\documentclass{article}

\PassOptionsToPackage{numbers, compress, square, comma}{natbib}
 \usepackage[preprint]{neurips_2026}

\usepackage[utf8]{inputenc} % allow utf-8 input
\usepackage[T1]{fontenc}    % use 8-bit T1 fonts
\usepackage[hidelinks]{hyperref}       % hyperlinks
\usepackage{url}            % simple URL typesetting
\usepackage{booktabs}       % professional-quality tables
\usepackage{amsfonts}       % blackboard math symbols
\usepackage{nicefrac}       % compact symbols for 1/2, etc.
\usepackage{microtype}      % microtypography
\usepackage{xcolor}         % colors

\usepackage{amsmath}
\usepackage{graphicx}
\usepackage{amssymb}
\usepackage{subcaption}

\renewcommand{\vec}[1]{\mathbf{#1}}

\usepackage{xcolor}

\newcommand{\Ltwonorm}[1]{\left\Vert{{#1}}\right\Vert_2}

\DeclareMathOperator*{\argmin}{argmin}

\usepackage{placeins}

\usepackage{amsmath}
\renewcommand{\vec}[1]{\mathbf{#1}}

\title{How Does Overparameterization Affect Machine Unlearning of Deep Neural Networks?}

\makeatletter
\newcommand{\correspondence}[1]{%
  \bgroup
  \let\thefootnote\relax % Clears the footnote symbol
  \footnotetext{#1}%     % Injects the text at the bottom
  \egroup
}
\makeatother

\author{%
  Gal Alon  and  Yehuda Dar\\
  Faculty of Computer and Information Science \\
  Ben-Gurion University of the Negev
}

\begin{document}

\maketitle
% --- Paste this block immediately AFTER \maketitle ---
\begingroup
\renewcommand\thefootnote{}%
\footnotetext{Correspondence to: \texttt{ydar@bgu.ac.il}}%
\endgroup
% ----------------------------------------------------

\begin{abstract}
Machine unlearning is the task of updating a trained model to forget specific training data without retraining from scratch. In this paper, we investigate how unlearning of deep neural networks (DNNs) is affected by the model parameterization level, which corresponds here to the DNN width. We define validation-based tuning for several unlearning methods from the recent literature, and show how these methods perform differently depending on (i) the DNN parameterization level, (ii) the unlearning goal (unlearned data privacy or bias removal), (iii) whether the unlearning method explicitly uses the unlearned examples. Our results show that unlearning usually excels on overparameterized models by significantly improving privacy/bias at a reasonable cost of utility (generalization) degradation; although for bias removal this requires the unlearning method to use the unlearned examples. Furthermore, we measure how much the unlearning changes the classification decision regions in the proximity of the unlearned examples, and avoids changing them elsewhere. By this we show that the unlearning success for overparameterized models stems from the ability to delicately change the model functionality in small regions in the input space while keeping much of the model functionality unchanged.
\end{abstract}

\section{Introduction}

Machine unlearning is motivated by regulatory requirements, such as the ``right to be forgotten'' in data protection laws like the GDPR \cite{GDPR}, which allows individuals to request the removal of their personal data from machine learning systems. While data can be forgotten by retraining the model from scratch on the original dataset without the data to unlearn, this is a computationally prohibitive approach. In contrast, machine unlearning offers a more computationally-efficient approach to forget specific data from a trained model while maintaining the model's overall performance.

Unlearning has increased importance for deep neural networks (DNNs) whose overparameterization facilitates training data memorization \cite{zhang2017understanding,arpit2017closer,maini2023can,carlini2021extracting,carlini2023extracting,radhakrishnan2020overparameterized,autoencoders_memorization_koren}.
Such memorization motivates unlearning applications. Memorization can increase privacy risks such as vulnerability to training data extraction \cite{carlini2021extracting,carlini2023extracting} and membership inference attacks \cite{parameters_or_privacy,MIA_shadow}; such privacy concerns can be addressed for selected training examples by unlearning them. Besides privacy issues, memorization increases the effect of biasing training examples on the model functionality, which can be corrected by unlearning the biasing examples.  Beyond the memorization implications, overparameterization increases the computational cost of training, making unlearning methods crucial as retraining from scratch becomes highly expensive or impossible in many cases.

\begin{figure*}
\centering
    \includegraphics[width=0.99\linewidth]{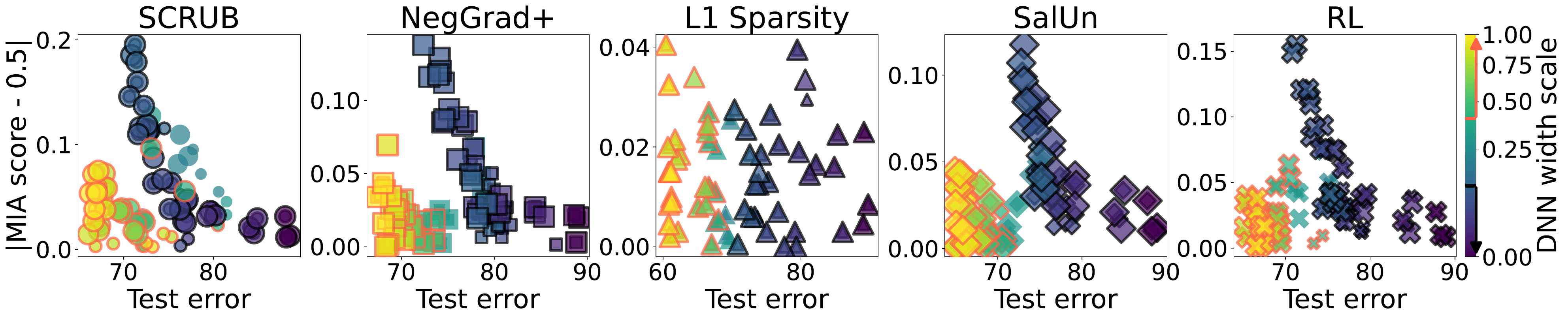}
    \caption{Unlearning results for \textbf{privacy} on ResNet-34, Tiny ImageNet, with 1000 unlearned examples. \textit{Ideal privacy-utility tradeoff is at the \textbf{bottom-left corner} of each diagram.} Markers' border line color: orange denotes overparameterization, black denotes underparameterization.}
    \label{fig:performance evaluation - resnet 50 tinet unlearn 1000 - privacy - main paper}
\end{figure*}
\begin{figure*}
\centering
    \includegraphics[width=0.99\linewidth]{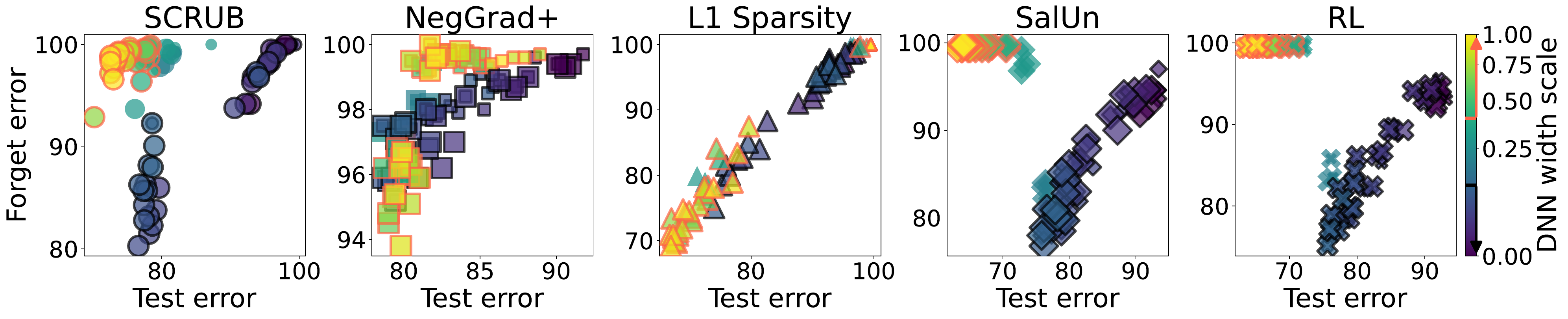}
    \caption{Unlearning results for \textbf{bias removal} on ResNet-34, Tiny ImageNet, with 1000 unlearned examples. \textit{Ideal bias-removal--utility tradeoff is at the \textbf{upper-left corner} of each diagram.}}
    \label{fig:performance evaluation - resnet 50 tinet unlearn 1000 - bias - main paper}
\end{figure*}

In this work, we empirically study how unlearning performance is affected by the parameterization level of the DNN model, ranging from underparameterized (where the model training error is nonzero) to overparameterized (where the model training error is zero, implying perfect fitting of training data). 
We show that the unlearning performance depends on the model parameterization level.

We examine unlearning methods from the recent literature, including SCRUB \cite{unbounded_MUL}, L1-sparsity aware unlearning \cite{l1_sparsity}, Saliency Unlearning (SalUn) \cite{salun}, Random Labeling (RL) \cite{eternal_sunshine}, Negative Gradient+ (NegGrad+) \cite{unbounded_MUL}.  As these methods were originally evaluated on DNNs at their standard width, we define here a validation-based hyperparameter-tuning procedure that makes the best out of each of the unlearning methods for the specific DNN width and unlearning goal. Specifically, we examine validation-based hyperparameter tuning for two different unlearning goals: 
\begin{enumerate}
    \item \textbf{Privacy of unlearned data:} The unlearned examples should not be identified as such w.r.t.~test data. Namely, the classification loss values on the unlearned data and on test data (from the same classes of the unlearned examples) should be close as possible such that loss-based membership inference attack would fail. 
    \item \textbf{Bias removal:} The model functionality should be changed, to predict other labels than those available in the biasing examples given to unlearn. Hence, the classification error on the unlearned examples should be worse as possible. 
\end{enumerate}
We implement each of these goals while balancing them with the basic need of generalization (i.e., model utility in the sense of a good classification performance on the overall test data). 

\noindent Our analysis provides the following insights:
\vspace{-0.01in}
\begin{enumerate}
    \item Unlearning usually excels on overparameterized models by significantly improving privacy/bias at a reasonable cost of utility degradation. 
    \item \textbf{Overparameterization promotes unlearning with a beneficial high-gain tradeoff.}  In contrast, \textbf{underparameterization restricts unlearning to low-gain tradeoffs.}
    \item Underparameterized models may have good unlearning privacy, if their original models already had good training data privacy. This allows unlearning methods to seemingly-succeed by operating a simpler tradeoff where low (or no) privacy gains are achieved in return to low utility degradation. 
    \item Overparameterization is important for successful bias removal, which requires an unlearning method that uses the forget set.
\end{enumerate}

For additional insights into the unlearning success on overparameterized models, we analyze the unlearning effect on the classification decision regions. 
We show that \textbf{overparameterization enables unlearning to modify the model functionality more delicately, by changing the decision regions locally around the unlearned examples without much change elsewhere.} 
This provides new insights into the inner-workings of unlearning methods, which are not observable by the applicative-performance analysis.

\section{Related Work}
\label{sec:related work}
%\subsection{The Overparameterization Effect on Classification DNNs}
\paragraph{The overparameterization effect on classification DNNs.} Recent research on overparameterized DNNs has explored how model architecture and parameterization influence the reproducibility and generalization of decision boundaries. A particularly relevant study by Somepalli et al.~\cite{somepalli_can_nn_learn_twice} showed the impact of the DNN width on decision regions formation and stability across different training runs. They showed that different DNN architectures, especially in terms of inductive bias, may have qualitatively and quantitatively different decision regions. Moreover, multiple training runs with the same architecture yield highly similar decision regions for highly overparameterized (wide) DNNs and noiseless data. 
\textit{In this research, we significantly extend the previous insights to show how overparameterization affects various unlearning methods in terms of decision region changes near the unlearned data.}

%\subsection{The Overparameterization Effect on Membership Inference}
\paragraph{The overparameterization effect on membership inference.} Tan et al.~\cite{parameters_or_privacy} studied the overparameterization effect on training data privacy by examining the inherent tradeoff between a model’s parameterization level and its vulnerability to membership inference attacks (MIA). Their work demonstrates that as the parameterization level of a model increases, so does its susceptibility to MIA, where adversaries can infer whether a particular example was part of the training dataset. This underscores the need for effective machine unlearning methods for overparameterized models, as these models are more likely to retain information about specific training examples and to risk their privacy.
\textit{However, existing research has not sufficiently explored how unlearning methods might address these privacy vulnerabilities in the context of overparameterization. Our work aims to bridge this gap by investigating how different machine unlearning techniques perform across models with varying parameterization levels. By examining unlearning performance on overparameterized DNNs, we aim to improve the understanding of tradeoffs among overparameterization, privacy, and unlearning effectiveness.}

\section{Machine Unlearning: Problem Definition and Existing Methods}
\label{sec:machine_unlearning_problem_definition_and_existing_methods}

\subsection{Definition of the Machine Unlearning Problem}

A training dataset \(\mathcal{D} = \{(\vec{x}_i, y_i)\}_{i=1}^N\) contains \(N\) examples. Each example \((\vec{x}_i, y_i)\) is a pair of an input vector \(\vec{x}_i \in \mathbb{R}^d\) and a label \(y_i \in \{1, \dots, C\}\). We define our model as the function \(f(\vec{x}; \vec{w})\) that gets $\vec{x}\in\mathbb{R}^d$ as its input and outputs a predicted label; here, \(\vec{w}\) represents the model’s learnable parameters that define the input-to-label mapping that $f$ implements. This function \(f\) corresponds to a DNN that is trained by supervised learning to minimize the loss over the dataset $\mathcal{D}$: $\mathcal{L}_{\rm train}\left(\vec{w}\right) = \frac{1}{N} \sum_{i=1}^N l_{\rm CE}(f(\vec{x}_i; \vec{w}), y_i)$, where \(l_{\rm CE}(\cdot,\cdot)\) denotes the cross-entropy loss.

In machine unlearning, we begin with an initial model \(f(\vec{x}; \vec{w}_0)\) that has been trained on the entire dataset \(\mathcal{D}\), where \(\vec{w}_0\) are the parameters obtained at the end of training. This model is the \textbf{original model} before any unlearning is applied. We further partition the training data into two subsets: The \textbf{forget set} \(\mathcal{D}_{\mathrm{f}} = \{(\vec{x}_i, y_i)\}_{i=1}^{N_\mathrm{f}}\) contains \(N_\mathrm{f}\) examples designated to be “forgotten”. The \textbf{retain set}  $\mathcal{D}_\mathrm{r} = \mathcal{D}\setminus\mathcal{D}_{\mathrm{f}}$ contains \(N_\mathrm{r}\triangleq N - N_\mathrm{f}\) examples that the model can continue to use. These subsets satisfy \(\mathcal{D}_\mathrm{f} \cup \mathcal{D}_\mathrm{r} = \mathcal{D}\) and \(\mathcal{D}_\mathrm{f} \cap \mathcal{D}_\mathrm{r} = \emptyset\).

The primary aim in machine unlearning is to modify the model’s parameters into new parameters \(\vec{w}_{\mathrm{u}}\), resulting in an \textbf{unlearned model} \(f(\vec{x}; \vec{w}_{\mathrm{u}})\). This unlearned model should, ideally, be unfamiliar with the specific information contained in the forget set \(\mathcal{D}_\mathrm{f}\) while preserving the accuracy and generalization performance of the original model on the retain set and other unseen test data.

\subsection{The Examined Unlearning Methods}
\label{subsec:The Examined Unlearning Methods}

Approximate unlearning methods aim to approximate a model retrained from scratch without the unlearned examples. This popular approach offers computational efficiency but may not fully remove the effect of the unlearned data. 
We consider the following approximate unlearning methods in this work: SCRUB \cite{unbounded_MUL}, Negative Gradient+ (NegGrad+) \cite{unbounded_MUL}, L1 Sparsity unlearning \cite{l1_sparsity}, Saliency-based Unlearning (SalUn) \cite{salun}, Random Labels (RL) \cite{eternal_sunshine}. See Appendix \ref{appendix:sec:Additional Details on Unlearning Methods} for more details, including formulations and hyperparameters. Importantly, unlike the other unlearning methods here,  L1 Sparsity uses only the retain set and does not use the forget set.

\subsection{Error and Loss Definitions}
\label{subsec:Error and Loss Definitions}
    In the following sections, we use a common formula for the classification error of a model \(f(\cdot; \vec{w})\) on a dataset  \(\widetilde{\mathcal{D}}\) that includes $\widetilde{N}$ statistically independent examples $(\vec{x}, y)$:
    \begin{equation}
        \label{eq:classification error - general formulation}
        \mathcal{E}\left(\vec{w};\widetilde{\mathcal{D}}\right) \triangleq \frac{1}{\left|\widetilde{\mathcal{D}}\right|} \sum_{\left(\vec{x},y\right)\in\widetilde{\mathcal{D}}} \mathbb{I}\left(f(\vec{x}; \vec{w})\ne y\right)
    \end{equation} 
     where $|\widetilde{\mathcal{D}}|=\widetilde{N}$, and $\mathbb{I}\left({\rm condition}\right)$ is an indicator function that returns 1 if the ${\rm condition}$ is satisfied, and 0 otherwise. Similarly, the cross-entropy loss on a dataset  \(\widetilde{\mathcal{D}}\) is 
      \begin{equation}
        \label{eq:cross-entropy loss on dataset - general formulation}
        \mathcal{L}\left(\vec{w};\widetilde{\mathcal{D}}\right) \triangleq \frac{1}{\left|\widetilde{\mathcal{D}}\right|} \sum_{\left(\vec{x},y\right)\in\widetilde{\mathcal{D}}} l_{\rm CE}(f(\vec{x}; \vec{w}), y).
    \end{equation}    
    Various datasets, which will be explicitly denoted, will take the role of \(\widetilde{\mathcal{D}}\) when we use (\ref{eq:classification error - general formulation}), (\ref{eq:cross-entropy loss on dataset - general formulation}).

\section{Validation-Based Tuning of Unlearning Hyperparameters}
\label{sec:Validation_Based _Tuning _of _Unlearning _Hyperparameters}

\subsection{Unlearning Goals}
\label{subsec:Unlearning Goals}
Motivated by \cite{unbounded_MUL}, we consider two \textbf{unlearning goals}: 
\paragraph{Privacy of unlearned data:}
    Unlearning supports users' ``right to be forgotten'' by enabling removal of specific examples from a model's training dataset. When the unlearning goal is to preserve privacy of unlearned examples, the unlearned model should be minimally vulnerable to membership inference attacks (MIAs) that try to identify unlearned examples (which could potentially expose private information). MIA usually works by identifying if a property (such as loss value) of an examined example is statistically more likely to be of an unlearned training example or not. Better privacy of unlearned examples in the sense of loss-based MIA implies that the unlearned model should ideally have its forget loss distribution as close as possible to its forget-test\footnote{The forget-test data is test data only from classes that have examples to unlearn in the forget set.} loss distribution; a larger discrepancy between these two loss distributions increases the unlearned model vulnerability to MIA. 
    The MIA accuracy represents the success rate of the attack in distinguishing the unlearned forget set examples from forget-test examples. This measures the extent to which the forget set examples can be identified as members or non-members of the training set based on their loss values. \textbf{Best privacy corresponds to MIA score of 0.5}.
    We use the loss-based MIA principles from \cite{forggeting_outside_the_box,unbounded_MUL}; more details are available in Appendix \ref{appendix:subsec:Membership Inference Attack (MIA) Methodology}.

    \paragraph{Bias removal:} Unlearning can resolve undesired model behavior that originates in specific training examples -- by directly unlearning these training examples. For this goal, the unlearned model's forget error should be (ideally) maximized; i.e., the model should no longer predict the labels associated with forgotten examples, to effectively remove unwanted biased functionality that the original examples induced. The \textbf{forget error} is the classification error for the unlearned examples \(\mathcal{D}_{\mathrm{f}} = \{(\vec{x}_i, y_i)\}_{i=1}^{N_\mathrm{f}}\), which is defined as $\mathcal{E}_{\rm forget} \left(\vec{w}\right)\triangleq \mathcal{E}\left(\vec{w};\mathcal{D}_{\mathrm{f}}\right)$.

The above unlearning goals are based on the forget error and forget loss that the unlearned model should ideally have. However, achieving the ideal forget error or loss might be limited by the need to maintain a good \textbf{utility} where the test error is sufficiently low. For a test set \(\mathcal{D}_{\rm test}\) with test examples that were not involved in the training nor the unlearning, the \textbf{test error} for a model \(f(\cdot; \vec{w})\) is $\mathcal{E}_{\rm test} \left(\vec{w}\right)\triangleq \mathcal{E}\left(\vec{w};\mathcal{D}_{\rm test}\right)$.
There is a possibly-challenging tradeoff between achieving the ideal unlearning goal and maintaining the utility (generalization performance) of the model.

\subsection{Validation-Based Hyperparameter Tuning}
\label{subsec:Validation-Based Hyperparameter Tuning}

As explained in the previous subsection, the performance criteria for unlearning are more intricate than for standard learning and depend on the specific unlearning goal (privacy or bias removal). Therefore, we define \textit{validation-based hyperparameter tuning} to optimize the performance of each unlearning method for the specific DNN architecture, parameterization level (width), and classification task (dataset) of the original model. This tuning is crucial, as the examined methods were originally tuned for specific settings and widths that differ from ours. Without it, suboptimal hyperparameters can severely degrade an unlearning method's performance.

We propose two new optimization objectives for validation-based tuning of unlearning hyperparameters, one objective is for the privacy goal and the second is for the bias removal goal. Both of these objectives aim to balance  between the unlearning goal and maintaining a good test error, which is estimated by a validation error using a validation set \(\mathcal{D}_{\rm val}\) that includes examples $(\vec{x}, y)$ not from the training set nor the test set.

 We denote the hyperparameters of an unlearning method $U$ as $\Psi$; the corresponding unlearned model is \(f(\cdot; \vec{w}_{\Psi})\) where $\vec{w}_{\Psi}$ is the unlearned parameter vector obtained by applying the unlearning method on the original parameter vector $\vec{w}_0$, i.e., $\vec{w}_{\Psi}\triangleq U(\vec{w}_0;\Psi)$. 
See Appendix \ref{grid search details} for more details on the hyperparameters of each unlearning method and their grid search.

Then, for a given unlearning method $U$ and a given discrete (finite) grid $\mathcal{G}_{\psi}$ of hyperparameter values to choose from, the validation-based tuning corresponds to solving 
\begin{equation}
    \label{eq:validation-based tuning of unlearning hyperparameters - argmin formula}
    \widehat{\Psi} = \argmin_{\Psi\in \mathcal{G}_{\psi}} S(\Psi;\vec{w}_0, \mathcal{D}_{\rm val},\mathcal{D}_{\mathrm{f}})
\end{equation}
where $S(\Psi;\vec{w}_0,\mathcal{D}_{\rm val},\mathcal{D}_{\mathrm{f}})$ is a validation score for the specific hyperparameter values $\Psi$; a lower score $S(\Psi;\vec{w}_0,\mathcal{D}_{\rm val},\mathcal{D}_{\mathrm{f}})$ implies better parameters $\Psi$ for unlearning. The original model parameters $\vec{w}_0$, the validation set $\mathcal{D}_{\rm val}$, the forget set $\mathcal{D}_{\mathrm{f}}$, the hyperparameter grid $\mathcal{G}_{\psi}$, and the unlearning method $U$ (which for brevity is not explicitly denoted here) are fixed in a hyperparameter tuning task. 

The validation score has the form of 
\begin{equation}
    \label{eq:validation score formula}
    S(\Psi;\vec{w}_0, \mathcal{D}_{\rm val},\mathcal{D}_{\mathrm{f}})\triangleq (1 - \lambda) \cdot 
S_{\rm goal}(\Psi;\vec{w}_0, \mathcal{D}_{\rm val},\mathcal{D}_{\mathrm{f}}) +   \lambda\cdot
S_{\rm util}(\Psi;\vec{w}_0, \mathcal{D}_{\rm val})
\end{equation}
where $S_{\rm goal}$ is a validation-proxy measure for the unlearning goal (will be defined below for privacy or bias removal), and $S_{\rm util}$ is a validation-proxy measure for utility degradation compared to the original model: 
\begin{equation}
    \label{eq:validation utility formula}
    S_{\rm util}(\Psi;\vec{w}_0, \mathcal{D}_{\rm val}) \triangleq \max\{0,\mathcal{E}\left(\vec{w}_{\Psi};\mathcal{D}_{\mathrm{val}}\right)-\mathcal{E}\left(\vec{w}_{0};\mathcal{D}_{\mathrm{val}}\right)\}.
\end{equation}
Note that $S_{\rm util}$ measures difference between the validation error of the unlearned model and the  original model; negative differences are clipped to zero, to promote the unlearning goal instead of strictly improving the original model utility. 
The validation parameter $\lambda\in(0,1)$ in (\ref{eq:validation score formula}) adjusts the balance between the unlearning goal and utility preservation; a lower $\lambda$ prioritizes more the unlearning goal (privacy or bias removal) over the utility (generalization performance, low test error close to the original model).

For \textbf{privacy of unlearned data}, we define the validation-proxy measure function $S_{\rm goal}(\Psi;\vec{w}_0, \mathcal{D}_{\rm val},\mathcal{D}_{\mathrm{f}})$ for (\ref{eq:validation score formula}) as 
\begin{align}
\label{eq:privacy_score_formula}
&S_{\rm goalP}(\Psi;\vec{w}_0,\mathcal{D}_{\rm val},\mathcal{D}_{\mathrm{f}}) =  
\Big|
\mathcal{L}\!\left(\vec{w}_{\Psi};\mathcal{D}_{\mathrm{f}}\right)
- \mathcal{L}\!\left(\vec{w}_{\Psi};\mathcal{D}_{\mathrm{valF}}\right)
\Big|
\end{align}
where $\vec{w}_{\Psi}\triangleq U(\vec{w}_0;\Psi)$ is the model parameter vector obtained from unlearning with hyperparameters $\Psi$; $\mathcal{L}$ is the function defined in (\ref{eq:cross-entropy loss on dataset - general formulation}) for computing the CE loss of a given model on a given dataset. 
Specifically, (\ref{eq:privacy_score_formula}) approximates the privacy level by the absolute difference between the forget loss $\mathcal{L}\left(\vec{w}_{\Psi};\mathcal{D}_{\mathrm{f}}\right)$ on the unlearned data $\mathcal{D}_{\mathrm{f}}$ and the validation forget loss $\mathcal{L}\left(\vec{w}_{\Psi};\mathcal{D}_{\mathrm{valF}}\right)$ that considers validation examples $\mathcal{D}_{\mathrm{valF}}\subseteq\mathcal{D}_{\mathrm{val}}$ only from the classes that have examples to unlearn, i.e., $\mathcal{D}_{\mathrm{valF}}=\{(\vec{x},y)\in\mathcal{D}_{\mathrm{val}} ~\text{such that}~\exists (\cdot,y)\in\mathcal{D}_{\mathrm{f}}\}$.

\vspace{0.1in}
For \textbf{bias removal}, we define the measure function $S_{\rm goal}(\Psi;\vec{w}_0, \mathcal{D}_{\rm val},\mathcal{D}_{\mathrm{f}})$ for (\ref{eq:validation score formula}) as
\begin{align}
\label{eq:unlearning_score_bias}
    &S_{\rm goalB}(\Psi;\vec{w}_0,\mathcal{D}_{\mathrm{f}}) = - 
    \mathcal{E}\!\left(\vec{w}_{\Psi};\mathcal{D}_{\mathrm{f}}\right).
\end{align}
This is the unlearned model's forget error, preceded by a minus sign to promote a higher forget error (thus better bias removal) in the minimization problem (\ref{eq:validation-based tuning of unlearning hyperparameters - argmin formula}).

\section{Performance-Based Analysis}
\label{sec:error_based_analysis}

Next, we explore the relation between unlearning performance and the model parameterization level. 

\paragraph{The examined models.}
We examine several DNN architectures: ResNet-18, ResNet-34, and a simple 3-layer fully-connected (FC) network across a range of parameterization levels. As in related overparameterization research \citep{nakkiran2019deep,somepalli_can_nn_learn_twice}, the parameterization level is implemented by varying the width of a given architecture (see Appendix \ref{appendix:Additional Experiment Details}). 
Our experiments are for the image classification datasets CIFAR-10 \cite{cifar10} and Tiny ImageNet \cite{tiny_imagenet}. Considering the computationally demanding validation-based tuning process for each parameterization level and unlearning method in each experiment (see Appendix \ref{grid search details}), we run each experiment 3 times with different random seeds and report the results as explained for each of the analysis parts. See Appendix \ref{appendix:Additional Experiment Details} for more details.

\subsection{How is Unlearning Performance Affected by the Parameterization Level?}
\label{How is Unlearning Performance Affected by the Parameterization Level?}
Here, we explore how the parameterization level affects the unlearning performance as reflected by pairing the test error (which indicates utility) and a score for the specific unlearning goal:

 The \textbf{privacy} performance is measured by the MIA score (as defined in Section \ref{subsec:Unlearning Goals}). A larger absolute difference of the MIA score from the ideal 0.5 value implies that the unlearned model is more vulnerable to loss-based membership inference attacks on the unlearned examples; thus better privacy is associated here with absolute difference closer to zero.
  
Better \textbf{bias removal} is reflected by a higher forget error, implying that the model functionality changed for more of the biasing examples that compose the forget set. 

Our results in Figs.~\ref{fig:performance evaluation - resnet 50 tinet unlearn 1000 - privacy - main paper}, \ref{fig:performance evaluation - resnet 50 tinet unlearn 1000 - bias - main paper}, \ref{fig:performance results - tinet resnet50 unlearn 1000 - privacy}-\ref{fig:performance results - fcnet3layer unlearn 200 - bias} show how different unlearning methods perform at different parameterization levels with respect to the tradeoff between generalization and achieving the unlearning goal. Specifically, in \textit{these} scatter plots, \textbf{the ideal performance for the \textit{privacy} goal is at the \textit{lower}-left corner of the diagram}, where both generalization and privacy are best;
\textbf{the ideal performance for the \textit{bias removal} goal is at the \textit{upper}-left corner of the diagram}, where both generalization and functionality change are best.

We now describe the three evaluation formats used to present our performance analysis; subsequently, in Sections \ref{subsec:When Can Unlearning Succeed on Overparameterized vs.~Underparameterized Models}-\ref{subsec:Insights into the Unlearning Tradeoff: Overparameterized vs. Underparameterized} we discuss the insights drawn from these evaluations.

\paragraph{Evaluation of parameterization levels.} Figures~\ref{fig:performance evaluation - resnet 50 tinet unlearn 1000 - privacy - main paper}, \ref{fig:performance evaluation - resnet 50 tinet unlearn 1000 - bias - main paper}, \ref{fig:performance results - tinet resnet50 unlearn 1000 - privacy}-\ref{fig:performance results - fcnet3layer unlearn 200 - bias} show the evaluated performance for various parameterization levels, corresponding to markers of different colors.
DNN width scale $1$ corresponds to the architecture at its standard width, here having the maximal parameterization level that appears as the yellowest markers. Smaller width scales correspond to narrower versions of the architecture. We denote overparameterization vs.~underparameterization as follows:
Widths whose original model's average train error is less than $0.1\%$ are considered \textbf{overparameterized} (due to memorization of at least $99.9\%$ of the training examples), the corresponding markers have orange border lines and the minimal width that is considered overparameterized is denoted by an orange threshold in the colorbar. In contrast, widths whose average train error is higher than $10\%$ are considered \textbf{underparameterized}; the corresponding markers are denoted by black border lines, which also appears in the colorbar as a threshold indicating the maximal width that is considered underparameterized. The marker size corresponds to the $\lambda$ value that balances between the unlearning goal and generalization in the validation score function (\ref{eq:validation score formula}). See Appendices \ref{appendix:Additional Experiment Details}, \ref{appendix: Additional Error-based Experiments} for further details and experiments on more model architectures and datasets, and the train errors of the original models in Fig.~\ref{fig:original_models_train_error_vs_test_error}. 

\begin{figure*}[t]
    \centering
    \hspace{-0.6in}
    \begin{subfigure}[b]{0.24\linewidth}
        \centering
        \includegraphics[height=0.12\textheight]{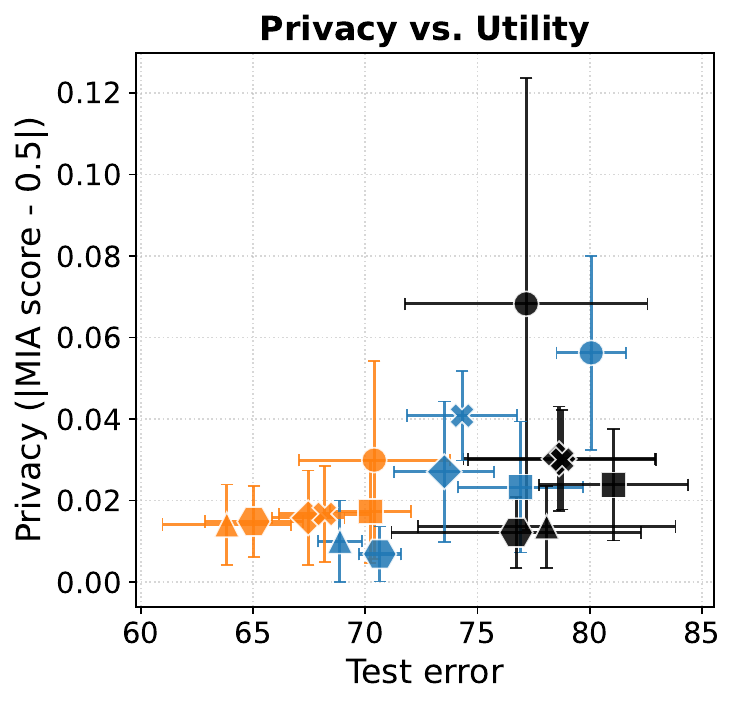}
        \caption{}
        \label{fig:category performance results - tinet resnet50 unlearn 1000 - main paper - privacy raw}
    \end{subfigure}
    \hspace{-0.18in}
    \begin{subfigure}[b]{0.24\linewidth}
        \centering
        \includegraphics[height=0.12\textheight]{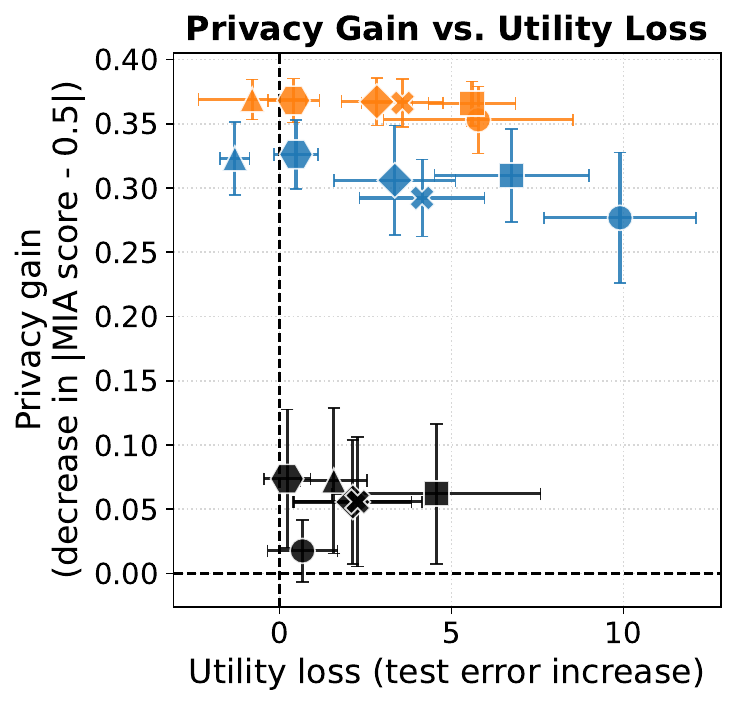}
        \caption{}
        \label{fig:category performance results - tinet resnet50 unlearn 1000 - main paper - privacy delta}
    \end{subfigure}
    \hspace{-0.18in}
    \begin{subfigure}[b]{0.24\linewidth}
        \centering
        \includegraphics[height=0.12\textheight]{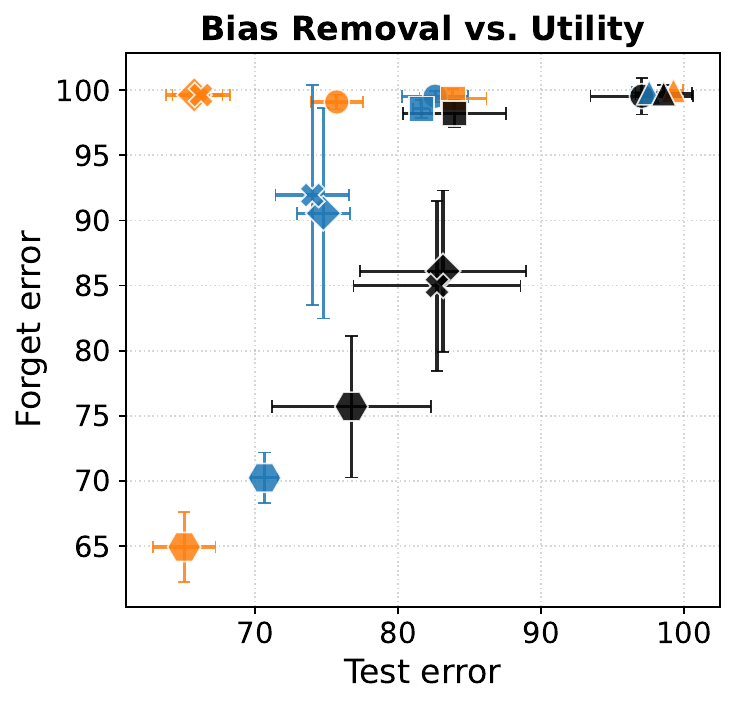}
        \caption{}
        \label{fig:category performance results - tinet resnet50 unlearn 1000 - main paper - bias raw}
    \end{subfigure}
    \hspace{-0.12in}
    \begin{subfigure}[b]{0.24\linewidth}
        \centering
        \includegraphics[height=0.12\textheight]{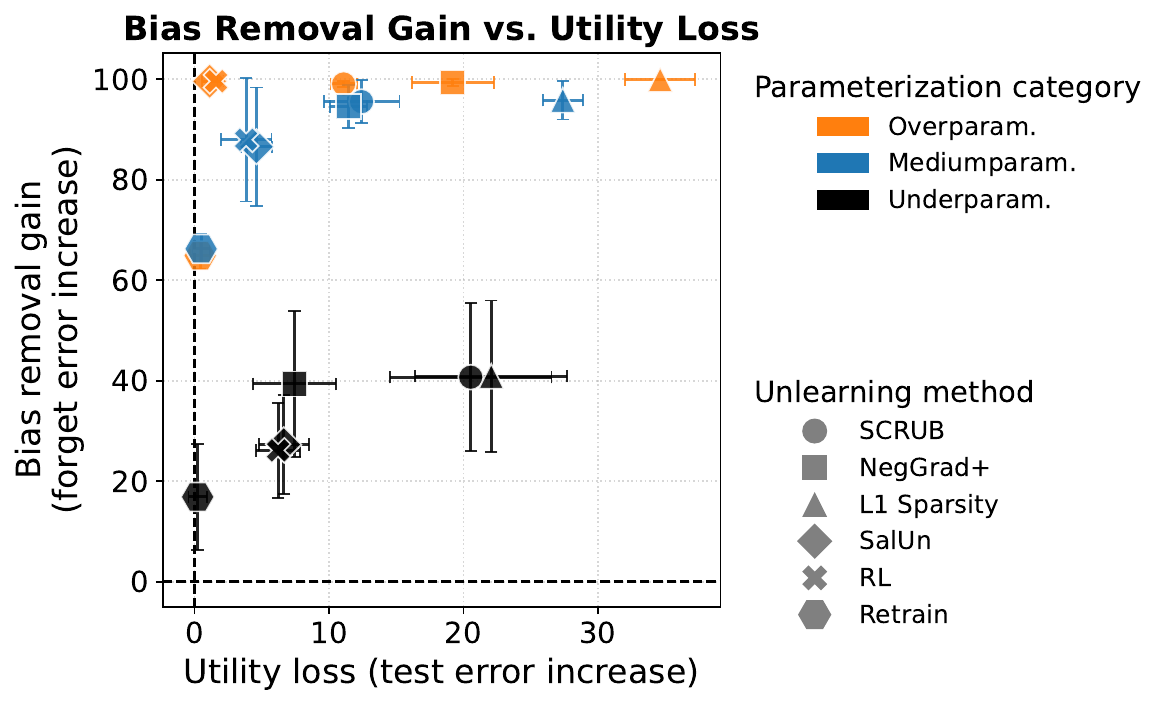}
        \caption{}
        \label{fig:category performance results - tinet resnet50 unlearn 1000 - main paper - bias delta}
    \end{subfigure}
    \vspace{-0.08in}
    \caption{Evaluation of parameterization categories and evaluation of unlearned vs.~original performance (ResNet-34, Tiny ImageNet, 1000 unlearned examples). The ideal performance corners: bottom-left for diagram (a), top-left for diagrams (b),(c),(d).}
     \label{fig:category performance results - tinet resnet50 unlearn 1000 - main paper}
\end{figure*}

\paragraph{Evaluation of parameterization categories: overparameterized vs.~underparameterized.} In addition to showing each of the evaluated parameterization levels, we show the results of the parameterization categories: overparameterized, medium-parameterized, underparameterized. Accordingly, in Figs.~\ref{fig:category performance results - tinet resnet50 unlearn 1000 - main paper - privacy raw}, \ref{fig:category performance results - tinet resnet50 unlearn 1000 - main paper - bias raw}, and in the left-side diagrams in Figs.~\ref{fig:category performance results - tinet resnet50 unlearn 1000 - privacy}-\ref{fig:category performance results - cifar10 fcnet3layer unlearn 200 - bias}, each marker denotes the \textit{average evaluation of all models that belong to a parameterization category} (by satisfying the above described train error conditions) and a specific unlearning method (denoted by the marker shape); the error bars denote standard deviations. This allows us to examine the performance of all the overparameterized models (orange markers) versus all the underparameterized models (black markers).

\paragraph{Evaluation of unlearned vs.~original performance.} Figs.~\ref{fig:category performance results - tinet resnet50 unlearn 1000 - main paper - privacy delta}, \ref{fig:category performance results - tinet resnet50 unlearn 1000 - main paper - bias delta}, and the right-side diagrams in Figs.~\ref{fig:category performance results - tinet resnet50 unlearn 1000 - privacy}-\ref{fig:category performance results - cifar10 fcnet3layer unlearn 200 - bias} show the performance of the unlearned models \textit{relative to their corresponding original models}. This shows how each unlearning method traded-off the improvements towards achieving its goal (privacy or bias removal) at the cost of degrading utility (test error). These figures provide \textit{clear insights into the underlying effectivity and activity levels that unlearning methods have depending on whether the model is overparameterized or underparameterized.}

\subsection{When Can Unlearning Succeed on Overparameterized vs.~Underparameterized Models?} 
\label{subsec:When Can Unlearning Succeed on Overparameterized vs.~Underparameterized Models}
\textbf{Unlearning excels on overparameterized models.} 
Overparameterized models (denoted by orange-border-lined or orange-filled markers) are usually closer to the ideal corner of the diagram, where both generalization and achieving the unlearning goal are best. Some of the unlearning methods may have a more noticeable variance in their privacy (MIA score) of overparameterized models, but even then there are overparameterized models that achieve the best performance in the diagram. 

\textbf{Underparameterized models may have good unlearning privacy, if their original models already had good training data privacy.}
Interestingly, for the privacy goal, underparameterized models with a relatively good original privacy level usually generalize worse than overparameterized models, but can still achieve relatively good unlearning privacy (reflected by having markers near the bottom of the scatter plots, but not necessarily near the left corner; see, e.g., Fig.~\ref{fig:performance evaluation - resnet 50 tinet unlearn 1000 - privacy - main paper}). This is because, for these specific cases (which do not represent all the underparameterized models), the original model before unlearning already had relatively good privacy; therefore, the unlearning method does not need to change the model too much to increase its privacy, it just needs to properly maintain its original privacy level --- and this is doable in unlearning of underparameterized models (as can be observed by the low privacy gains and low utility degradation of underparametrized models, e.g., in Fig.~\ref{fig:category performance results - tinet resnet50 unlearn 1000 - main paper - privacy delta} and the right-side diagram of Figs.~\ref{fig:category performance results - tinet resnet50 unlearn 1000 - privacy}, \ref{fig:category performance results - tinet resnet18 unlearn 500 - privacy}, \ref{fig:category performance results - cifar10 resnet18 unlearn 200 - privacy}, \ref{fig:category performance results - cifar10 resnet18 unlearn 600 - privacy}, \ref{fig:category performance results - cifar10 fcnet3layer unlearn 200 - privacy}). This sharply contrasts the overparameterized case where unlearning should actively change the model to increase its privacy, due to the original overparameterized models having significant differences between their forget loss and forget test loss, a consequence of perfect-fitting training examples (implying having forget loss near zero whereas the forget test loss remains much higher).

\textbf{Overparameterization is important for successful bias removal.}
Our bias removal results show that the forget error can be intentionally increased more successfully (without overly increasing the test error) for overparameterized models than for underparameterized models; see, e.g., Figs.~\ref{fig:performance evaluation - resnet 50 tinet unlearn 1000 - bias - main paper}, \ref{fig:category performance results - tinet resnet50 unlearn 1000 - main paper - bias raw}, \ref{fig:category performance results - tinet resnet50 unlearn 1000 - main paper - bias delta}. This is due to the high functional complexity of overparameterized models, which yields perfect fitting of training data -- that increases risks of biasing examples, but also makes their removal easier. This easier bias removal can be perceived as changing the model functionality locally in the input space around the biasing examples, and without affecting too much the overall functionality and generalization performance (this viewpoint will be further studied in the Section \ref{sec:Unlearning Analysis using Decision Regions}).

\textbf{Successful bias removal in overparameterized models requires unlearning method that uses the forget set.}
An important exception for the great unlearning performance on overparameterized models is observed for the L1-sparsity unlearning method that does not use the forget set and therefore fails in the bias removal goal (see that the overparameterized markers of \textit{L1-Sparsity} have relatively large distance from the upper-left corner in the bias-removal diagrams, e.g., Figs.~\ref{fig:performance evaluation - resnet 50 tinet unlearn 1000 - bias - main paper}, \ref{fig:category performance results - tinet resnet50 unlearn 1000 - main paper - bias raw}, \ref{fig:category performance results - tinet resnet50 unlearn 1000 - main paper - bias delta}).
%; interestingly, this behavior also appears in the retrain-from-scratch diagrams, as the forget set is not used there either).
This failure in bias removal can be more consequential for overparameterized models than for underparameterized models, due to the zero (or very low) training error of overparameterized models that yields memorization of biasing examples; underparameterized models have limited model complexity that prevents memorization and thus inherently provides some (although usually insufficient) protection against biasing examples. This exemplifies that choosing a suitable unlearning method might be crucial for some goals, here we show it for bias removal.

\subsection{Insights into the Unlearning Tradeoff: Overparameterized vs.~Underparameterized}
\label{subsec:Insights into the Unlearning Tradeoff: Overparameterized vs. Underparameterized}
\textbf{Underparameterization restricts unlearning to low-gain tradeoffs.} The results show that underparameterized models have relatively low gains in achieving their privacy and bias removal goals (see the black markers in Figs.~\ref{fig:category performance results - tinet resnet50 unlearn 1000 - main paper - privacy delta}, \ref{fig:category performance results - tinet resnet50 unlearn 1000 - main paper - bias delta}, and in the right-side diagrams in Figs.~\ref{fig:category performance results - tinet resnet50 unlearn 1000 - privacy}-\ref{fig:category performance results - cifar10 fcnet3layer unlearn 200 - bias}). These low gains are due to the low model complexity that constrains the validation-tuned unlearning methods to opt for low privacy/bias-removal gains at the cost of relatively low utility degradation.

\textbf{Overparameterization promotes unlearning of high-gain tradeoffs.} The results show that overparameterized models enable the validation-tuned unlearning methods to achieve high gains in privacy/bias-removal usually at a reasonable cost of utility degradation. This makes the unlearning performance tradeoffs much more beneficial (orange markers closer to the ideal corner in Figs.~\ref{fig:category performance results - tinet resnet50 unlearn 1000 - main paper - privacy delta}, \ref{fig:category performance results - tinet resnet50 unlearn 1000 - main paper - bias delta}, and in the right-side diagrams in Figs.~\ref{fig:category performance results - tinet resnet50 unlearn 1000 - privacy}-\ref{fig:category performance results - cifar10 fcnet3layer unlearn 200 - bias}) in the overparameterized case than in the underparameterized case.  This improved unlearning tradeoff is enabled by the high complexity of overparameterized models; the high model complexity makes the original models more prone to privacy and bias issues, but it also unleashes the unlearning ability to impressively resolve them.

Our insights emphasize that machine learning systems that employ unlearning tasks may often benefit from using overparameterized models.

\section{Unlearning Analysis using Decision Regions}
\label{sec:Unlearning Analysis using Decision Regions}

The above performance-based analysis elucidates the high-level behavior of unlearning performance, which naturally raises questions about the inner-workings of unlearning. Hence, here we \textbf{empirically analyze how unlearning affects the decision regions of a trained classifier and how this depends on the model’s parameterization level}.

One may argue that a good unlearning method should update the model such that   
at the \textbf{proximity} of unlearned training samples, the learned function (e.g., the predicted class label) can (or should, depending on the unlearning goal) be \textit{modified}, in order to make the model forget the specified training examples; 
elsewhere (i.e., \textbf{sufficiently far} from the unlearned training examples), the learned function (e.g., the predicted class label) should be kept \textit{unchanged}, to avoid unnecessary modifications in the model.

\subsection{Similarity and Change Scores}
\label{subsec:decision region analysis - Case I - Unlearning of Specific Training Samples}
For methods that unlearn a subset of training examples $\mathcal{D}_{\mathrm{f}} = \{(\vec{x}_i, y_i)\}_{i=1}^{N_\mathrm{f}}$, we will define a \textit{similarity score} that excludes input-space regions that correspond to $\delta$-small neighborhoods around the unlearned examples from $\mathcal{D}_{\mathrm{f}}$:   
\begin{equation}
\label{eq: unlearning similarity score - unlearning of specific samples}
\begin{split}
    &S_{\rm far}(f_0, f_{\rm u}; \mathcal{D}_{\mathrm{f}}, \delta)
    =
\frac{1}{m} \sum_{j=1}^{m} 
    \frac{1}{|\mathcal{P}_{j,\text{far}}(\mathcal{D}_{\mathrm{f}}, \delta)|}  
    \sum_{\vec{x} \in \mathcal{P}_{j,\text{far}}(\mathcal{D}_{\mathrm{f}}, \delta)} 
    \mathbb{I}\!\left(f_0(\vec{x}) = f_{\rm u}(\vec{x})\right)
\end{split}
\end{equation}
\begin{equation}
\label{eq: unlearning similarity score - unlearning of specific samples - regions far from unlearned samples}
\begin{split}
    &\text{where }~~\mathcal{P}_{j,\text{far}}(\mathcal{D}_{\mathrm{f}}, \delta)
    = \bigl\{\, \vec{x} \in \mathcal{P}_j ~\big|~
    \Ltwonorm{\vec{x}-\vec{x}_{\mathrm{u}}} > \delta,
    ~\text{for any } (\vec{x}_{\mathrm{u}}, y_{\mathrm{u}}) \in \mathcal{D}_{\mathrm{f}}
    \,\bigr\}.
\end{split}
\end{equation}   
Here, $f_0 (\cdot)\triangleq f(\cdot;\vec{w}_0)$ and $ f_{\rm u} (\cdot)\triangleq f(\cdot;\vec{w}_{\mathrm{u}})$ are the classification functions before and after unlearning, respectively.
$\mathbb{I}({\rm condition})$ is an indicator function that returns 1 if the ${\rm condition}$ holds, and 0 otherwise. Here, $\mathcal{P}_j$ is a discrete grid over a plane which is spanned by a randomly chosen triplet of training examples of $f_0$ (specifically, $\mathcal{P}_j$ is a rectangular part of the plane that includes the three training examples, one of them is an unlearned example); we consider $m$ independently chosen training example triplets and, hence, there are $m$ planes over which we examine the decision regions. We use $m=300$ planes, each with a discrete uniform grid of $2500$ points. Note that our definition in Eq.~(\ref{eq: unlearning similarity score - unlearning of specific samples}) is an updated version of the similarity score that was used in \cite{somepalli_can_nn_learn_twice}  for comparing classifiers that are trained using different random seeds or different architectures; importantly, here, we update the previous similarity score to disregard the input regions around each of the unlearned examples (see the definition of $\mathcal{P}_{j,\text{far}}(\mathcal{D}_{\mathrm{f}}, \delta)$ in Eq.~(\ref{eq: unlearning similarity score - unlearning of specific samples - regions far from unlearned samples})).  
    
Here, $\delta>0$ is a constant that determines the size of the region around the unlearned examples. The experiments that compute Eq.~(\ref{eq: unlearning similarity score - unlearning of specific samples}) were done for several different values of $\delta$; we show the results for $\delta=10$. For a proper value of $\delta>0$, a higher similarity score $S_{\rm far}(f_0, f_{\rm u}; \mathcal{D}_{\mathrm{f}}, \delta)$ reflects that the unlearning preserves more of the originally learned function $f_0$ in areas that are at least $\delta$-far from unlearned examples; such avoidance of unnecessary changes in the original model is a desired property for a good unlearning method.
Complementarily to Eq.~(\ref{eq: unlearning similarity score - unlearning of specific samples}), we also define a \textit{change score} for areas in the $\delta$-proximity of the unlearned examples. We formulate this change score as 
    \begin{align}
        \label{eq:unlearning_change_score}
        &C_{\rm prox}(f_0, f_{\rm u}; \mathcal{D}_{\mathrm{f}}, \delta)
        =
\frac{1}{m} \sum_{j=1}^{m} 
        \frac{1}{|\mathcal{P}_{j,\text{prox}}(\mathcal{D}_{\mathrm{f}}, \delta)|}  \sum_{\vec{x} \in \mathcal{P}_{j,\text{prox}}(\mathcal{D}_{\mathrm{f}}, \delta)} 
        \mathbb{I}\big(f_0(\vec{x}) \ne f_{\rm u}(\vec{x})\big)
    \end{align}
where $\mathcal{P}_{j,\text{prox}}(\mathcal{D}_{\mathrm{f}}, \delta)= \mathcal{P}_j \setminus \mathcal{P}_{j,\text{far}}(\mathcal{D}_{\mathrm{f}}, \delta)$ includes inputs from $\mathcal{P}_j$ that are $\delta$-close to unlearned examples.
    Then, for a proper value of $\delta>0$, a higher change score $C_{\rm prox} (f_0, f_{\rm u}; \mathcal{D}_{\mathrm{f}}, \delta)$ reflects that the unlearning process modifies more of the originally learned function $f$ in areas that are \textit{close to the unlearned examples}; one may argue that such changes in the original model are allowed (or even necessary for bias removal) to successfully unlearn the requested examples.

\begin{figure*}[]
    \centering    
    \includegraphics[height=0.185\textwidth]{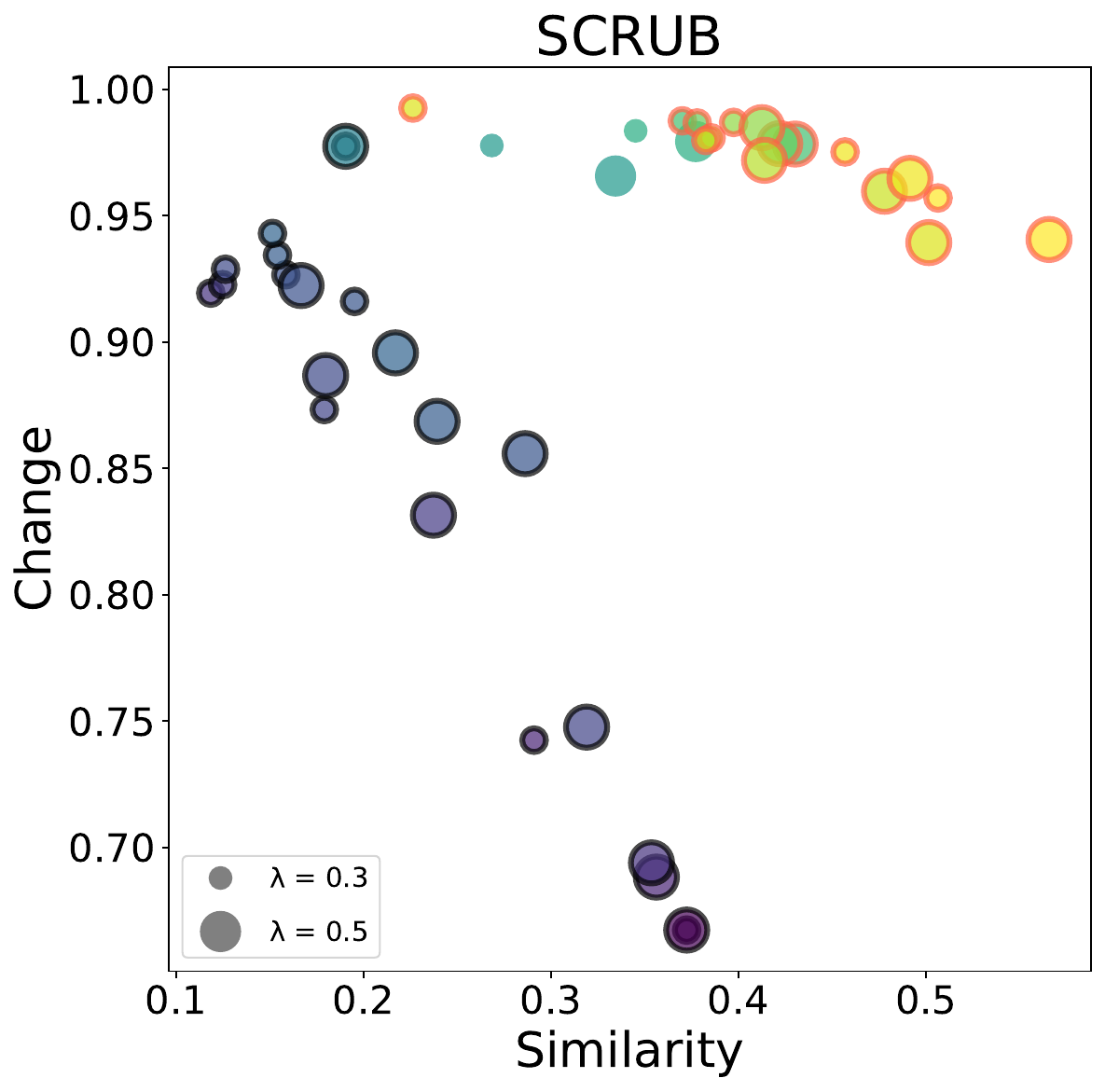}
    \includegraphics[height=0.185\textwidth]{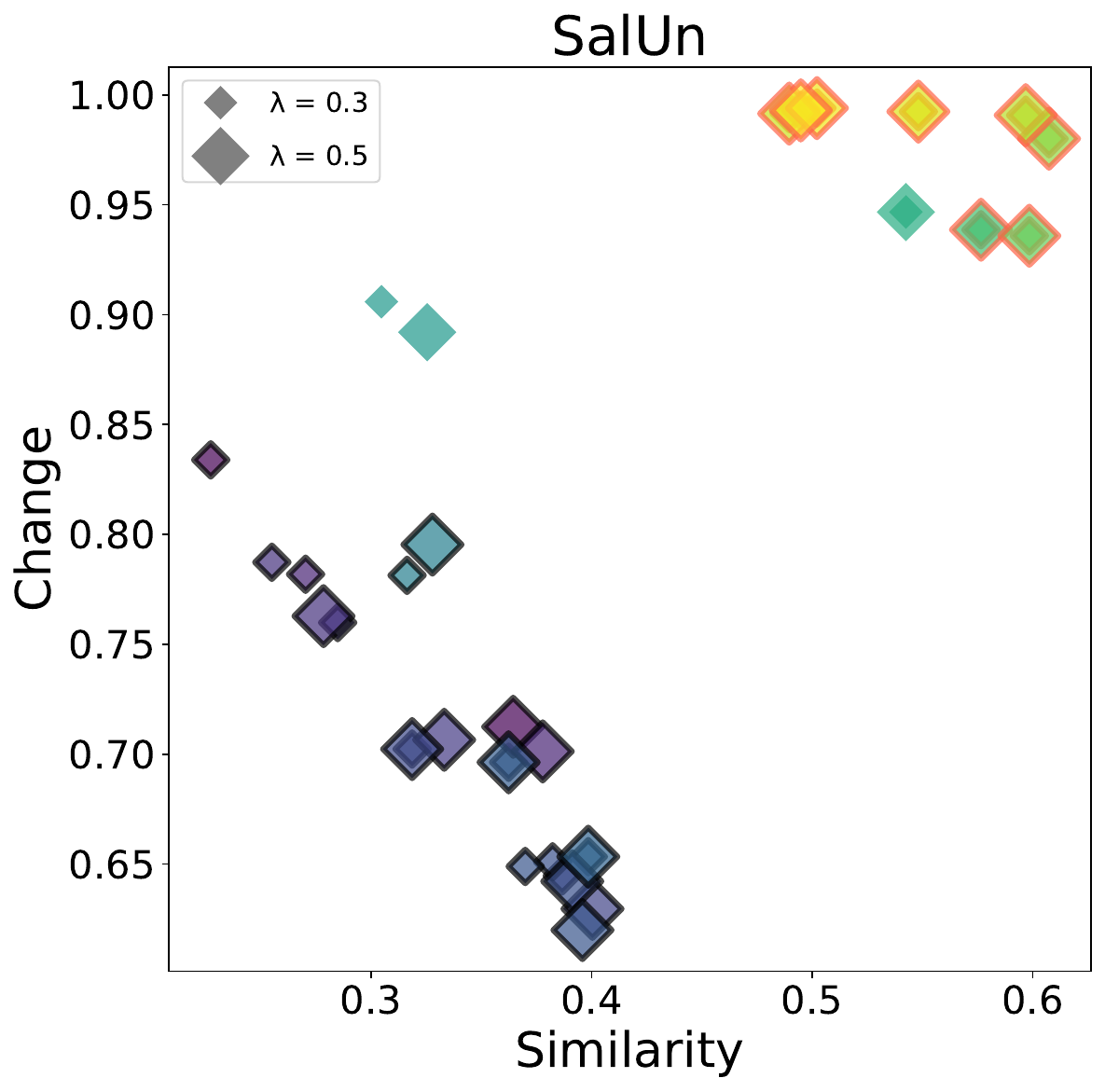}
    \includegraphics[height=0.185\textwidth]{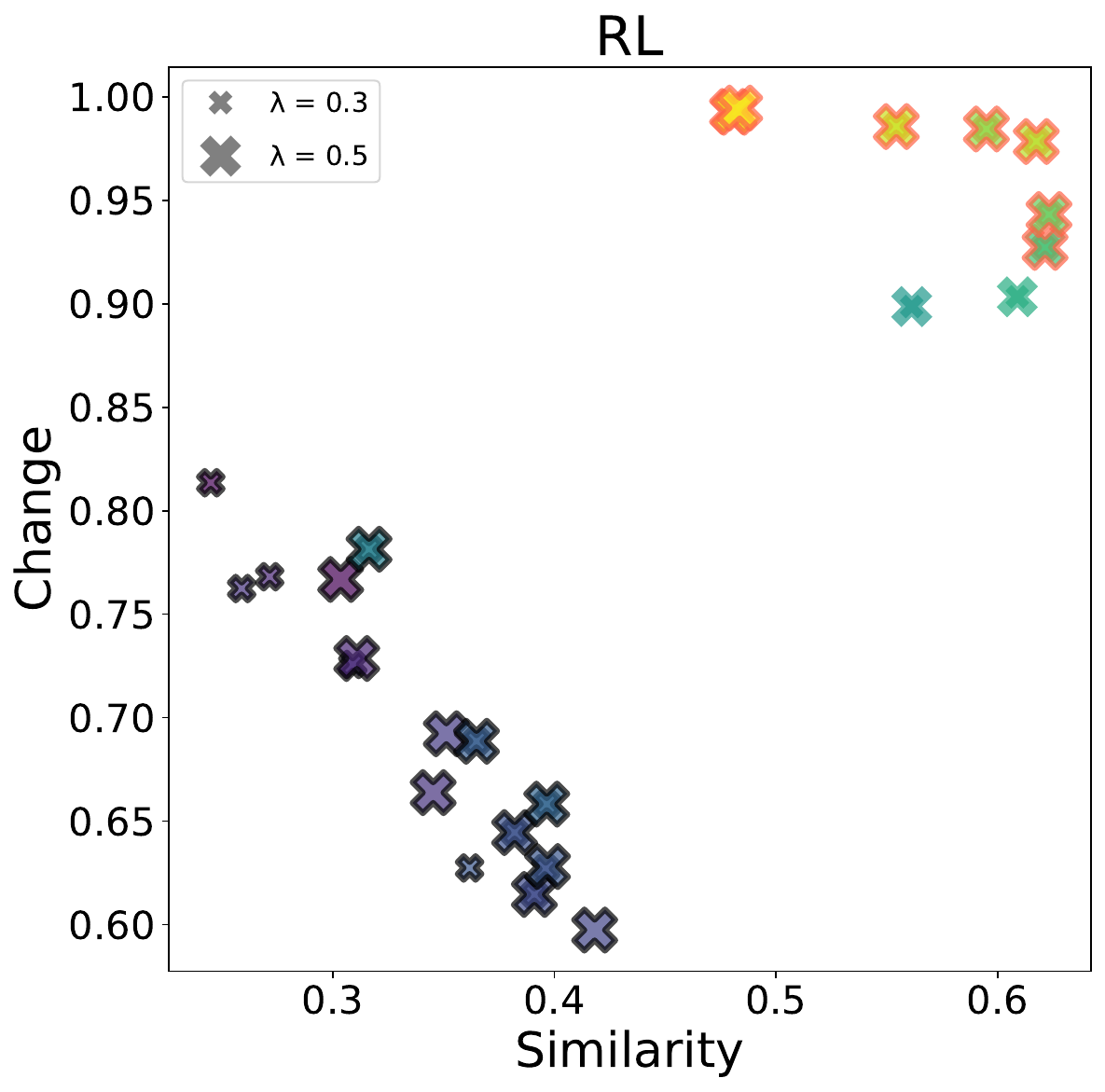}
    \includegraphics[height=0.185\textwidth]{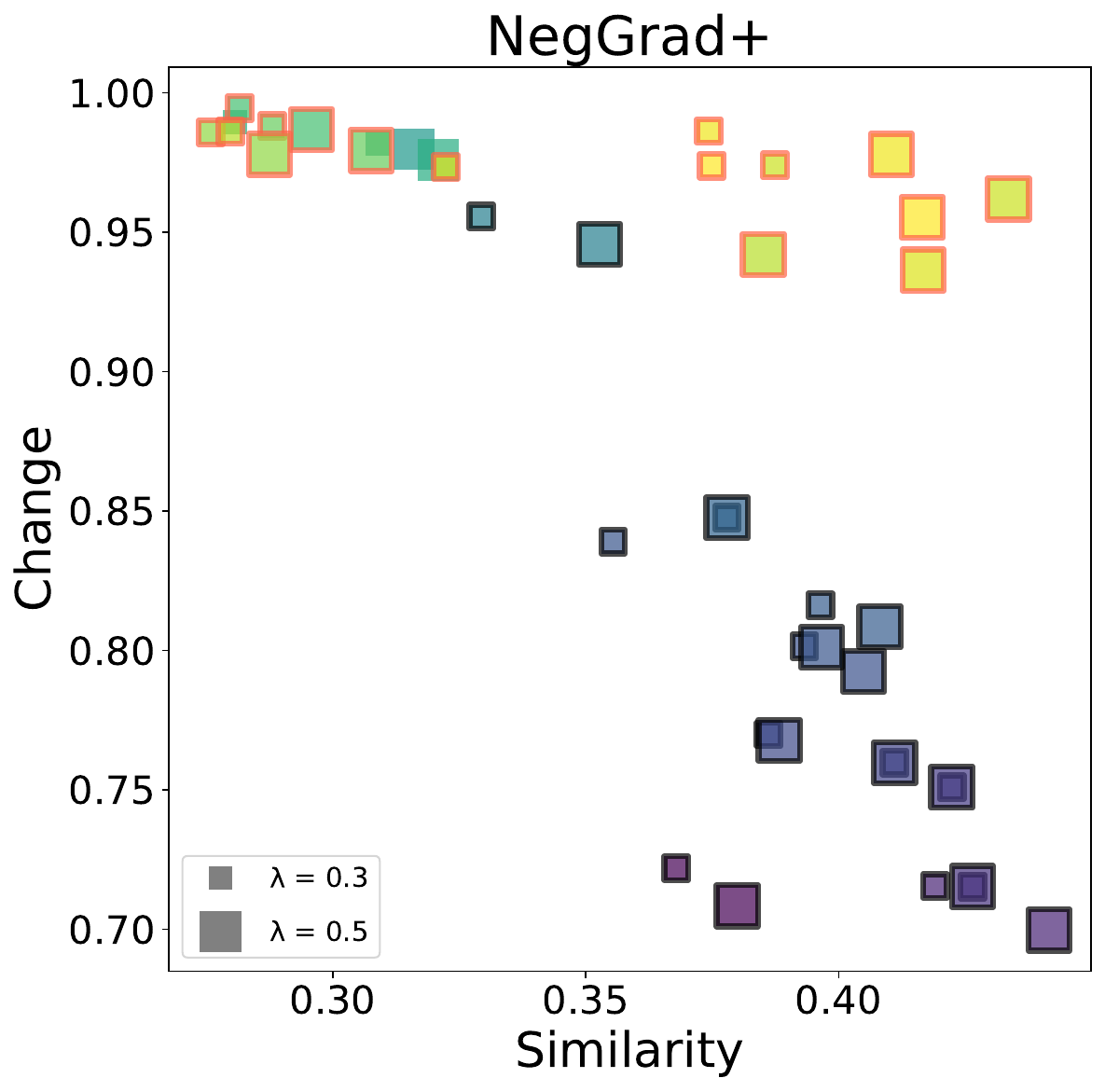}
    \includegraphics[height=0.185\textwidth]{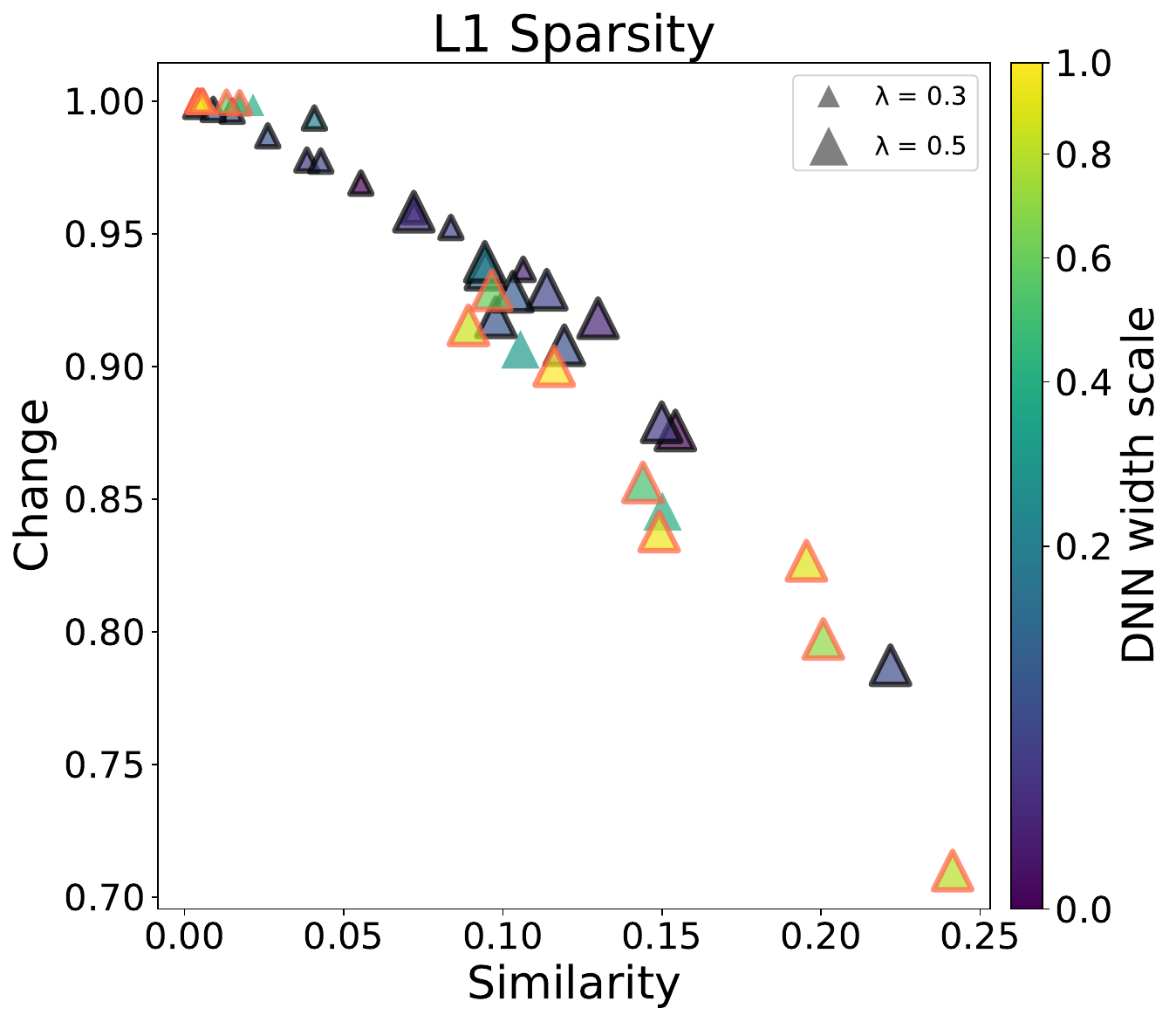}
    \caption{Decision-region similarity and change scores for \textbf{bias removal} unlearning. ResNet-18 on Tiny ImageNet (500 unlearned examples, $\delta=10$). Ideal similarity-change (local changes of decision regions) is at the \textit{top-right corner} of each diagram. See larger versions of these diagrams in Fig.~\ref{fig:similarity_change_bias_resnet18_tinet_500unlearned}.}
    \label{fig:similarity_change_bias_resnet18_tinet_500unlearned - main}
\end{figure*}

\subsection{Empirical Evaluations}
Given an unlearning method, a goal, a DNN architecture, and classification task, we evaluate the decision region similarity and change scores for various parameterization levels. This will demonstrate the parameterization level effect on the unlearning modification of decision regions. Additional details are provided in Appendix \ref{appendix:subsec:Decision Region Sampling and Similarity Calculation}.

Figs.~\ref{fig:similarity_change_bias_resnet18_tinet_500unlearned - main}, \ref{fig:similarity_change_privacy_resnet50_tinet_1000unlearned}-\ref{fig:similarity_change_bias_fcnet3layer_cifar10_200unlearned} show the similarity and complementary change scores across different parameterization levels and different unlearning methods. 
Ideally, for a reasonable choice of $\delta$: \textbf{For the privacy goal}, the similarity-change performance is better if it is closer to the similarity-change marker of the retrained model at the same parameterization level. \textbf{For the bias removal goal}, both the similarity and change scores should be high (i.e., the top-right corner of the diagram), implying local changes around the unlearned examples.

We observe that overparameterized models usually have better similarity-change values (markers at better coordinates) than underparameterized models. As in the performance analysis, for bias removal this requires unlearning that uses the forget set (i.e., not the L1 sparsity method). 
These findings align with the trends in our performance analysis, showing that \textbf{unlearning usually benefits from overparameterization}. Here, \textbf{the decision regions' similarity and change scores further elucidate this behavior, showing that overparameterization facilitates delicate unlearning that can locally modify the necessary decision regions while preserving most of the model functionality.}

\section{Conclusions}
\label{sec:discussion_and_conclusions}

In this work, we examined the parameterization level effect on machine unlearning performance. 
Our results show the benefits of overparameterization in terms of performance-based unlearning evaluations and effects on classification decision regions. 
Our analysis elucidates the interaction between the parameterization level, unlearning goal (unlearned data privacy or bias removal), and unlearning performance in a way that contributes to the interpretability and usage of existing unlearning methods. We believe that our insights can help practitioners to select unlearning methods best suited to their learning setting. Moreover, our new observations may inspire new unlearning methods that consider the parameterization level and unlearning goal in their core definition instead of as an external generic validation process like we examined in this work.

\bibliographystyle{IEEEtran}
\bibliography{references}

%%%%%%%%%%%%%%%%%%%%%%%%%%%%%%%%%%%%%%%%%%%%%%%%%%%%%%%%%%%%

\appendix
\section*{Appendices}

\section{Experiments Compute Resources}
\label{appendix:sec:compute_resources}

All model training and unlearning was performed using an array of NVIDIA RTX 3090 and RTX 4090 GPUs in an internal GPU cluster. 
%We used a storage space of 1.5TB for all the experiments. 

We conducted experiments across \textbf{5 different settings that differently combine model architecture, dataset and forget set size}:
\begin{itemize}
    \item ResNet-34, Tiny ImageNet, 1000 unlearning examples from 20 classes
    \item ResNet-18, Tiny ImageNet, 500 unlearning examples from 10 classes
    \item ResNet-18, CIFAR-10, 200 unlearning examples from 2 classes
    \item ResNet-18, CIFAR-10, 600 unlearning examples from 3 classes
    \item 3-layer fully connected network, CIFAR-10, 200 unlearning examples from 2 classes.
\end{itemize}
For each setting, we evaluated \textbf{21 distinct model widths for each ResNet-18 or ResNet-34 setting}, \textbf{13 distinct model widths for the fully-connected network setting}, with \textbf{3 random seeds} per configuration. For unlearning, we applied five different methods: \textbf{SCRUB} \cite{unbounded_MUL}, \textbf{NegGrad+} \cite{unbounded_MUL}, \textbf{L1-sparsity} \cite{l1_sparsity}, \textbf{SalUn} \cite{salun}, \textbf{RL} \cite{eternal_sunshine}. 
To tune the hyperparameters for each model's unlearning, we performed a grid search over:
\begin{itemize}
    \item \textbf{100 hyperparameter combinations} for SCRUB on ResNet-18 or ResNet-34; \textbf{450 hyperparameter combinations} for SCRUB on 3-layer FC network.
    \item \textbf{25 hyperparameter combinations} for NegGrad+ on ResNet-18 or ResNet-34; \textbf{42 hyperparameter combinations} for NegGrad+ on 3-layer FC network.
    \item \textbf{25 hyperparameter combinations} for L1-sparsity.
    \item \textbf{18 hyperparameter combinations} for SalUn.
    \item \textbf{5 hyperparameter combinations} for RL.
\end{itemize}

This resulted in a total of 
\[
\left({(100 + 25 + 25 + 18 + 5) \times 21 \times 4 + (450 + 42 + 25 + 18 + 5) \times 13}\right) \times 3  = \textbf{{64,656}}
\] 
individual unlearning runs.

In addition to unlearning, in our experiments we trained \textbf{519 models from scratch}, among them $(21\times 3 + 13)\times 3 = 228$ are the trained models before unlearning and the other $(21\times 4 + 13)\times 3 = 291$ are models retrained from scratch without the forget data.

\section{Software Packages}
\label{appendix:subsec:Assets and Software Packages}

We utilized several publicly available datasets and software tools throughout our experiments.

\textbf{Datasets.} We used the following publicly available datasets for our image classification and unlearning experiments:

\begin{itemize}
    \item \textbf{CIFAR-10}: Available at \url{https://www.cs.toronto.edu/~kriz/cifar.html}. This dataset includes 10 object categories with 50{,}000 training images (5{,}000 per class) and 10{,}000 test images. All images are RGB and have a resolution of $32 \times 32 \times 3$ pixels.
    
    \item \textbf{Tiny ImageNet}: A subset of the full ImageNet dataset containing 200 object categories, with 500 training images and 50 validation images per class, totaling 100{,}000 training samples. We used a resized version where all images were scaled to $32 \times 32 \times 3$ to align with our model architectures.

\end{itemize}

\textbf{Software and Libraries.} We made use of several open-source packages and codebases during the course of our experiments:

\begin{itemize}
    \item \textbf{PyTorch}: All model training and GPU-accelerated computations were carried out using PyTorch, a widely adopted deep learning framework. PyTorch is available under the Apache Contributor License Agreement (CLA). More details can be found at \url{https://pytorch.org/}.
    
    \item \textbf{dbViz Toolkit}: For our decision region similarity analysis, we employed the techniques and tools introduced by Somepalli et al. \cite{somepalli_can_nn_learn_twice}. The official implementation is publicly available at \url{https://github.com/somepago/dbViz}.
    
    \item \textbf{SCRUB Codebase}: We used the code repository from \url{https://github.com/meghdadk/SCRUB} to implement the SCRUB \cite{unbounded_MUL} and NegGrad+ unlearning methods, as well as for portions of the base model training procedures.
    
    \item \textbf{L1-sparsity Unlearning}: For experiments involving the L1-sparsity method \cite{l1_sparsity}, we used code from \url{https://github.com/OPTML-Group/Unlearn-Sparse}.

    \item \textbf{SalUn and RL Unlearning}: For the SalUn \cite{salun} and RL \cite{eternal_sunshine} unlearning methods, we used the implementation provided at:
    
    \url{https://github.com/OPTML-Group/Unlearn-Saliency}.

\end{itemize}

\section{Additional Details on the Examined Unlearning Methods}
\label{appendix:sec:Additional Details on Unlearning Methods}

In this appendix, we overview the examined unlearning methods and their optimization formulations. The formulations are important for understanding the hyperparameters that participate in the validation-based tuning process, as described in Section \ref{subsec:Validation-Based Hyperparameter Tuning} and Appendix \ref{grid search details}.

\begin{itemize}  
    \item \textbf{SCRUB Unlearning} \cite{unbounded_MUL}: SCRUB is a data-driven unlearning method that uses a teacher-student model approach. The original model (the teacher) remains unchanged, while a new model (the student) is trained to imitate the teacher on the data that should be retained and diverge from the teacher on the data that should be unlearned. This divergence is achieved by maximizing the Kullback-Leibler (KL) divergence between the teacher and student predictions for the examples to unlearn, appearing as the third sum in the following optimization:
    \begin{equation}
    \label{eq:SCRUB_optimization_formula}
    \begin{split}
    \vec{w}_{\rm u} = \argmin_{\vec{w}} \Bigg\{ 
        &\frac{\alpha}{N_\mathrm{r}} \sum_{(\vec{x}_\mathrm{r}, \cdot) \in \mathcal{D}_\mathrm{r}} 
        d_{\text{KL}}(\vec{x}_\mathrm{r}; \vec{w_0}, \vec{w}) + \frac{\gamma}{N_\mathrm{r}} \sum_{(\vec{x}_\mathrm{r}, y_\mathrm{r}) \in \mathcal{D}_\mathrm{r}} 
        l_{\rm CE}(f(\vec{x}_\mathrm{r}; \vec{w}), y_\mathrm{r}) \\
        &- \frac{1}{N_\mathrm{f}} \sum_{(\vec{x}_\mathrm{f}, \cdot) \in \mathcal{D}_\mathrm{f}} 
        d_{\text{KL}}(\vec{x}_\mathrm{f}; \vec{w_0}, \vec{w})
    \Bigg\}
    \end{split}
    \end{equation}
    which is solved by iterative optimization that is initialized to the original model \(\vec{w}_0\). Here \(l_{\rm CE}(\cdot,\cdot)\) represents the cross-entropy loss; \(d_{\text{KL}}(\vec{x};\vec{w_0}, \vec{w})\) measures the KL-divergence between the softmax probabilities of the student $\vec{w}$ and teacher $\vec{w}_0$ for an input $\vec{x}$; and \(\alpha\) and \(\gamma\) are scalar hyperparameters to determine the importance of the optimization objective terms (including to determine the importance of the retain and forget sets).

    \item \textbf{Negative Gradient+ (NegGrad+)} \cite{unbounded_MUL}: Similar to fine-tuning, NegGrad+ starts from the original model and fine-tunes it both on data from the retain and forget sets, but negates the gradient for the latter. The unlearning optimization is 
    \begin{equation}
    \label{eq:NegGrad_optimization_equation}
    \begin{split}
    \vec{w}_{\rm u} = \argmin_{\vec{w}} \Bigg\{
        &\frac{\alpha}{N_{\mathrm{r}}} \sum_{(\vec{x}_\mathrm{r}, y_\mathrm{r}) \in \mathcal{D}_\mathrm{r}}
        l_{\rm CE}(f(\vec{x}_\mathrm{r}; \vec{w}), y_\mathrm{r}) - \frac{(1 - \alpha)}{N_{\mathrm{f}}} \sum_{(\vec{x}_\mathrm{f}, y_\mathrm{f}) \in \mathcal{D}_\mathrm{f}}
        l_{\rm CE}(f(\vec{x}_\mathrm{f}; \vec{w}), y_\mathrm{f})
    \Bigg\}
    \end{split}
    \end{equation}  
    which is solved by iterative optimization, initialized to the original model $\vec{w}_0$. The hyperparameter \(\alpha\) balances the contributions of the retain set and the forget set.

    \item \textbf{L1 Sparsity Unlearning} \cite{l1_sparsity}:
    This method employs model pruning for machine unlearning, implemented by solving an optimization problem that modifies the original parameters \(\vec{w}_0\) using \textit{only the retained training data} \(\mathcal{D}_\mathrm{r}\), and an \(\ell_1\)-norm penalty that sparsifies the unlearned model parameters:    
    \begin{align}
    \label{eq:L1-sparsity unlearning - optimization}
    \vec{w}_{\rm u}
    = \argmin_{\vec{w}} 
        \frac{1}{N_{\mathrm{r}}}
        \sum_{(\vec{x}_\mathrm{r}, y_\mathrm{r}) \in \mathcal{D}_\mathrm{r}}
        l_{\rm CE}(f(\vec{x}_\mathrm{r}; \vec{w}), y_\mathrm{r})
        \;+\;
        \gamma \|\vec{w}\|_1
    \end{align}
    This optimization is solved by iterative optimization, initialized to the original model $\vec{w}_0$. The regularization parameter \(\gamma\) controls the level of the sparsity penalty, targeting the reduction of parameter values that were previously influenced by the forget set \(\mathcal{D}_\mathrm{f}\). This pruning reduces the model dependency on the forget set by eliminating parameters that are less essential for the retained data.

    \item \textbf{Saliency-based Unlearning (SalUn)} \cite{salun}: SalUn integrates weight saliency with random labeling to selectively fine-tune only the salient parameters. A saliency mask  is computed by thresholding the gradient of the forgetting loss with respect to the original model parameters $\vec{w}_0$; the threshold hyperparameter is denoted as $\tau$. 
    The saliency mask determines which parameters are updated during unlearning. Parameters with saliency below the chosen threshold are frozen, while those above threshold are updated. Lower threshold values allow more parameters to change, whereas higher threshold values restrict updates to only the most salient parameters.    
    Then, the masked version of the model $\vec{w}_0$ is fine-tuned on a merged dataset consisting of the retain set $\mathcal{D}_\mathrm{r}$ and the randomly-labeled forget set $\widetilde{\mathcal{D}}_\mathrm{f}$ where forget input examples from $\mathcal{D}_\mathrm{f}$ are paired with uniformly-drawn random labels:
    \begin{equation}
    \label{eq:SalUn_optimization}
    \begin{split}
        \vec{w}_{\rm u}
        = \argmin_{\vec{w}}
        \sum_{(\vec{x}, y) \in \mathcal{D}_{\mathrm{r}} \cup \widetilde{\mathcal{D}}_{\mathrm{f}}}
        l_{\rm CE}\!\left(f(\vec{x}; \vec{w}), y\right)
    \end{split}
    \end{equation}
    which is solved by iterative optimization initialized to $\vec{w}_0$, allowing to update only the salient parameters; non-salient parameters are fixed to their corresponding values in $\vec{w}_0$. 

    \item \textbf{Random Labeling (RL)} \cite{eternal_sunshine}:
    The RL method follows the same general approach as SalUn, but \emph{without} using any saliency-based masking to restrict the parameter updates. Specifically, the model is fine-tuned on a merged dataset consisting of the retain set $\mathcal{D}_\mathrm{r}$ and the randomly-labeled forget set $\widetilde{\mathcal{D}}_\mathrm{f}$ where forget input examples from $\mathcal{D}_\mathrm{f}$ are paired with uniformly-drawn random labels. Unlike SalUn, all of the model parameters are updated during this fine-tuning stage using same objective like in (\ref{eq:SalUn_optimization}).
    The random labeling disrupts the model’s learned association with the forget data, while the retain loss preserves performance on $\mathcal{D}_\mathrm{r}$.

\end{itemize}

\section{Additional Experimental Details}
\label{appendix:Additional Experiment Details}

\subsection{Model Architecture Details}
\label{appendix:subsec:Model Architecture details}

In this research, we use the term \textbf{width scale} (\(s\)) to describe the parameterization level of a model. The width scale adjusts the number of channels in each layer of the architecture relative to its base configuration. Specifically, for a given base number of channels \(c\), the scaled number of channels is defined as \(\lfloor c \times s \rfloor\). We experiment with width scales in the range \([0.02, 1]\), where \(s = 1\) corresponds to the original architecture.

We evaluate three architecture forms: ResNet-18, ResNet-34, and a 3-layer fully-connected network. Following approaches used in prior work \cite{somepalli_can_nn_learn_twice,nakkiran2019deep}, we vary the parameterization level by adjusting the width scale. Below, we provide a detailed description of each architecture. 
Importantly, note that \cite{somepalli_can_nn_learn_twice,nakkiran2019deep} denoted the ResNet-18 width by a width parameter whose value should be 64 for forming the standard ResNet-18; in contrast, we use a width scale whose value should be 1 for forming the standard ResNet-18.

\paragraph{ResNet-18} 
We create a family of scaled ResNet-18 models by adjusting their width scale \(s\). The architecture starts with an initial convolutional layer, followed by residual stages with downsampling at the beginning of each stage. ResNet-18 has four stages of preactivation residual blocks, each with two BatchNorm-ReLU-convolution layers and skip connections, with channel widths \([64 \times s, 128 \times s, 256 \times s, 512 \times s]\).

\paragraph{ResNet-34} includes four stages of preactivation basic blocks, where each block consists of two $3 \times 3$ convolutional layers. The network follows a block depth layout of $[3, 4, 6, 3]$ across its four stages, with scaled channel widths of $[64 \times s, 128 \times s, 256 \times s, 512 \times s]$.

The ResNet architectures ends with global average pooling and a fully connected layer. Our implementations follow the design in \cite{unbounded_MUL,somepalli_can_nn_learn_twice}.

\paragraph{3-layer fully connected (FC) network.} This architecture is a three-layer fully connected neural network (MLP) that flattens a multi-channel input and processes it through two ReLU-activated hidden layers before generating final class predictions via a linear output layer. The two hidden layers scale up to 6144 and 3072 units depending on the chosen scaling factor in the range $(0,1]$.

\subsection{Batch Sizes}
We used a batch size of 128 for original model training. For SCRUB and NegGrad+, the retain set was processed with a batch size of 256 examples;  The entire forget set was used as a single batch if it does not include more than 200 examples, otherwise, the forget set batch size is 64. For SalUn and RL, the batch sizes of both retain and forget sets are set to 32. As commonly done in machine unlearning research (for example, \cite{unbounded_MUL}), the training is without data augmentation.

\subsection{Grid Search Setup}
\label{grid search details}

We denote the hyperparameters of an unlearning method $U$ as $\Psi$; the corresponding unlearned model is $f(\cdot; \vec{w}_{\Psi})$ where $\vec{w}_{\Psi}$ is the unlearned parameter vector obtained by applying the unlearning method on the original parameter vector $\vec{w}_0$, i.e., $\vec{w}_{\Psi} \triangleq U(\vec{w}_0; \Psi)$.

Recall the description and formulations in Appendix ~\ref{appendix:sec:Additional Details on Unlearning Methods}. 
The hyperparameter and their grids for each method were as follows:
\begin{itemize}
    \item \textbf{SCRUB (\ref{eq:SCRUB_optimization_formula}):}
    \begin{itemize}
        \item Hyperparams:\newline$\Psi =\nolinebreak \{\alpha, \gamma, \text{msteps}, \text{learning rate}, \text{number of unlearning iterations}\}$.
        \item Grids for ResNet-18 and ResNet-34:
        \begin{itemize}
            \item Learning rate (\(lr\)): \(\{0.0005, 0.001,0.005,0.01\}\)
            \item CE loss factor (\(\gamma\)): \(\{0.1,0.5,0.9, 0.95, 0.99\}\)
            \item Distillation loss weight (\(\alpha\)): \(\{0.001, 0.005, 0.01, 0.1, 0.5\}\)
            \item \text{msteps}=2 (fixed)
        \end{itemize}
        \item Grids for 3-layer FC network:
        \begin{itemize}
            \item Learning rate (\(lr\)): \(\{0.001, 0.005, 0.01, 0.05, 0.1\}\)
            \item CE loss factor (\(\gamma\)): \(\{0.01, 0.05, 0.1, 0.5, 0.9, 0.95\}\)
            \item Distillation loss weight (\(\alpha\)): \(\{0.001, 0.005, 0.01, 0.1, 0.5\}\)
            \item \text{msteps}: \(\{2,3,4\}\)
        \end{itemize}
    \end{itemize}
    \item \textbf{NegGrad+ (\ref{eq:NegGrad_optimization_equation}):}
    \begin{itemize}    
        \item Hyperparams: \newline$\Psi =\nolinebreak \{\alpha, \text{learning rate}, \text{number of unlearning iterations}\}$.
        \item Grids for ResNet-18 and ResNet-34:
        \begin{itemize}
        \item Learning rate (\(lr\)): \(\{0.001, 0.01, 0.05, 0.1, 0.5\}\)
        \item Forget loss weight (\(\alpha\)): \(\{0.1,0.5, 0.9, 0.999,0.9999\}\)
        \end{itemize}
        \item Grids for 3-layer FC network:
        \begin{itemize}
        \item Learning rate (\(lr\)): \(\{0.00005, 0.0001, 0.001, 0.01, 0.05, 0.1\}\)
        \item Forget loss weight (\(\alpha\)): \(\{0.1,0.5, 0.9, 0.95, 0.98, 0.999, 0.9999\}\)
        \end{itemize}
    \end{itemize}
    \item \textbf{L1 Sparsity (\ref{eq:L1-sparsity unlearning - optimization}):}
    \begin{itemize}
    \item Hyperparams: $\Psi = \{\gamma, \text{ratio of unlearning iterations}\}$
    \item Grids:
    \begin{itemize}
        \item $\ell_1\text{-norm penalty factor }(\gamma)$: $\{0.0001, 0.0005, 0.001, 0.005, 0.01\}$
        \item Ratio of unlearning iterations (see description below): \(\{0, 0.1, 0.3, 0.5, 0.8\}\)
    \end{itemize}
    \end{itemize}
    \item \textbf{SalUn (\ref{eq:SalUn_optimization}):}
    \begin{itemize}
    \item Hyperparams: $\Psi = \{\text{learning rate}, \tau\}$ where $\tau$ is the saliency threshold.
    \item Grids:
    \begin{itemize}   
        \item Learning rate (\(lr\)): \(\{0.005, 0.01, 0.1\}\)
        \item Mask threshold: \(\{0.1, 0.3, 0.5, 0.7, 0.9, 1.0\}\)
    \end{itemize}
    \end{itemize}
    \item \textbf{RL (Random Labeling):}
    \begin{itemize}
        \item Hyperparams: $\Psi = \{\text{learning rate}\}$.
        \item Grids:
        \begin{itemize}  
        \item Learning rate (\(lr\)): \(\{0.001, 0.005, 0.01, 0.05, 0.1\}\)
        \end{itemize}
    \end{itemize}
\end{itemize}

For all methods, besides L1 Sparsity, we employ epochwise evaluation during the unlearning process. This allows to select the model from the epoch that achieved the lowest validation score (like a retroactive early stopping of the unlearning process). 

For \textbf{L1 Sparsity}, the ratio of unlearning iterations is implemented following \cite{l1_sparsity} by varying the proportion of epochs during which the \(\ell_1\)-norm penalty is applied. Specifically, training is divided into two phases: an initial unlearning phase where the model is trained using the \(\ell_1\)-norm penalty, and a subsequent fine-tuning phase on the retain set without the penalty, as described in \cite{l1_sparsity}.

\section{Performance-Based Analysis: Additional Results}
\label{appendix: Additional Error-based Experiments}

\begin{figure*}[]
    \centering
    \includegraphics[width=\textwidth]{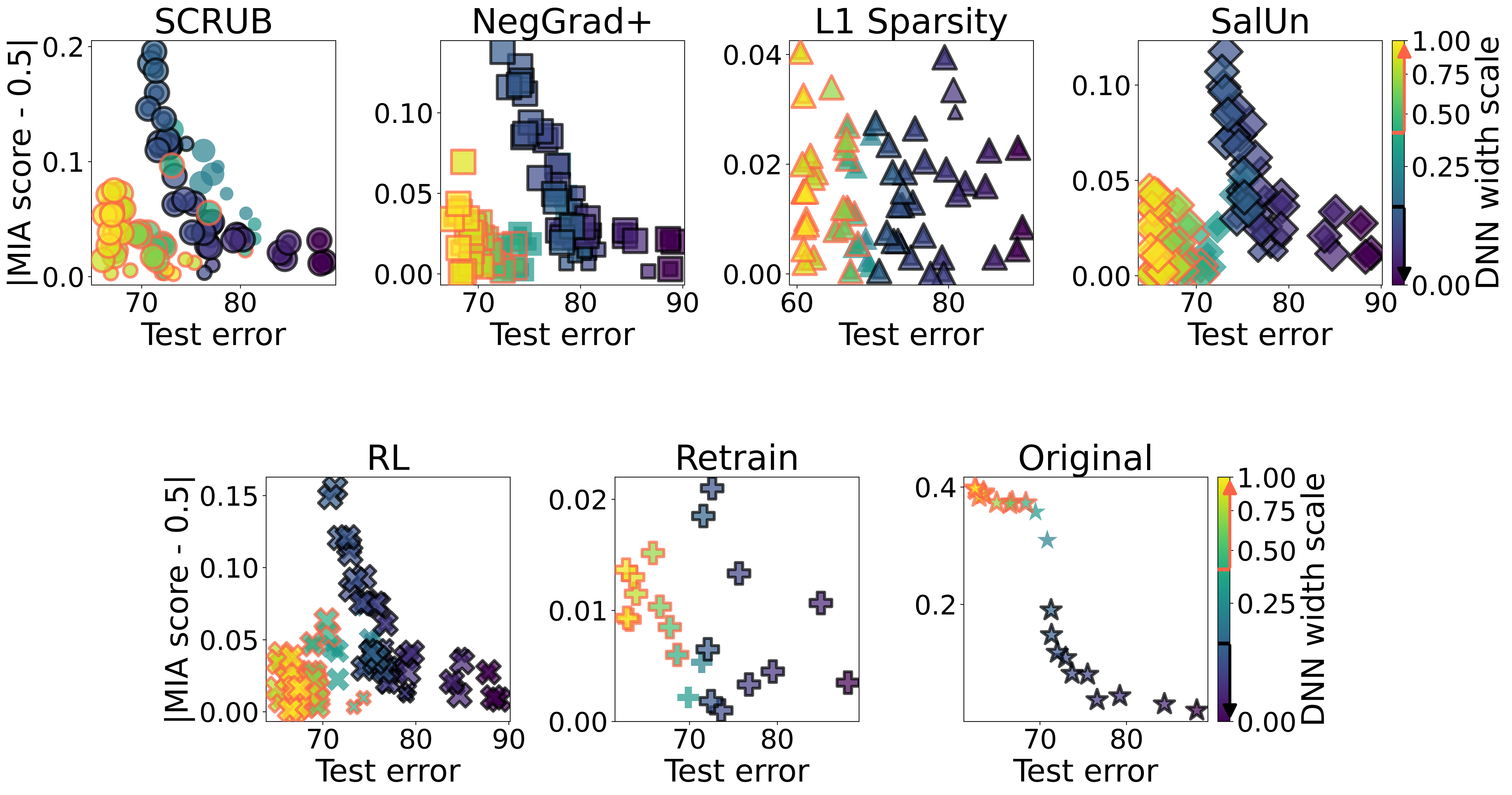}
    \caption{Unlearning for \textbf{privacy} (ResNet-34, Tiny ImageNet, 1000 unlearned examples from 20 out of 200 classes). Ideal privacy-utility tradeoff is at the \textbf{bottom-left corner} of each diagram. Markers' border line color: orange denotes overparameterization, black denotes underparameterization.}
    \label{fig:performance results - tinet resnet50 unlearn 1000 - privacy}
\end{figure*}

\begin{figure*}[]
    \centering
    \includegraphics[width=\textwidth]{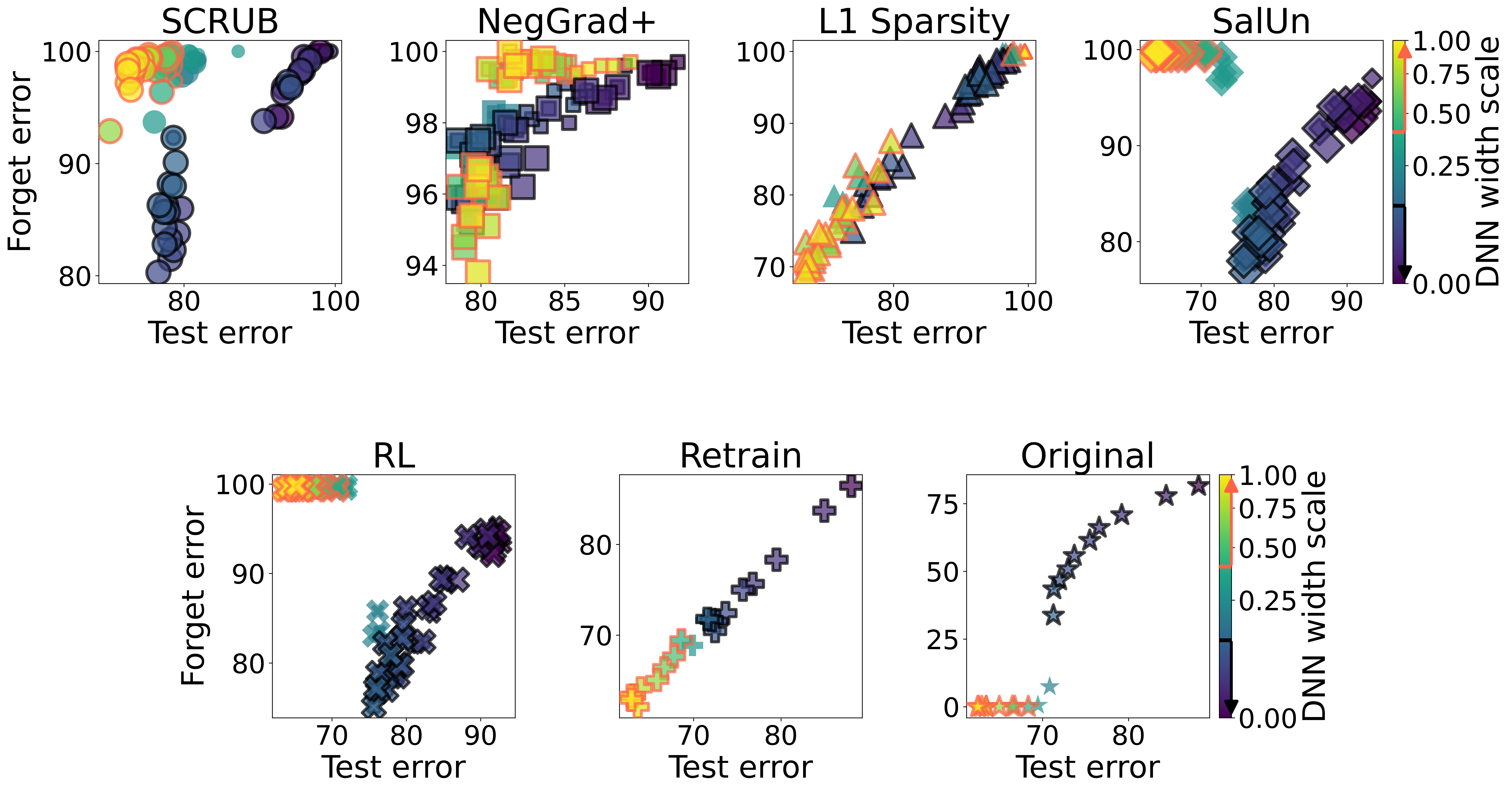}
    \caption{Unlearning for \textbf{bias removal} (ResNet-34, Tiny ImageNet, 1000 unlearned examples from 20 out of 200 classes). Ideal tradeoff of bias removal vs.~utility is at the \textbf{top-left corner} of each diagram. Markers' border line color: orange denotes overparameterization, black denotes underparameterization.}
    \label{fig:performance results - tinet resnet50 unlearn 1000 - bias}
\end{figure*}

\begin{figure*}[]
    \centering
    \includegraphics[width=\textwidth]{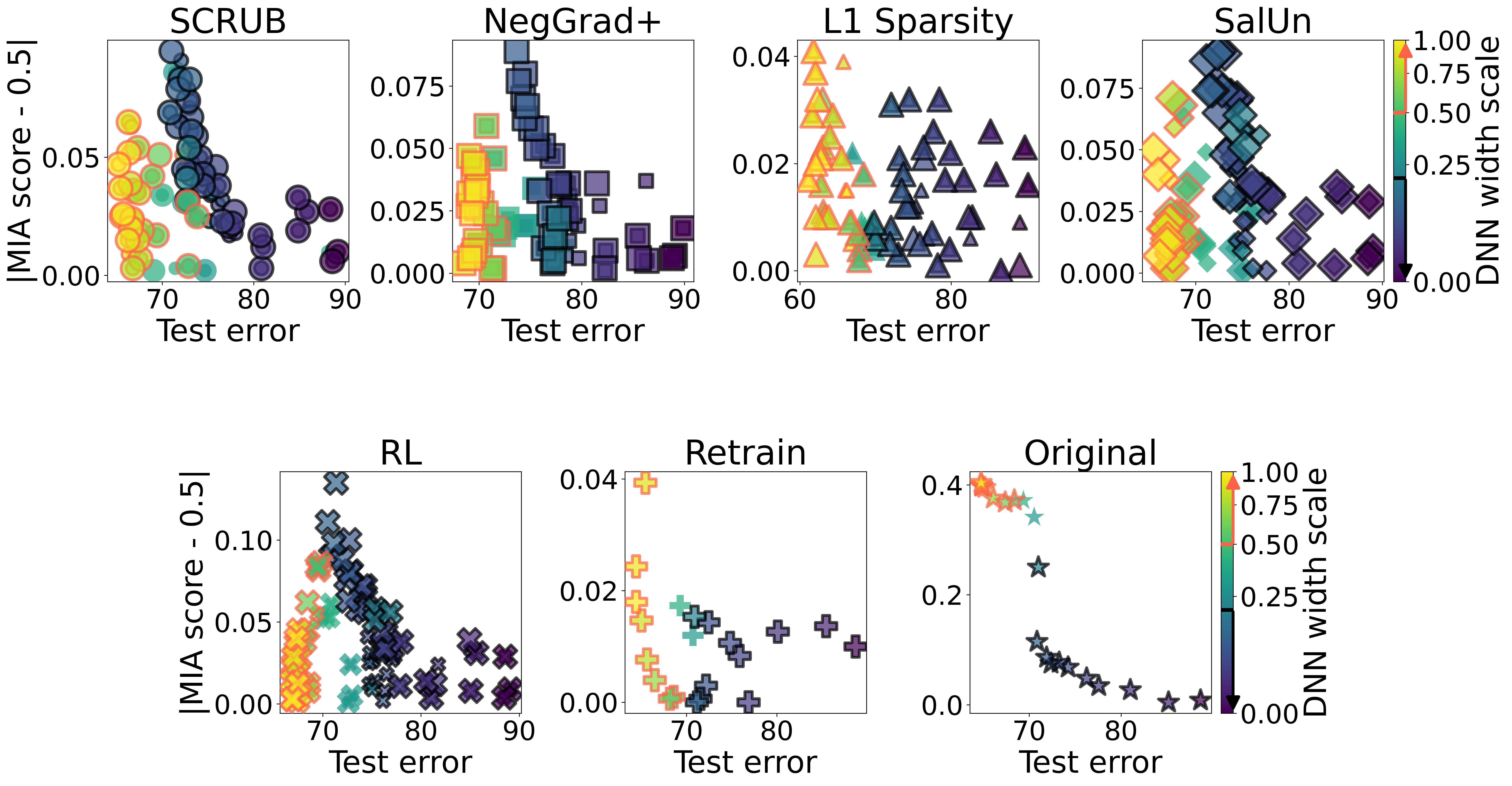}
    \caption{Unlearning for \textbf{privacy} (ResNet-18, Tiny ImageNet, 500 unlearned examples from 10 out of 200 classes). Ideal privacy-utility tradeoff is at the \textbf{bottom-left corner} of each diagram. Markers' border line color: orange denotes overparameterization, black denotes underparameterization.}
    \label{fig:performance results - tinet resnet18 unlearn 500 - privacy}
\end{figure*}

\begin{figure*}[]
    \centering
    \includegraphics[width=\textwidth]{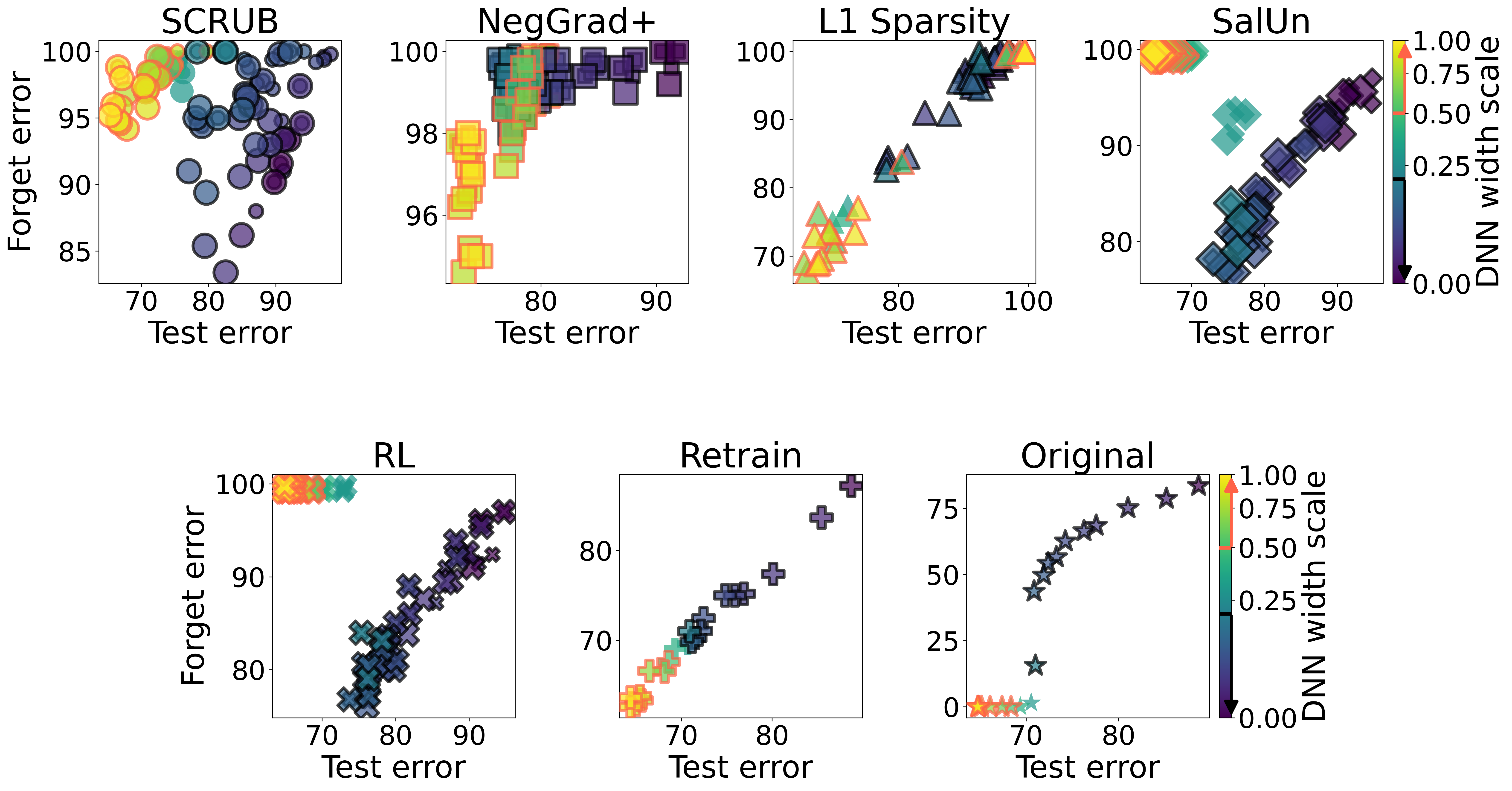}
    \caption{Unlearning for \textbf{bias removal} (ResNet-18, Tiny ImageNet, 500 unlearned examples from 10 out of 200 classes). Ideal tradeoff of bias removal vs.~utility is at the \textbf{top-left corner} of each diagram. Markers' border line color: orange denotes overparameterization, black denotes underparameterization.}
    \label{fig:performance results - tinet resnet18 unlearn 500 - bias}
\end{figure*}

\begin{figure*}[]
    \centering
    \includegraphics[width=\textwidth]{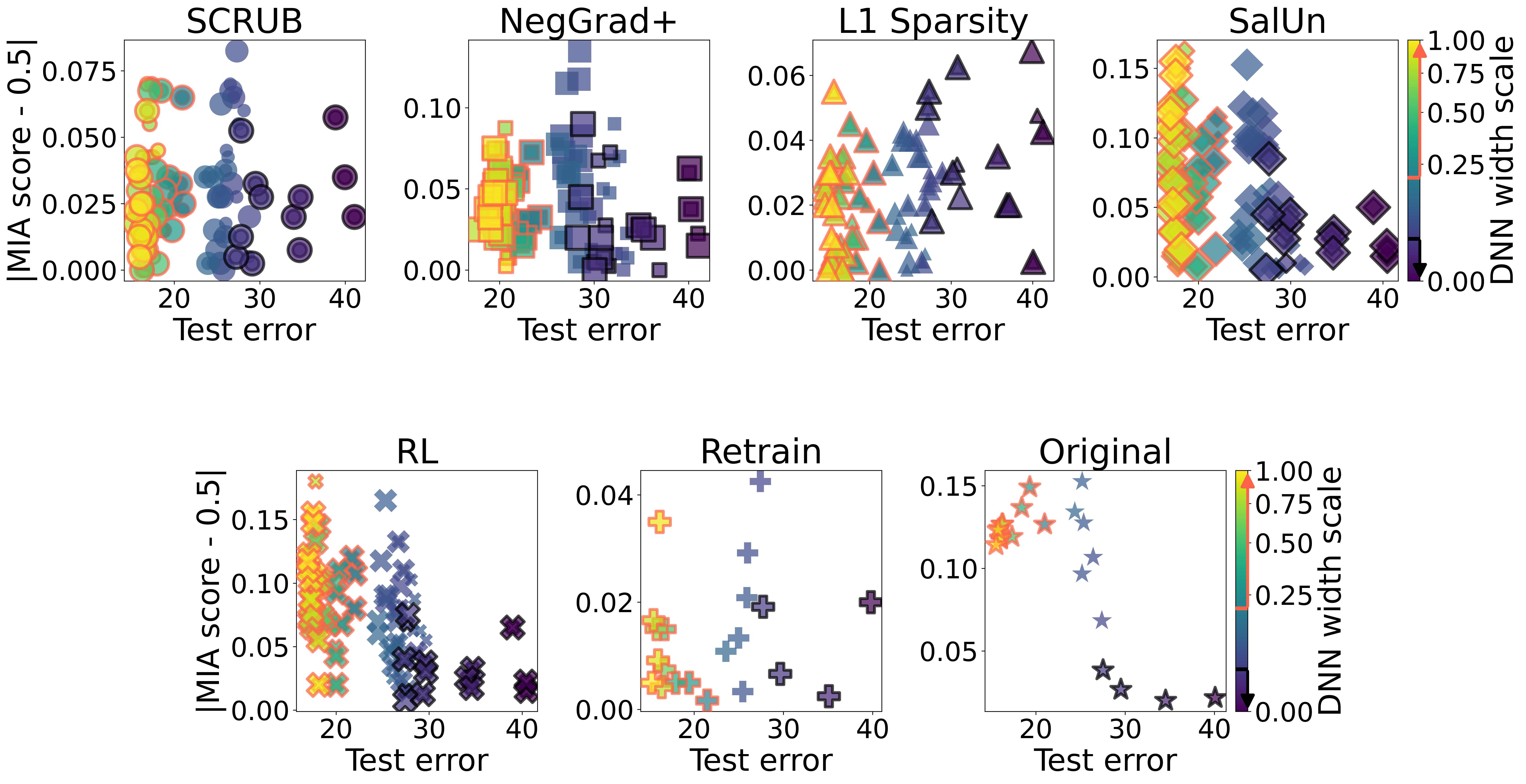}
    \caption{Unlearning for \textbf{privacy} (ResNet-18, CIFAR-10, 200 unlearned examples from 2 out of 10 classes). Ideal privacy-utility tradeoff is at the \textbf{bottom-left corner} of each diagram. Markers' border line color: orange denotes overparameterization, black denotes underparameterization.}
    \label{fig:performance results - cifar10 resnet18 unlearn 200 - privacy}
\end{figure*}

\begin{figure*}[]
    \centering
    \includegraphics[width=\textwidth]{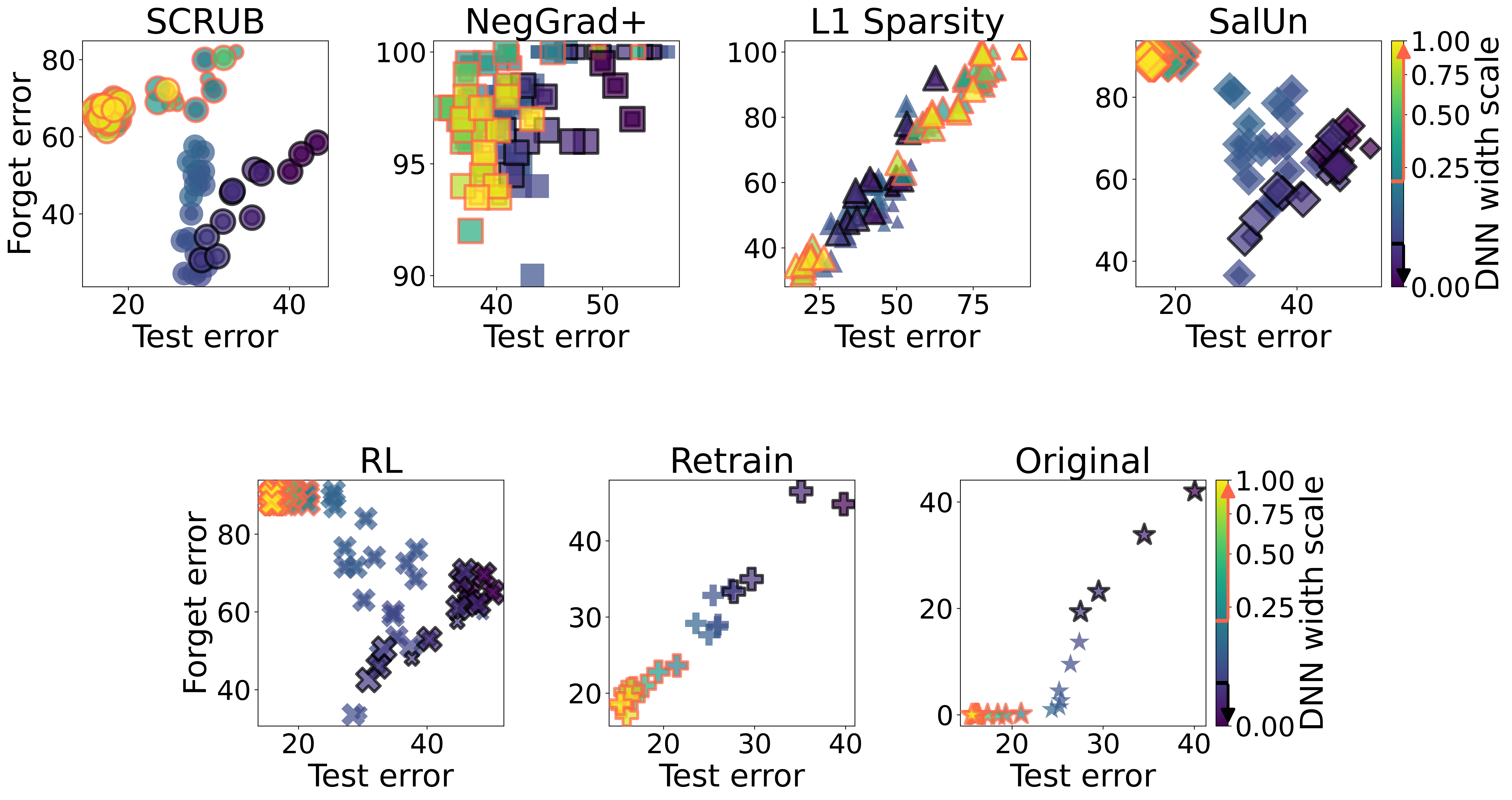}
    \caption{Unlearning for \textbf{bias removal} (ResNet-18, CIFAR-10, 200 unlearned examples from 2 out of 10 classes). Ideal tradeoff of bias removal vs.~utility is at the \textbf{top-left corner} of each diagram. Markers' border line color: orange denotes overparameterization, black denotes underparameterization. }
    \label{fig:performance results - cifar10 resnet18 unlearn 200 - bias}
\end{figure*}

\begin{figure*}[]
    \centering
    \includegraphics[width=\textwidth]{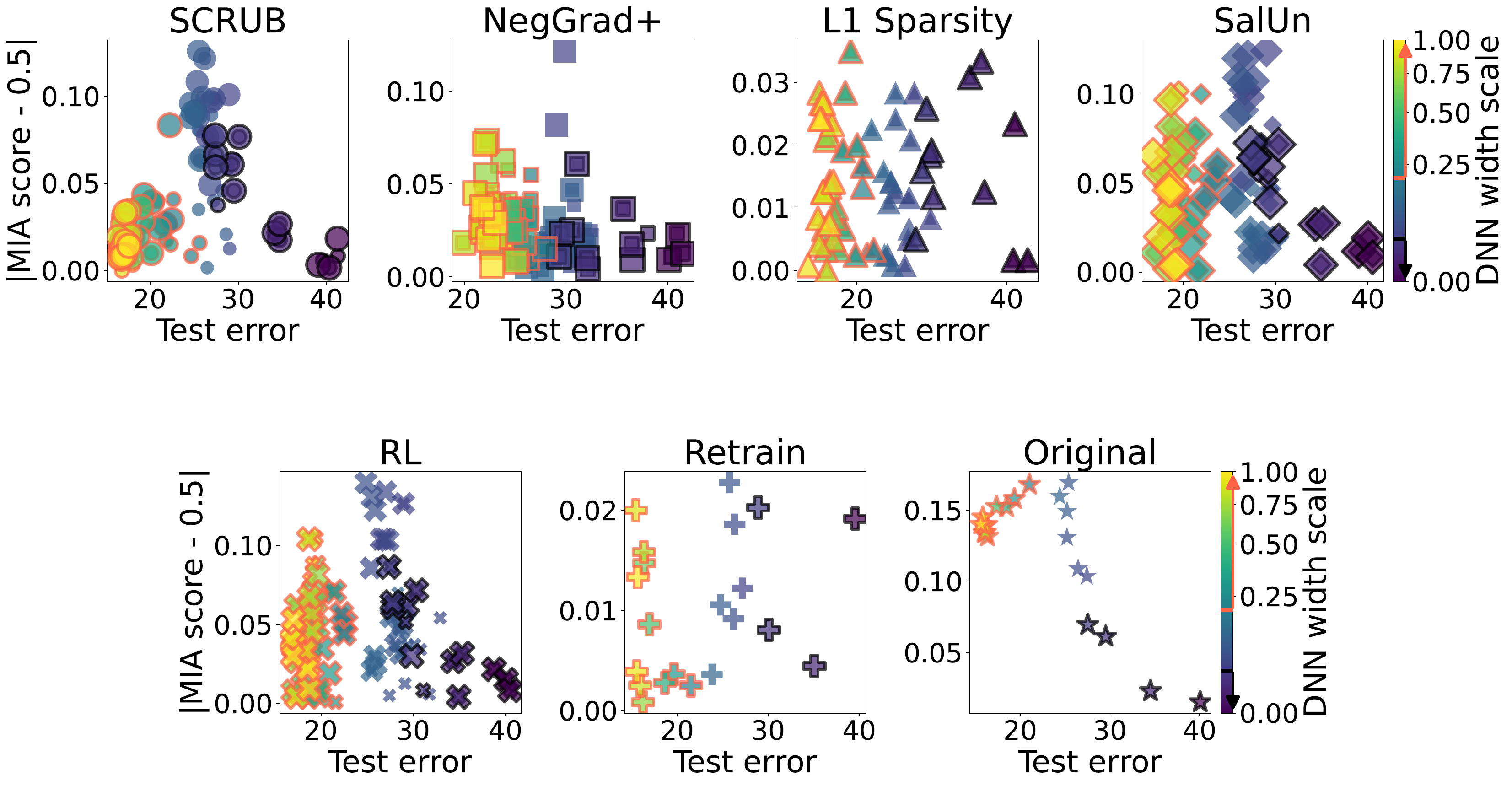}
    \caption{Unlearning for \textbf{privacy} (ResNet-18, CIFAR-10, 600 unlearned examples from 3 out of 10 classes). Ideal privacy-utility tradeoff is at the \textbf{bottom-left corner} of each diagram. Markers' border line color: orange denotes overparameterization, black denotes underparameterization.}
    \label{fig:performance results - cifar10 resnet18 unlearn 600 - privacy}
\end{figure*}

\begin{figure*}[]
    \centering
    \includegraphics[width=\textwidth]{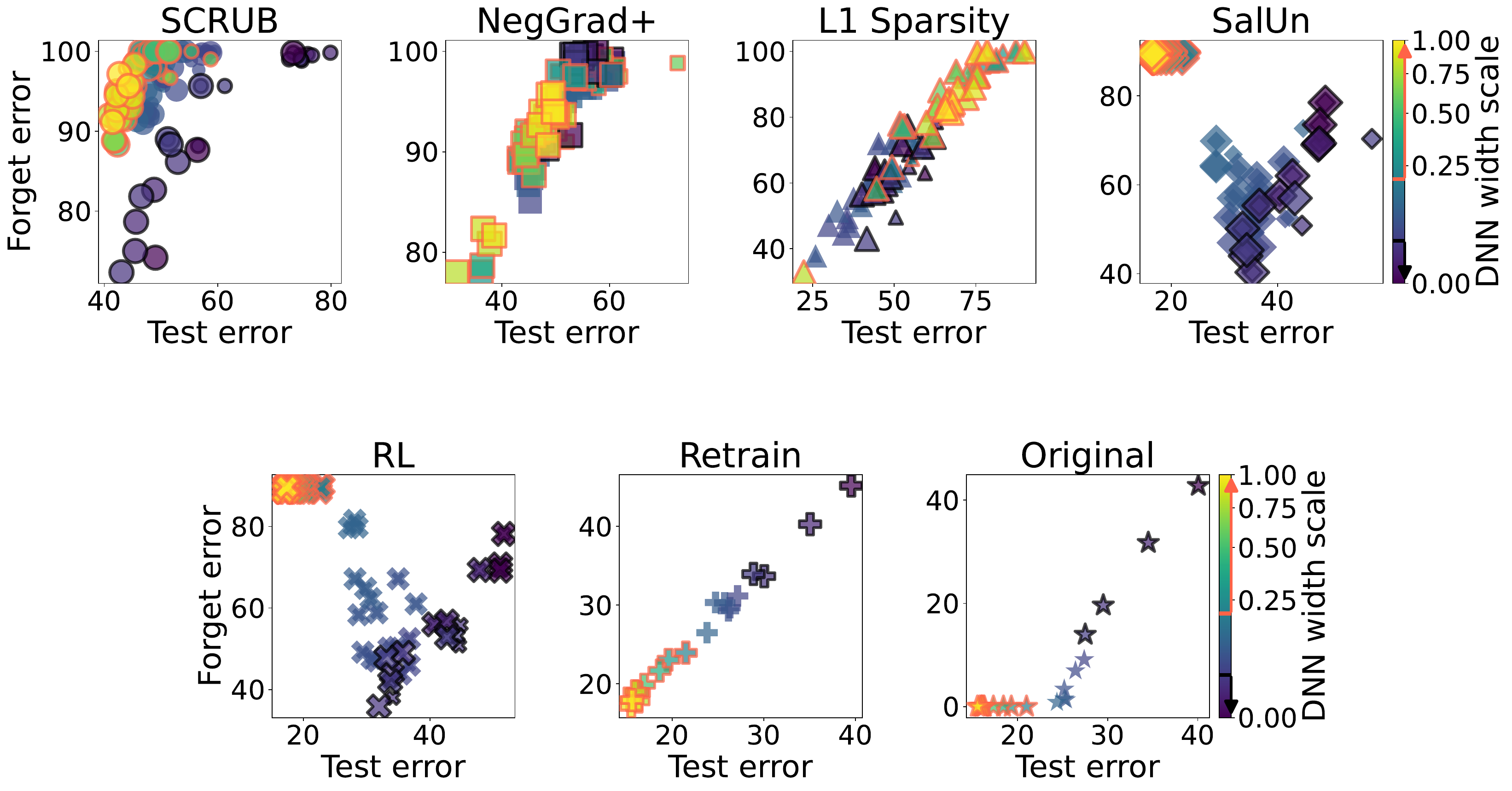}
    \caption{Unlearning for \textbf{bias removal} (ResNet-18, CIFAR-10, 600 unlearned examples from 3 out of 10 classes). Ideal tradeoff of bias removal vs.~utility is at the \textbf{top-left corner} of each diagram. Markers' border line color: orange denotes overparameterization, black denotes underparameterization.}
    \label{fig:performance results - cifar10 resnet18 unlearn 600 - bias}
\end{figure*}

\begin{figure*}[]
    \centering
    \includegraphics[width=\textwidth]{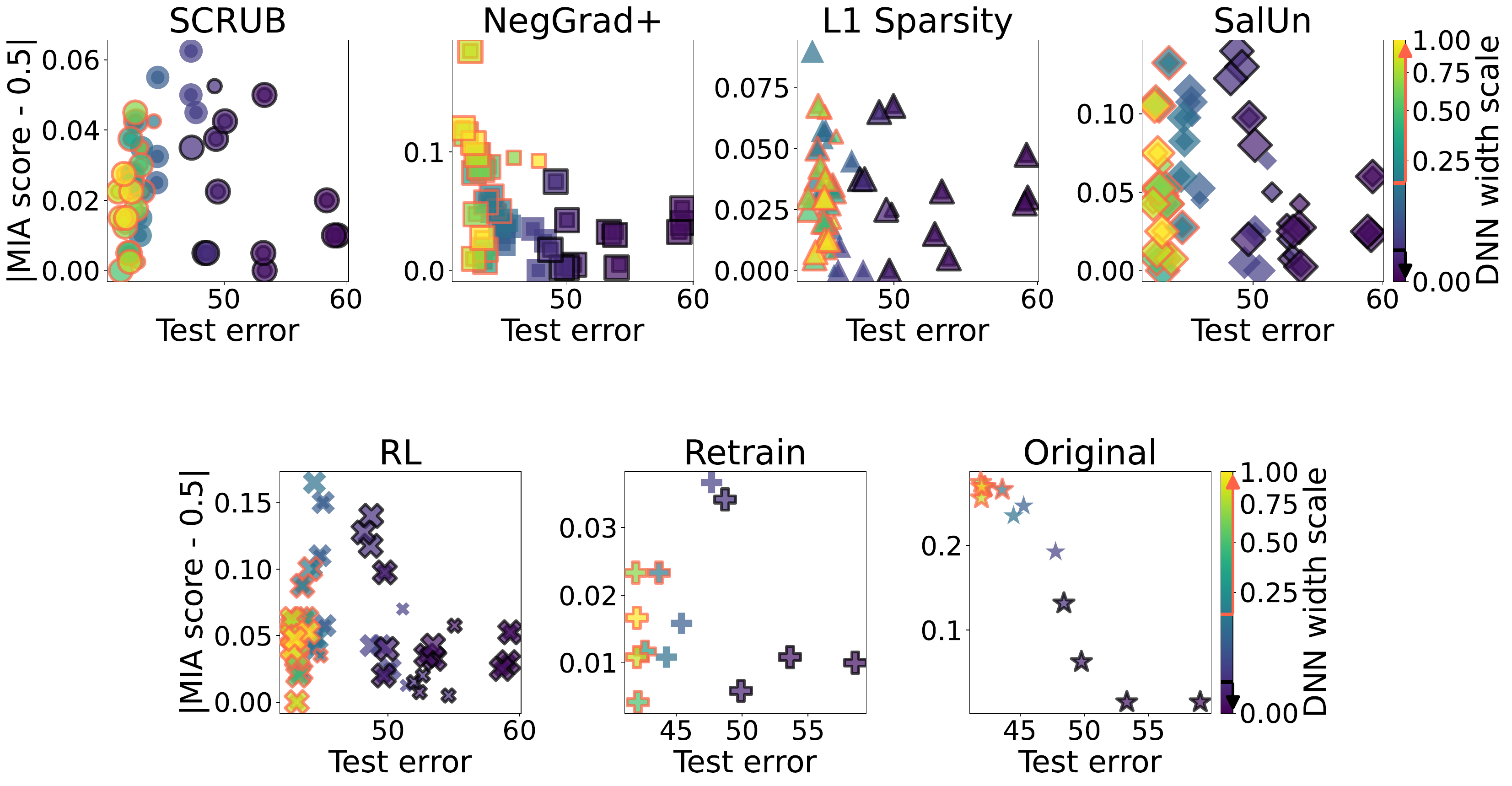}
    \caption{Unlearning for \textbf{privacy} (3-layer FC network, CIFAR-10, 200 unlearned examples from 2 out of 10 classes). Ideal privacy-utility tradeoff is at the \textbf{bottom-left corner} of each diagram. Markers' border line color: orange denotes overparameterization, black denotes underparameterization.}
    \label{fig:performance results - fcnet3layer unlearn 200 - privacy}
\end{figure*}

\begin{figure*}[]
    \centering
    \includegraphics[width=\textwidth]{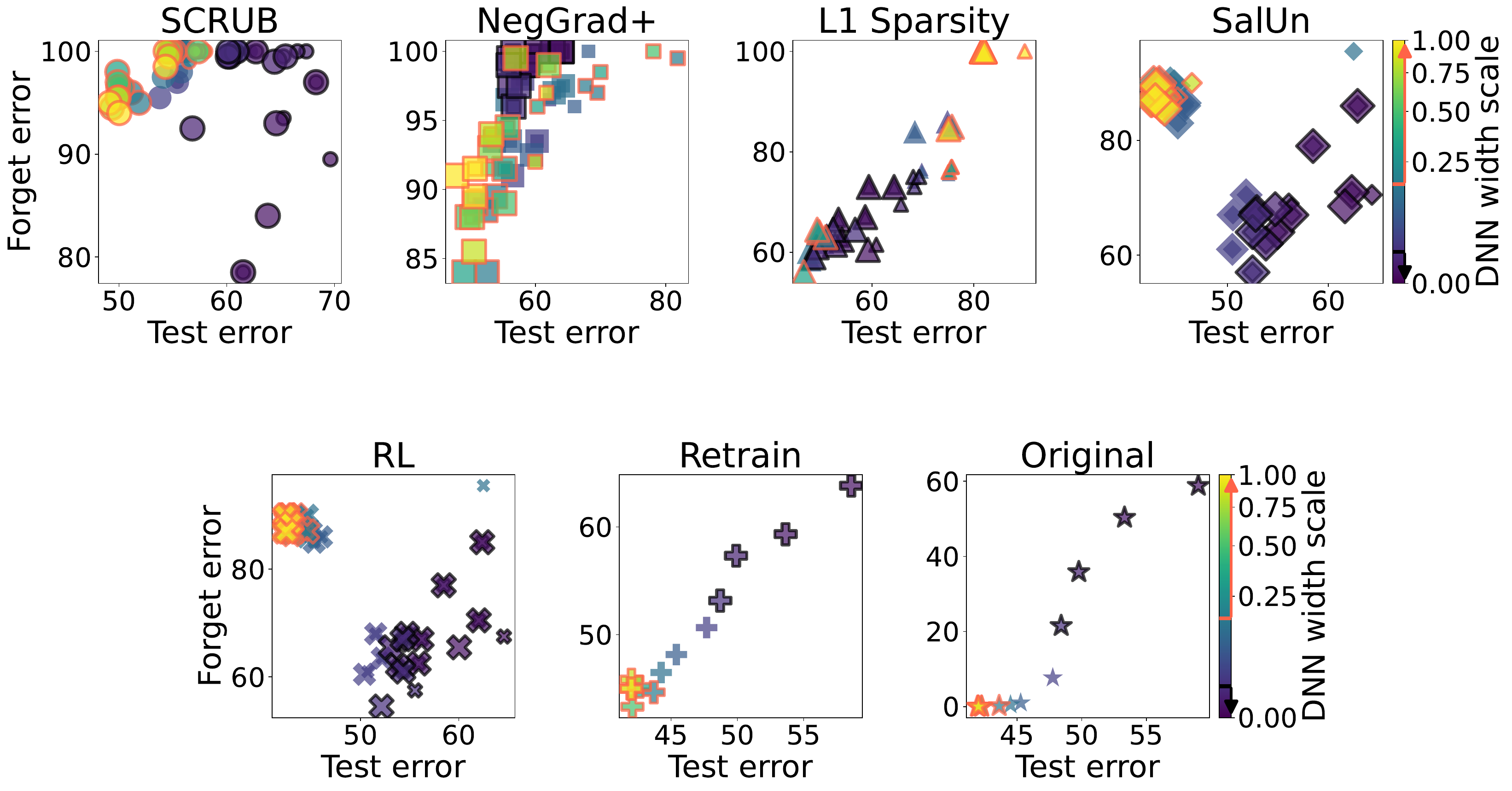}
    \caption{Unlearning for \textbf{bias removal} (3-layer FC network, CIFAR-10, 200 unlearned examples from 2 out of 10 classes). Ideal tradeoff of bias removal vs.~utility is at the \textbf{top-left corner} of each diagram. Markers' border line color: orange denotes overparameterization, black denotes underparameterization.}
    \label{fig:performance results - fcnet3layer unlearn 200 - bias}
\end{figure*}

\begin{figure*}[]
    \centering
    \includegraphics[height=0.22\textheight]{figures/tinet_resnet50_forget1000_absolute_metrics_scatter_compact_nolegend.pdf}
    \includegraphics[height=0.22\textheight]{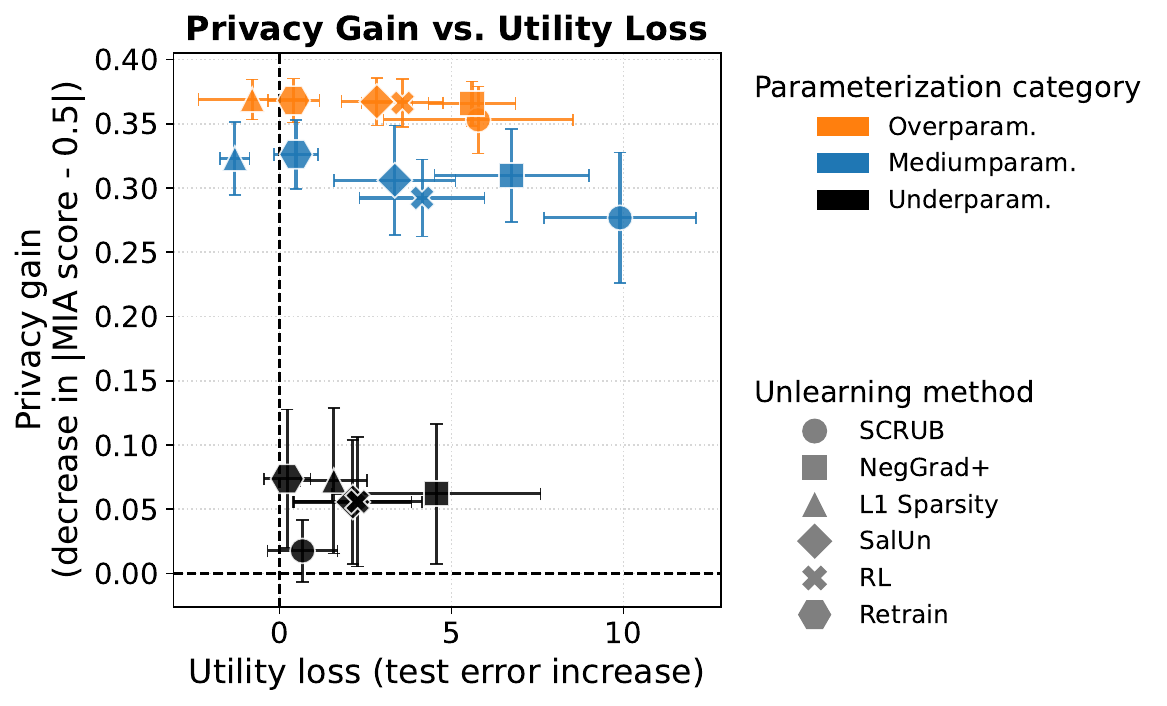}
    \caption{Unlearning for \textbf{privacy} (ResNet-34, Tiny ImageNet, 1000 unlearned examples from 20 out of 200 classes). Ideal tradeoff of the left diagram is at its bottom-left corner. Ideal tradeoff of the right diagram is at its top-left corner.}
    \label{fig:category performance results - tinet resnet50 unlearn 1000 - privacy}
\end{figure*}

\begin{figure*}[]
    \centering
    \includegraphics[height=0.22\textheight]{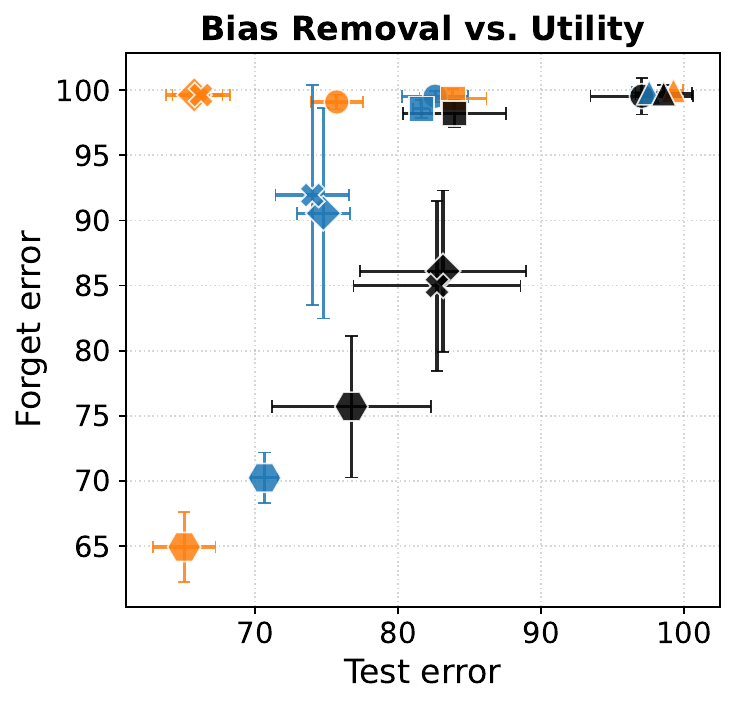}
    \includegraphics[height=0.22\textheight]{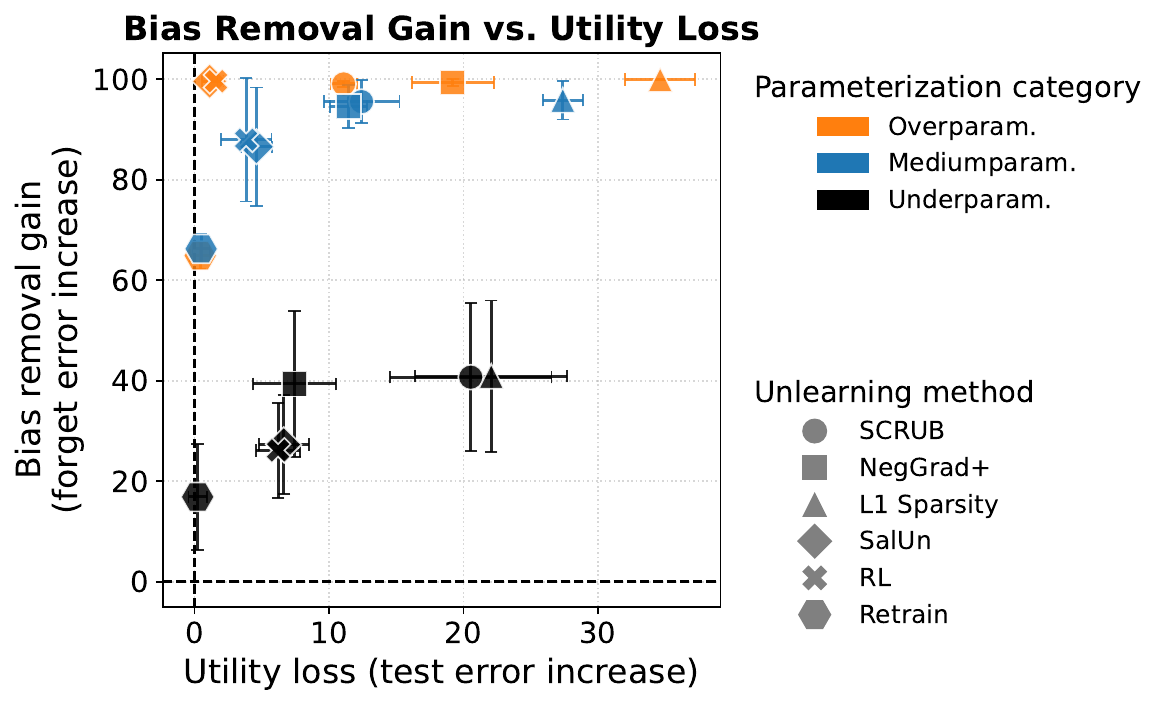}
    \caption{Unlearning for \textbf{bias removal} (ResNet-34, Tiny ImageNet, 1000 unlearned examples from 20 out of 200 classes). Ideal tradeoff of the each diagram is at its top-left corner.}
    \label{fig:category performance results - tinet resnet50 unlearn 1000 - bias}
\end{figure*}

\begin{figure*}[]
    \centering
    \includegraphics[height=0.22\textheight]{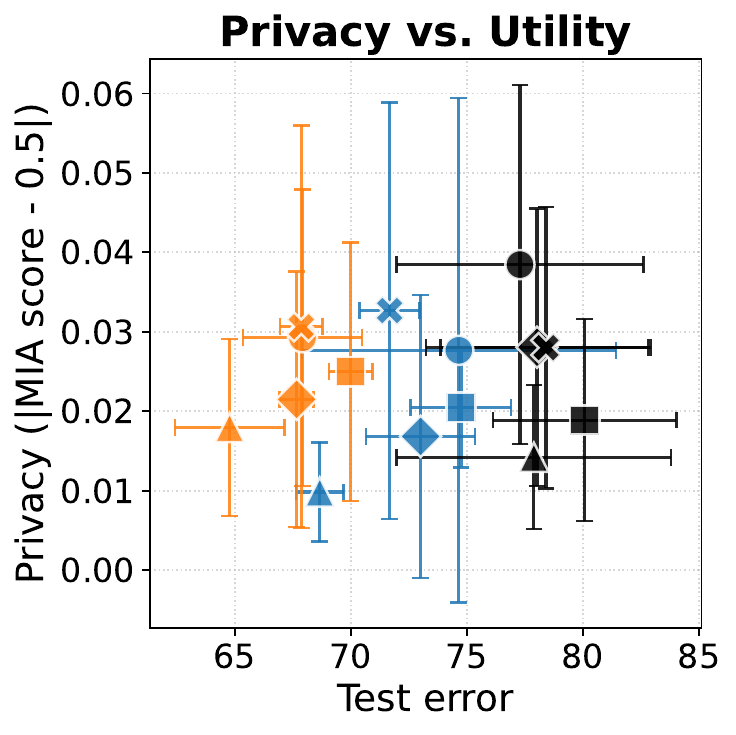}
    \includegraphics[height=0.22\textheight]{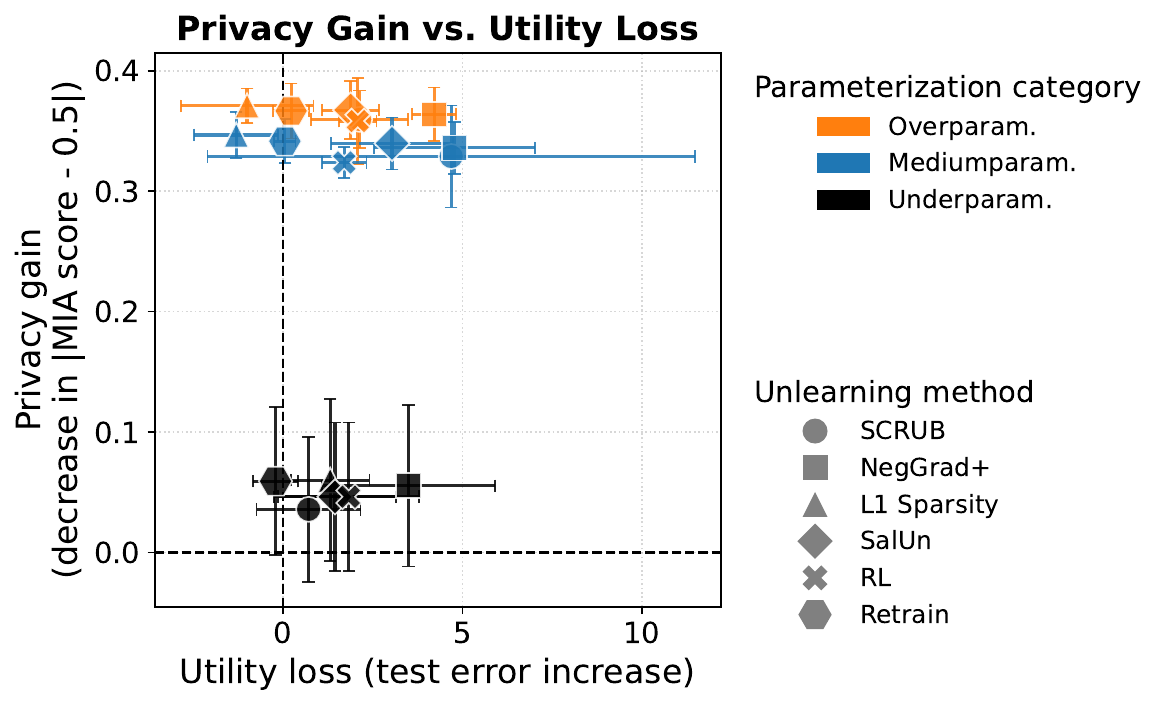}
    \caption{Unlearning for \textbf{privacy} (ResNet-18, Tiny ImageNet, 500 unlearned examples from 10 out of 200 classes). Ideal tradeoff of the left diagram is at its bottom-left corner. Ideal tradeoff of the right diagram is at its top-left corner.}
    \label{fig:category performance results - tinet resnet18 unlearn 500 - privacy}
\end{figure*}

\begin{figure*}[]
    \centering
    \includegraphics[height=0.22\textheight]{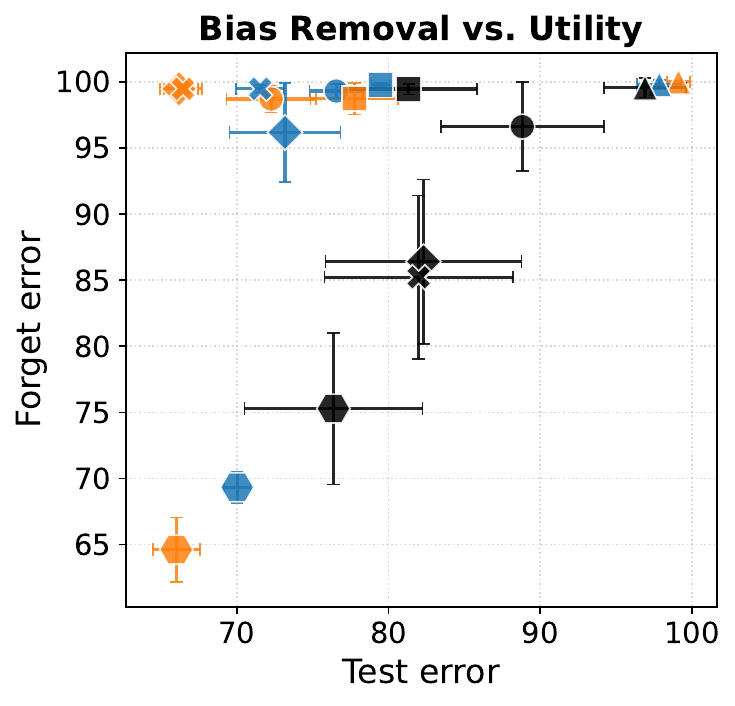}
    \includegraphics[height=0.22\textheight]{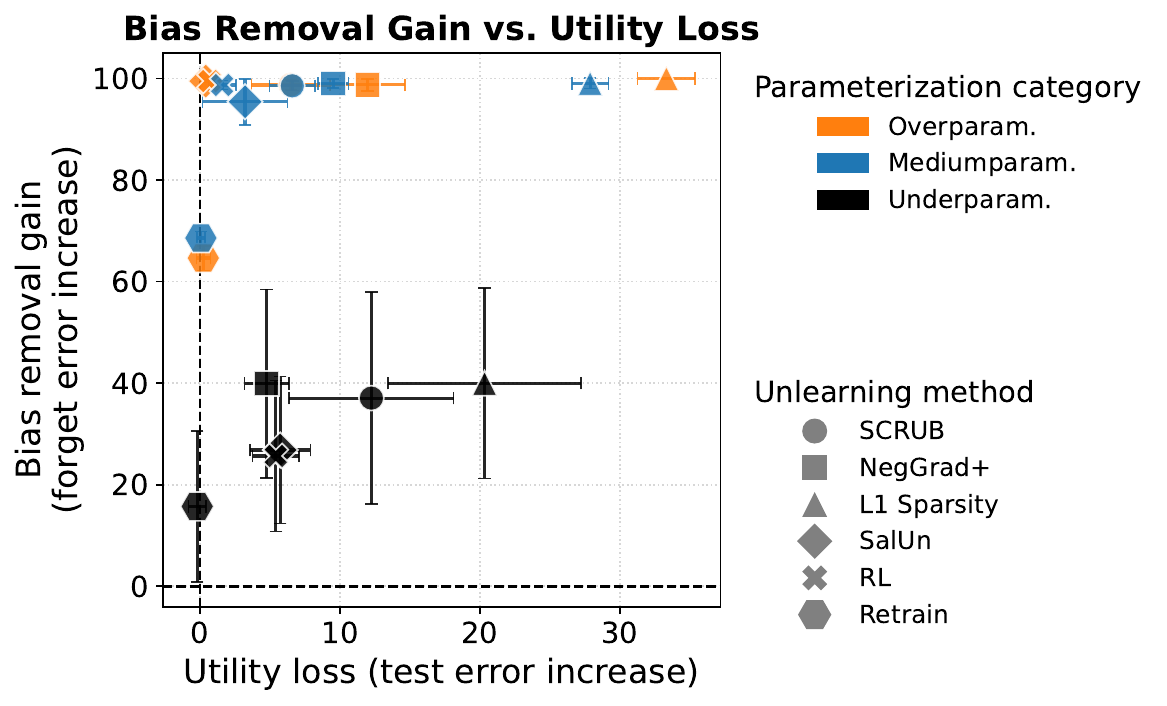}
    \caption{Unlearning for \textbf{bias removal} (ResNet-18, Tiny ImageNet, 500 unlearned examples from 10 out of 200 classes). Ideal tradeoff of the each diagram is at its top-left corner.}
    \label{fig:category performance results - tinet resnet18 unlearn 500 - bias}
\end{figure*}

\begin{figure*}[]
    \centering
    \includegraphics[height=0.22\textheight]{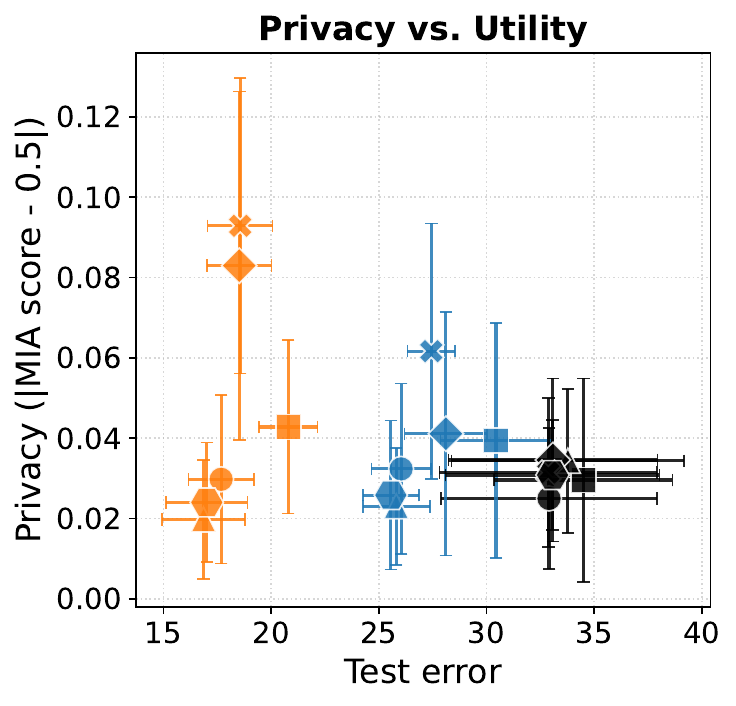}
    \includegraphics[height=0.22\textheight]{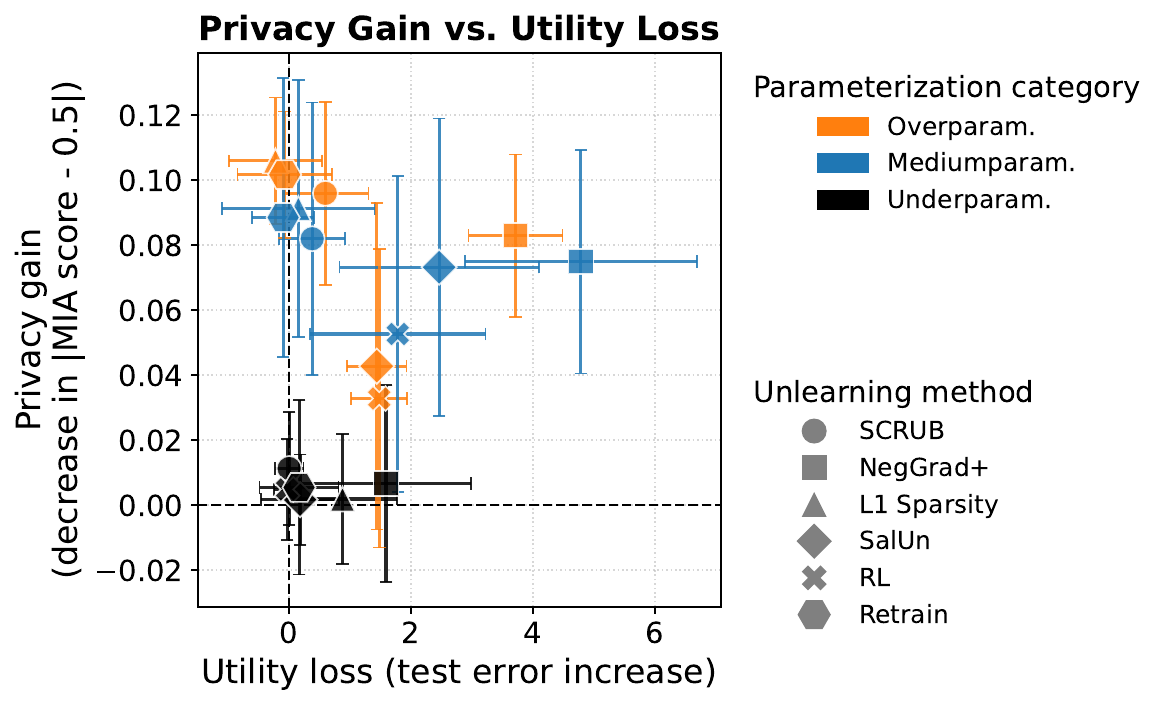}
    \caption{Unlearning for \textbf{privacy} (ResNet-18, CIFAR-10, 200 unlearned examples from 2 out of 10 classes). Ideal tradeoff of the left diagram is at its bottom-left corner. Ideal tradeoff of the right diagram is at its top-left corner.}
    \label{fig:category performance results - cifar10 resnet18 unlearn 200 - privacy}
\end{figure*}

\begin{figure*}[]
    \centering
    \includegraphics[height=0.22\textheight]{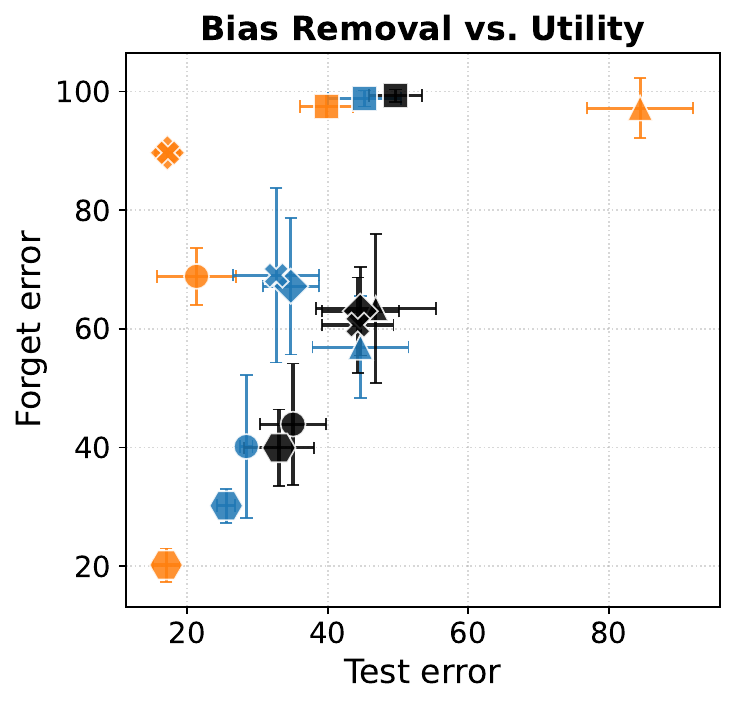}
    \includegraphics[height=0.22\textheight]{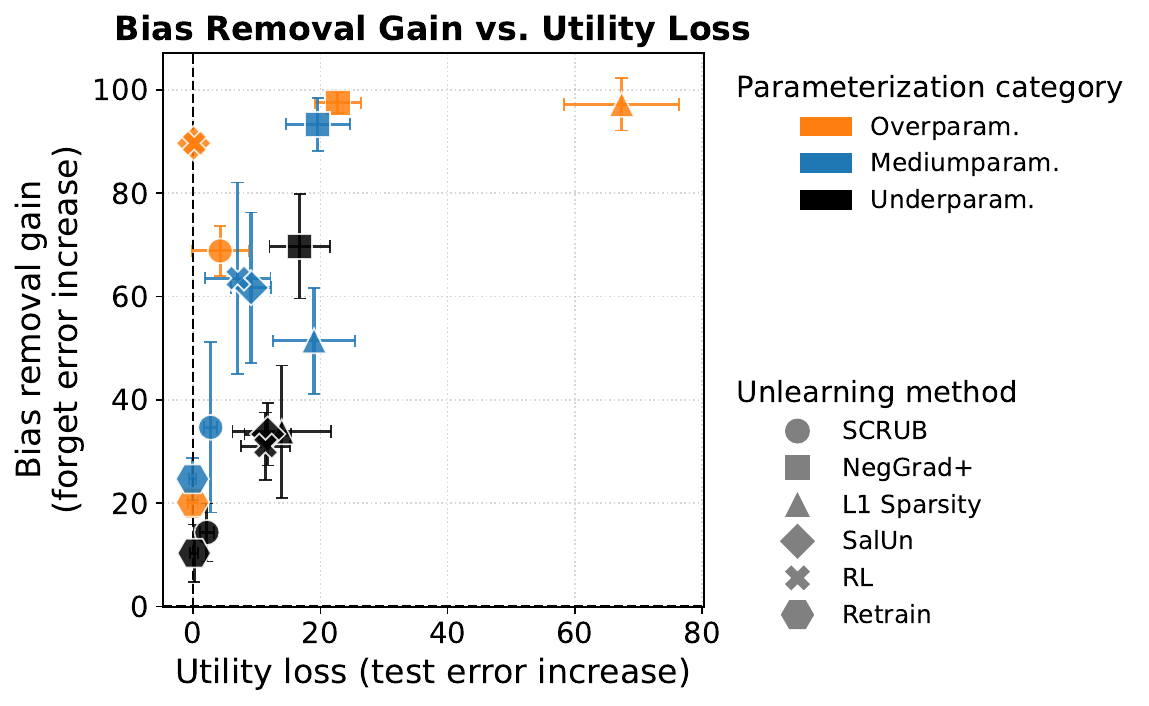}
    \caption{Unlearning for \textbf{bias removal} (ResNet-18, CIFAR-10, 200 unlearned examples from 2 out of 10 classes). Ideal tradeoff of the each diagram is at its top-left corner.}
    \label{fig:category performance results - cifar10 resnet18 unlearn 200 - bias}
\end{figure*}

\begin{figure*}[]
    \centering
    \includegraphics[height=0.22\textheight]{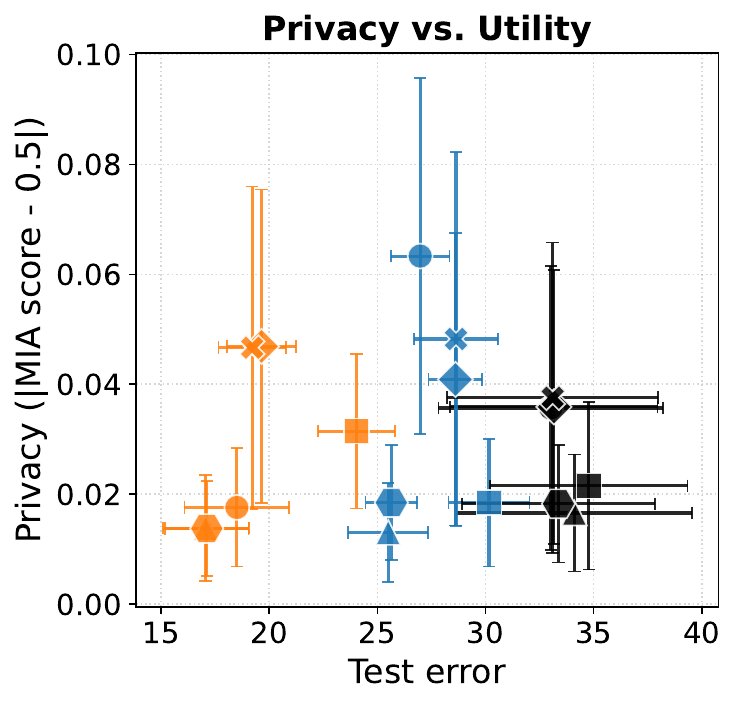}
    \includegraphics[height=0.22\textheight]{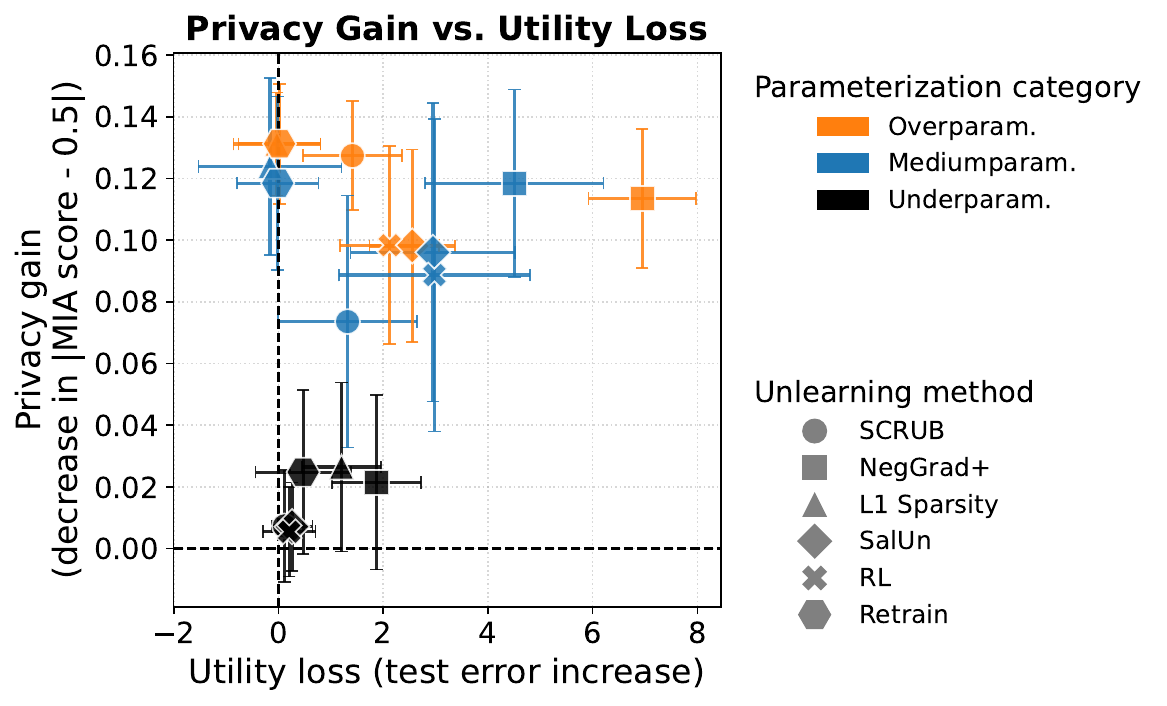}
    \caption{Unlearning for \textbf{privacy} (ResNet-18, CIFAR-10, 600 unlearned examples from 3 out of 10 classes). Ideal tradeoff of the left diagram is at its bottom-left corner. Ideal tradeoff of the right diagram is at its top-left corner.}
    \label{fig:category performance results - cifar10 resnet18 unlearn 600 - privacy}
\end{figure*}

\begin{figure*}[]
    \centering
    \includegraphics[height=0.22\textheight]{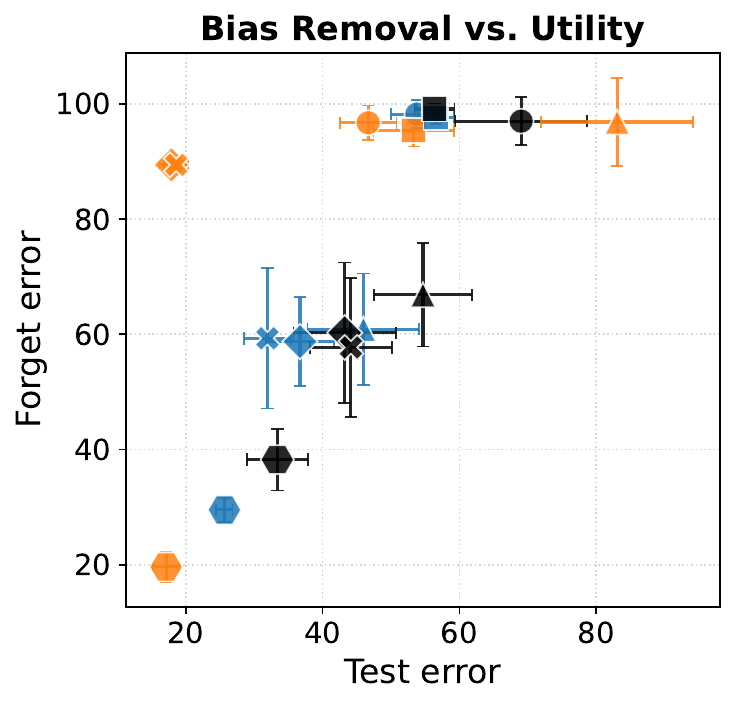}
    \includegraphics[height=0.22\textheight]{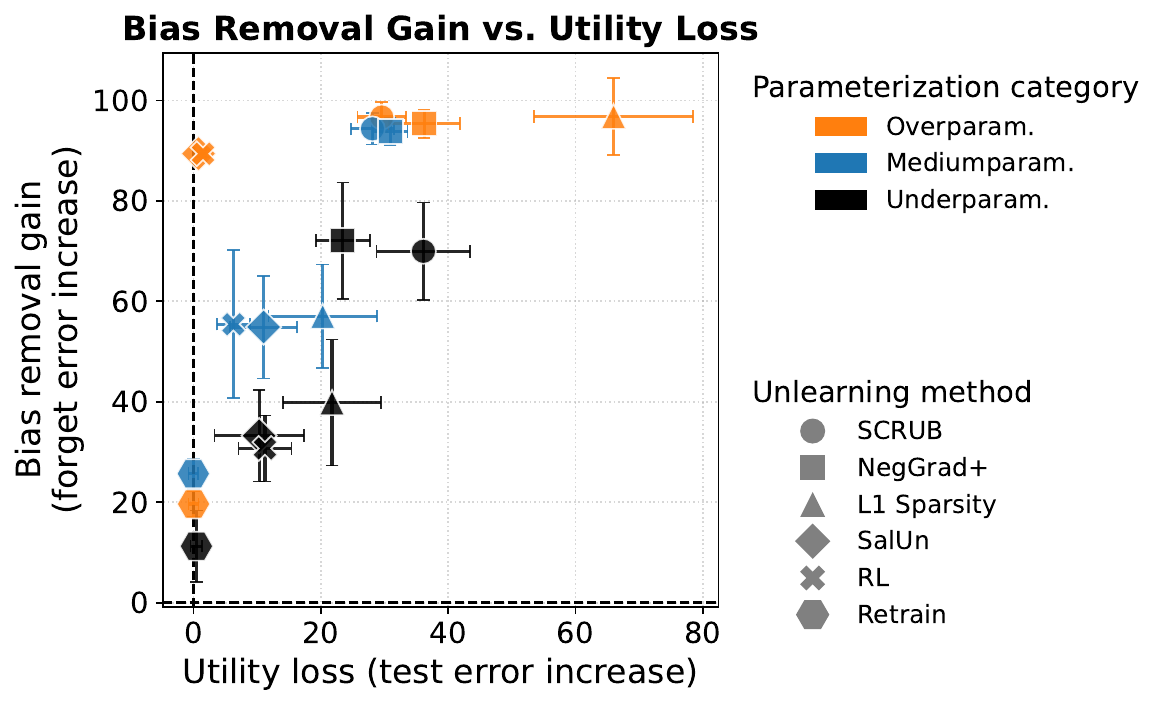}
    \caption{Unlearning for \textbf{bias removal} (ResNet-18, CIFAR-10, 600 unlearned examples from 3 out of 10 classes). Ideal tradeoff of the each diagram is at its top-left corner.}
    \label{fig:category performance results - cifar10 resnet18 unlearn 600 - bias}
\end{figure*}

\begin{figure*}[]
    \centering
    \includegraphics[height=0.22\textheight]{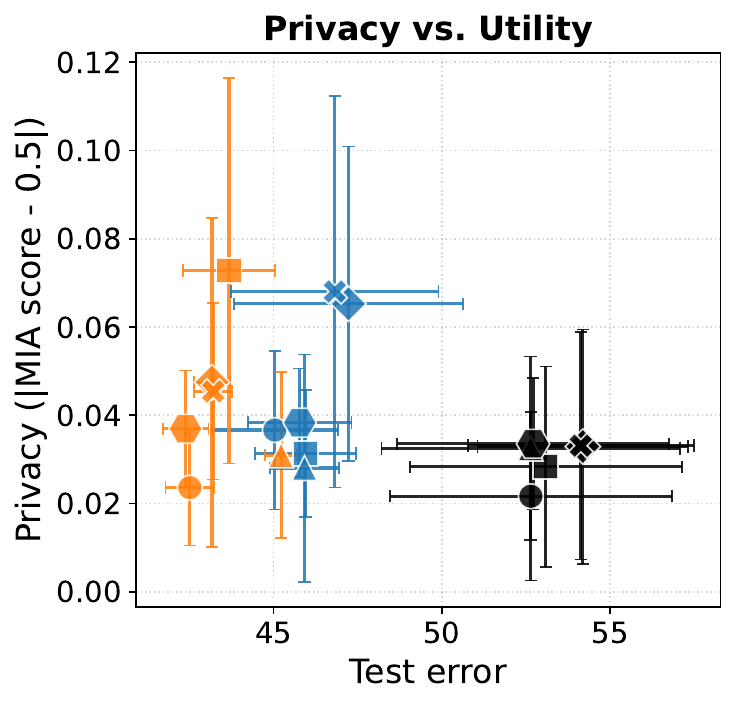}
    \includegraphics[height=0.22\textheight]{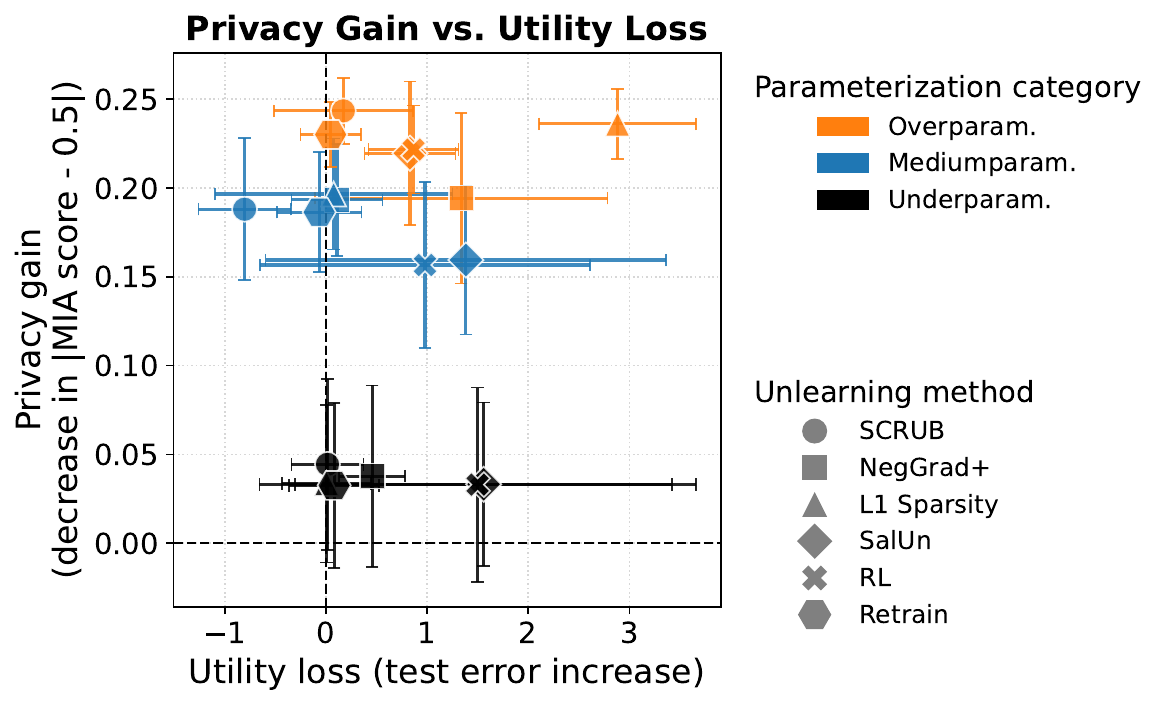}
    \caption{Unlearning for \textbf{privacy} (3-layer FC network, CIFAR-10, 200 unlearned examples from 2 out of 10 classes). Ideal tradeoff of the left diagram is at its bottom-left corner. Ideal tradeoff of the right diagram is at its top-left corner.}
    \label{fig:category performance results - cifar10 fcnet3layer unlearn 200 - privacy}
\end{figure*}

\begin{figure*}[]
    \centering
    \includegraphics[height=0.22\textheight]{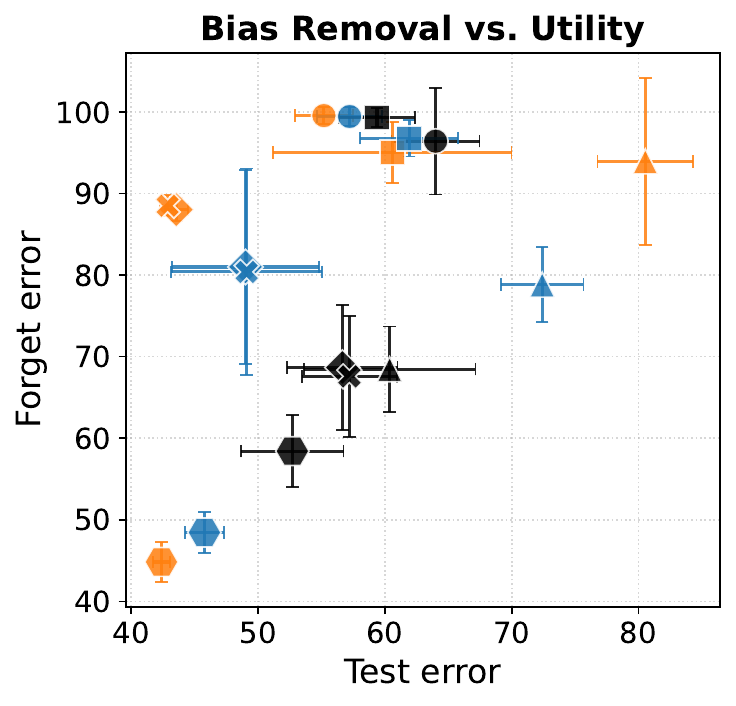}
    \includegraphics[height=0.22\textheight]{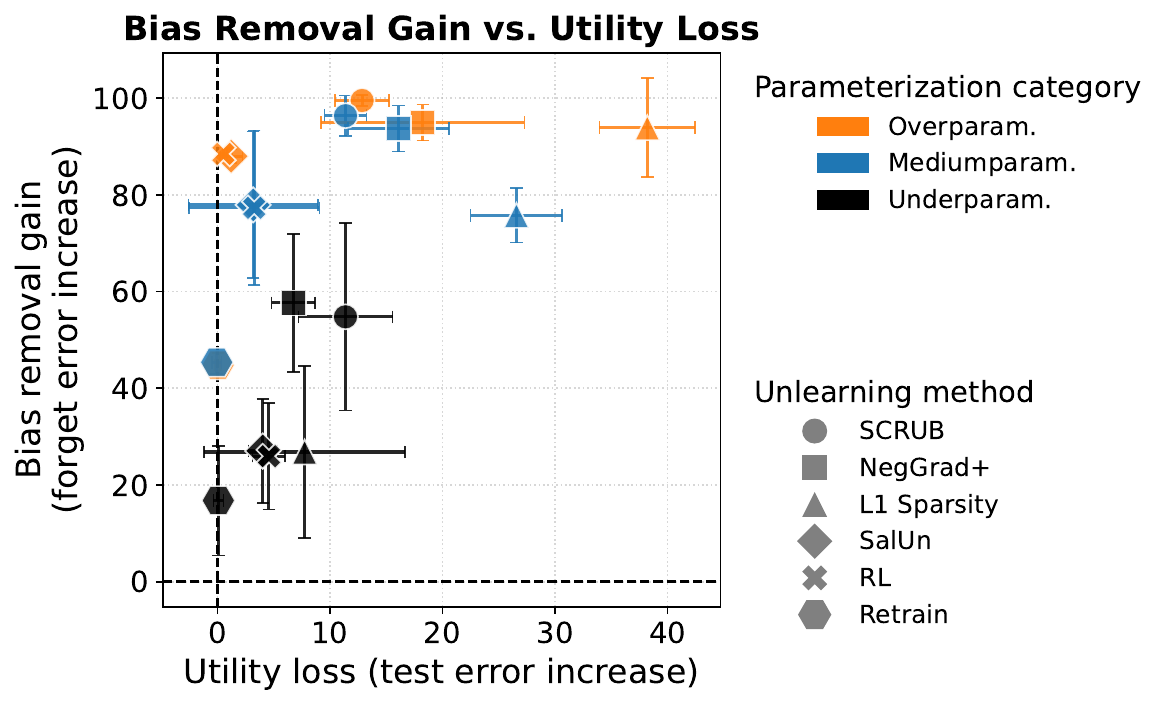}
    \caption{Unlearning for \textbf{bias removal} (3-layer FC network, CIFAR-10, 200 unlearned examples from 2 out of 10 classes). Ideal tradeoff of the each diagram is at its top-left corner.}
    \label{fig:category performance results - cifar10 fcnet3layer unlearn 200 - bias}
\end{figure*}

\begin{figure*}[t] % added [t] to suggest placing it at the top of the page
    \centering  
    
    % --- First Row ---
    \begin{subfigure}[b]{0.48\textwidth}
        \centering
        \includegraphics[height=0.65\textwidth]{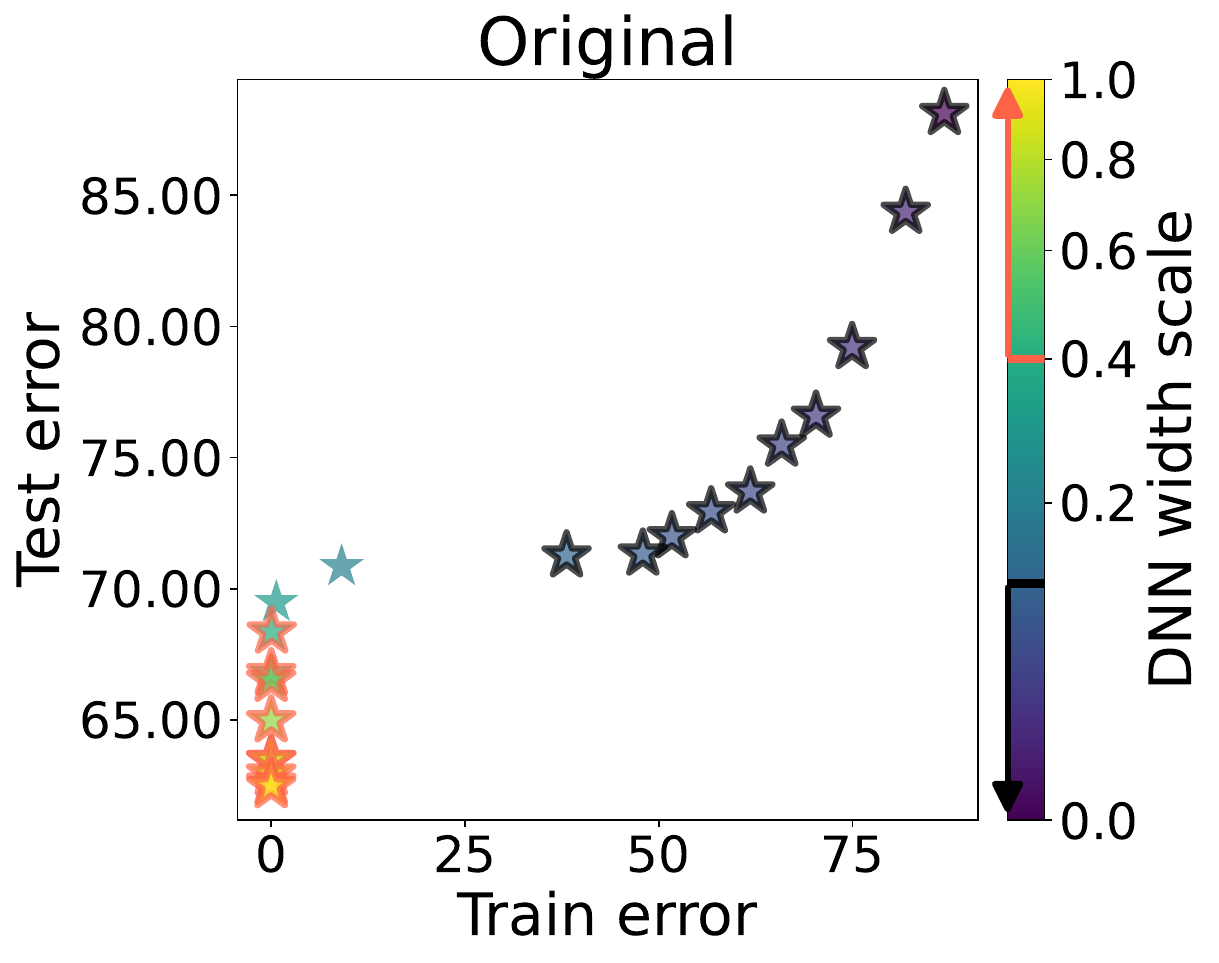}
        \caption{ResNet-34 on Tiny ImageNet}
        \label{fig:orig_res34_tinet_train_error_vs_test_error}
    \end{subfigure}
    \hfill
    \begin{subfigure}[b]{0.48\textwidth}
        \centering
        \includegraphics[height=0.65\textwidth]{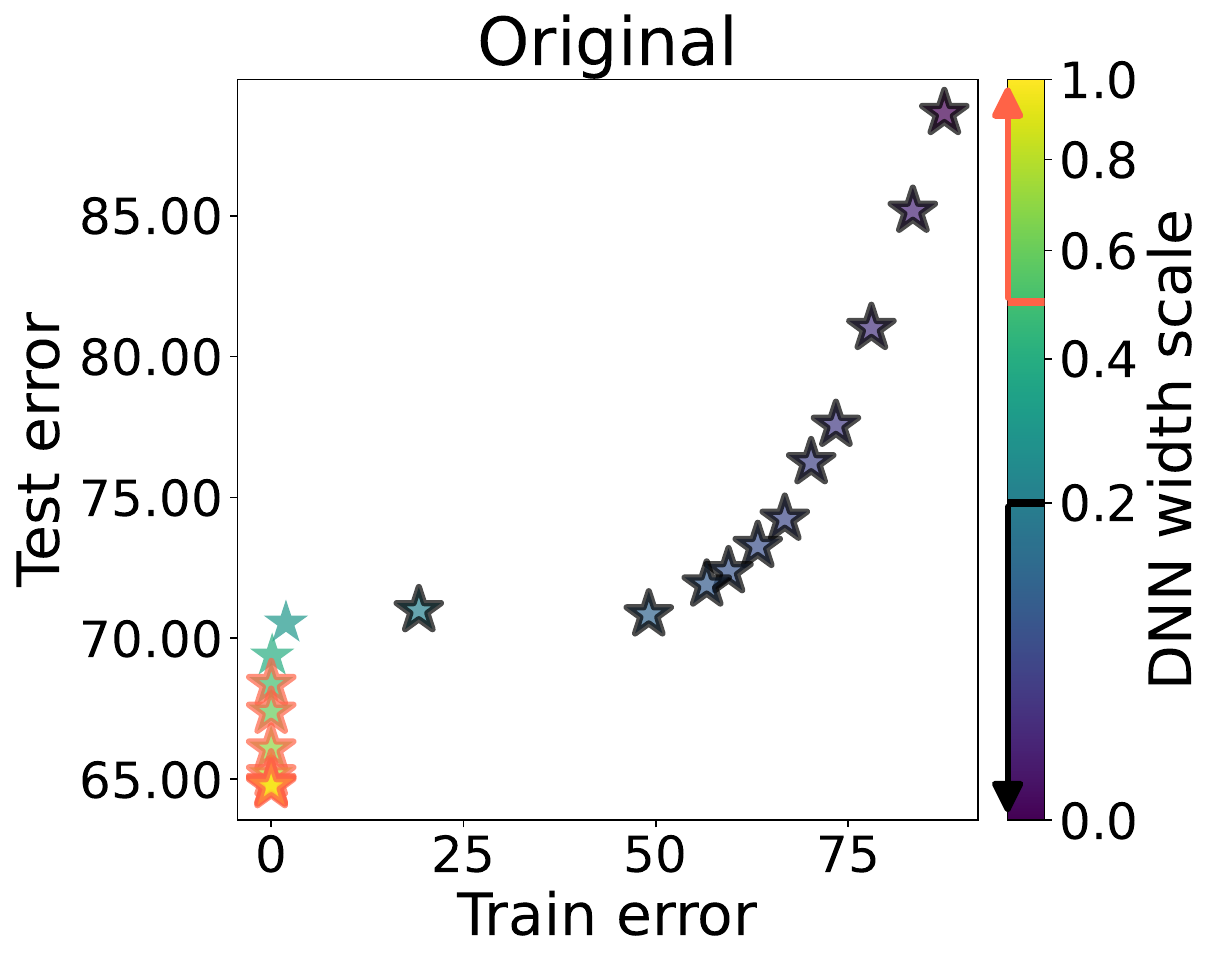}
        \caption{ResNet-18 on Tiny ImageNet}
        \label{fig:orig_res18_tinet_train_error_vs_test_error}
    \end{subfigure}
    
    \vspace{0.5cm} % Adds a little vertical breathing room between the rows
    
    % --- Second Row ---
    \begin{subfigure}[b]{0.48\textwidth}
        \centering
        \includegraphics[height=0.65\textwidth]{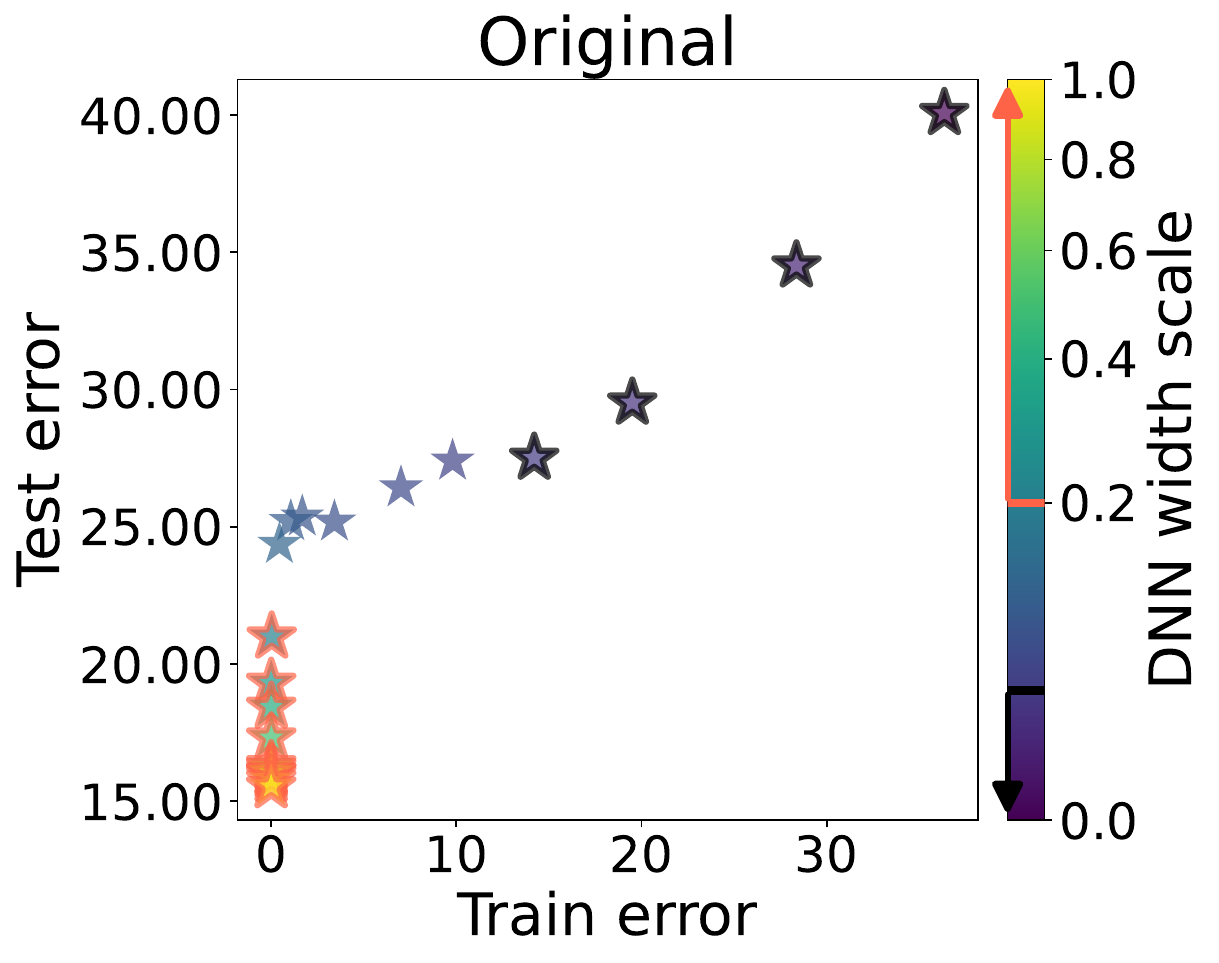}
        \caption{ResNet-18 on CIFAR-10}
        \label{fig:orig_res18_cifar_train_error_vs_test_error}
    \end{subfigure}
    \hfill
    \begin{subfigure}[b]{0.48\textwidth}
        \centering
        \includegraphics[height=0.65\textwidth]{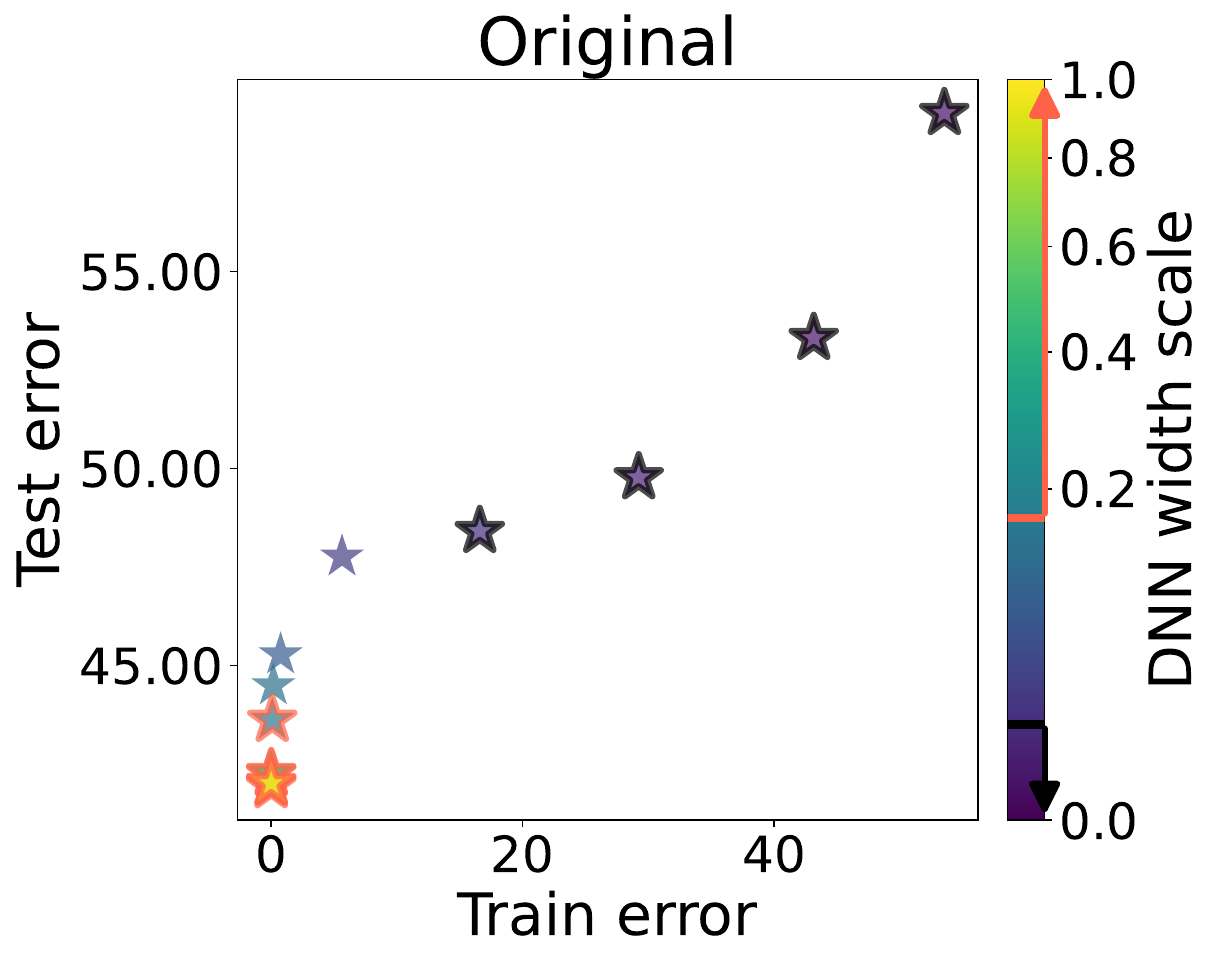}
        \caption{3-layer FC network on CIFAR-10}
        \label{fig:orig_fc3_cifar_train_error_vs_test_error}
    \end{subfigure}
    
    \caption{Test error vs.~\textbf{train error} of the original models before unlearning.}
    \label{fig:original_models_train_error_vs_test_error}
\end{figure*}

In Figs.~\ref{fig:performance results - tinet resnet50 unlearn 1000 - privacy}-\ref{fig:category performance results - cifar10 fcnet3layer unlearn 200 - bias}  we extend the performance evaluations presented in the main paper by exploring a broader range of unlearning settings. 
In Fig.~\ref{fig:original_models_train_error_vs_test_error} we provide diagrams of the train errors vs.~test errors of the original models at the various widths.

Note that in Figures~\ref{fig:performance evaluation - resnet 50 tinet unlearn 1000 - privacy - main paper}, \ref{fig:performance evaluation - resnet 50 tinet unlearn 1000 - bias - main paper}, \ref{fig:performance results - tinet resnet50 unlearn 1000 - privacy}-\ref{fig:performance results - fcnet3layer unlearn 200 - bias}, the diagrams of the five unlearning methods show the performance evaluation markers separately for each of the three runs (with three different random seeds); in contrast, the diagrams of the retrain and original model show markers for the average performance evaluation over the three runs. All the other diagrams in this paper show markers for evaluations that include averaging over the three runs (in some cases the averaging may include additional aspects, e.g., in the parameterization-category diagrams).

The provided evaluations use the following $\lambda$ values that balance between the unlearning goal and generalization in the validation score function (\ref{eq:validation score formula}):
\begin{itemize}
    \item Evaluation of parameterization levels (Figs.~\ref{fig:performance evaluation - resnet 50 tinet unlearn 1000 - privacy - main paper}, \ref{fig:performance evaluation - resnet 50 tinet unlearn 1000 - bias - main paper}, \ref{fig:performance results - tinet resnet50 unlearn 1000 - privacy}-\ref{fig:performance results - fcnet3layer unlearn 200 - bias}) show diagrams that each considers two values of $\lambda$; the marker size corresponds to the $\lambda$ value. The results for the privacy goal are for $\lambda\in\{0.1,0.2\}$ (except for CIFAR-10 on 200 unlearned examples where $\lambda\in\{0.1,0.3\}$). The results for the bias removal goal are for $\lambda\in\{0.3,0.5\}$.
    \item Evaluation of parameterization categories (overparameterized vs.~underparameterized) and the evaluation of unlearned vs.~original performance  (Figs.~\ref{fig:category performance results - tinet resnet50 unlearn 1000 - main paper}, \ref{fig:category performance results - tinet resnet50 unlearn 1000 - privacy}-\ref{fig:category performance results - cifar10 fcnet3layer unlearn 200 - bias}) are for $\lambda=0.1$ for the privacy goal, and $\lambda=0.3$ for the bias removal goal. 
\end{itemize}

%%%%%%%%%%%%%%%%%% category-based graphs




%%%%%%%%%%%%%%

%%% Original models train vs test errors
%%%

\section{Membership Inference Attack (MIA) Experiments}
\label{appendix:subsec:Membership Inference Attack (MIA) Methodology}

The privacy experiments include measurements of the Membership Inference Attack (MIA) accuracy. To calculate the MIA accuracy, we implemented the principles from \cite{forggeting_outside_the_box,unbounded_MUL} as follows:

\begin{enumerate}
    \item \textit{Dataset creation}:  We constructed a balanced binary classification dataset using the cross-entropy loss as the single input feature. Examples from the forget set were labeled as ``members'', while an equal number of randomly sampled examples from the unseen test set (specifically, from classes that also appear in the forget set) were labeled as ``non-members''. To stabilize the solver, the loss features were clipped to the range $[-100,100]$.
    
    \item \textit{Model training}: Using this balanced dataset, we evaluated a standard logistic regression model using 10-fold stratified cross-validation to ensure robustness against overfitting.
    
    \item \textit{MIA accuracy calculation}: The MIA accuracy represents the success rate of the attack in distinguishing the unlearned forget set examples from the unseen test examples (specifically, from classes that also appear in the forget set). A model with the \textbf{best possible privacy yields an MIA score of 0.5}, equivalent to random guessing.

\end{enumerate}

\clearpage

\section{Additional Details and Results for Decision Region Analysis}

\subsection{Decision Region Sampling and Similarity Calculation}
\label{appendix:subsec:Decision Region Sampling and Similarity Calculation}
To evaluate the similarities and changes across decision regions of different models, we extended the approach from \cite{somepalli_can_nn_learn_twice} for constructing planes in the input space for decision region similarity evaluation. Here, our extension considers the aspects of unlearning, which are beyond the scope of \cite{somepalli_can_nn_learn_twice}. Our process is as follows:

\begin{enumerate}
    \item \textit{Image selection}: For each plane, we randomly selected three images: one from the forget set and two from the retain set (composed of the remaining images in the training dataset).

    \item \textit{Plane construction}: Using the three selected images, we constructed a (truncated) plane that is spanned by, and contains, the three images. The construction of the plane given three images is as explained in \cite{somepalli_can_nn_learn_twice}.

    \item \textit{Grid sampling}: From the constructed plane, we sampled \(50 \times 50 \) uniformly-spaced points that form a 2D discrete grid on the plane to create new images and evaluating decision regions.

    \item \textit{Unlearning similarity calculation}: 
    Unlearning similarity scores were computed using Eq.~(\ref{eq: unlearning similarity score - unlearning of specific samples})-(\ref{eq: unlearning similarity score - unlearning of specific samples - regions far from unlearned samples}) for unlearning. See the detailed explanations in Section \ref{subsec:decision region analysis - Case I - Unlearning of Specific Training Samples}.

    \item \textit{Plane averaging}: We repeated the above process for \(300\) planes. The similarity scores were averaged across these planes to obtain stable and reliable metrics for similarity and change.
\end{enumerate}

This method provides a comprehensive evaluation of decision regions, enabling us to analyze the effects of unlearning methods.

\subsection{Additional Decision Regions Results}

\begin{figure*}[]
    \centering    
    \includegraphics[height=0.3\textwidth]{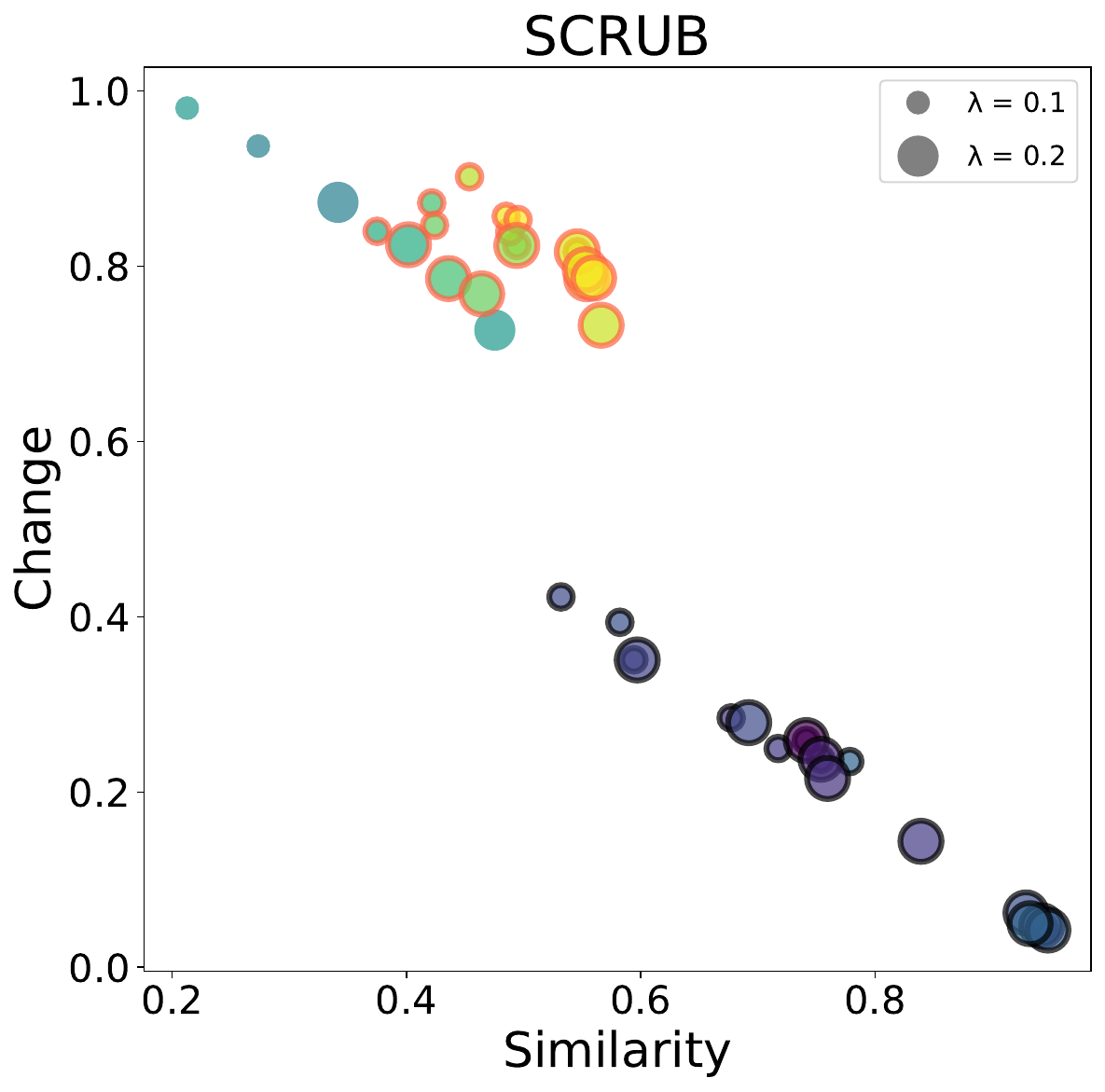}
    \includegraphics[height=0.3\textwidth]{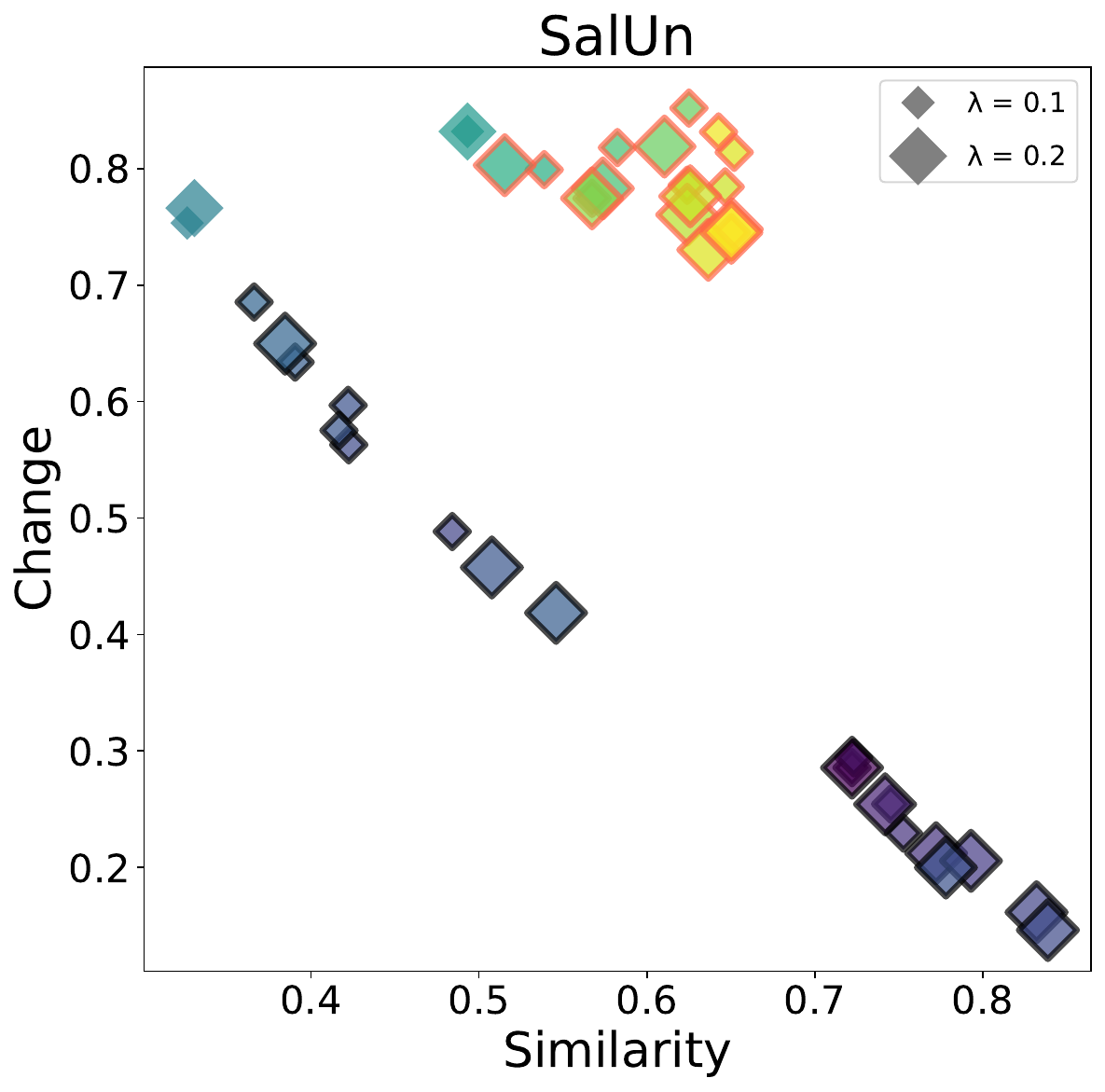}
    \includegraphics[height=0.3\textwidth]{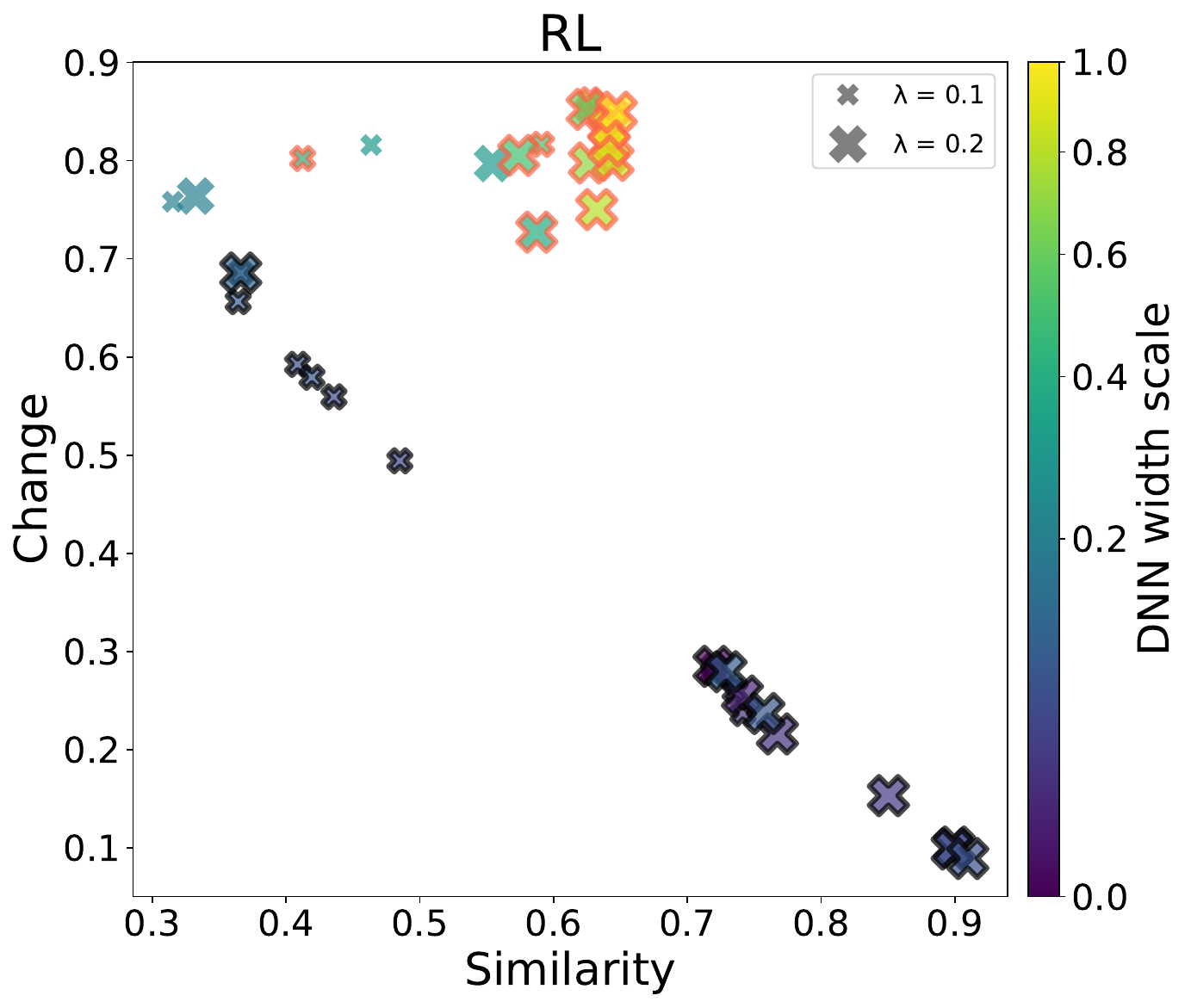}
    \\
    \includegraphics[height=0.3\textwidth]{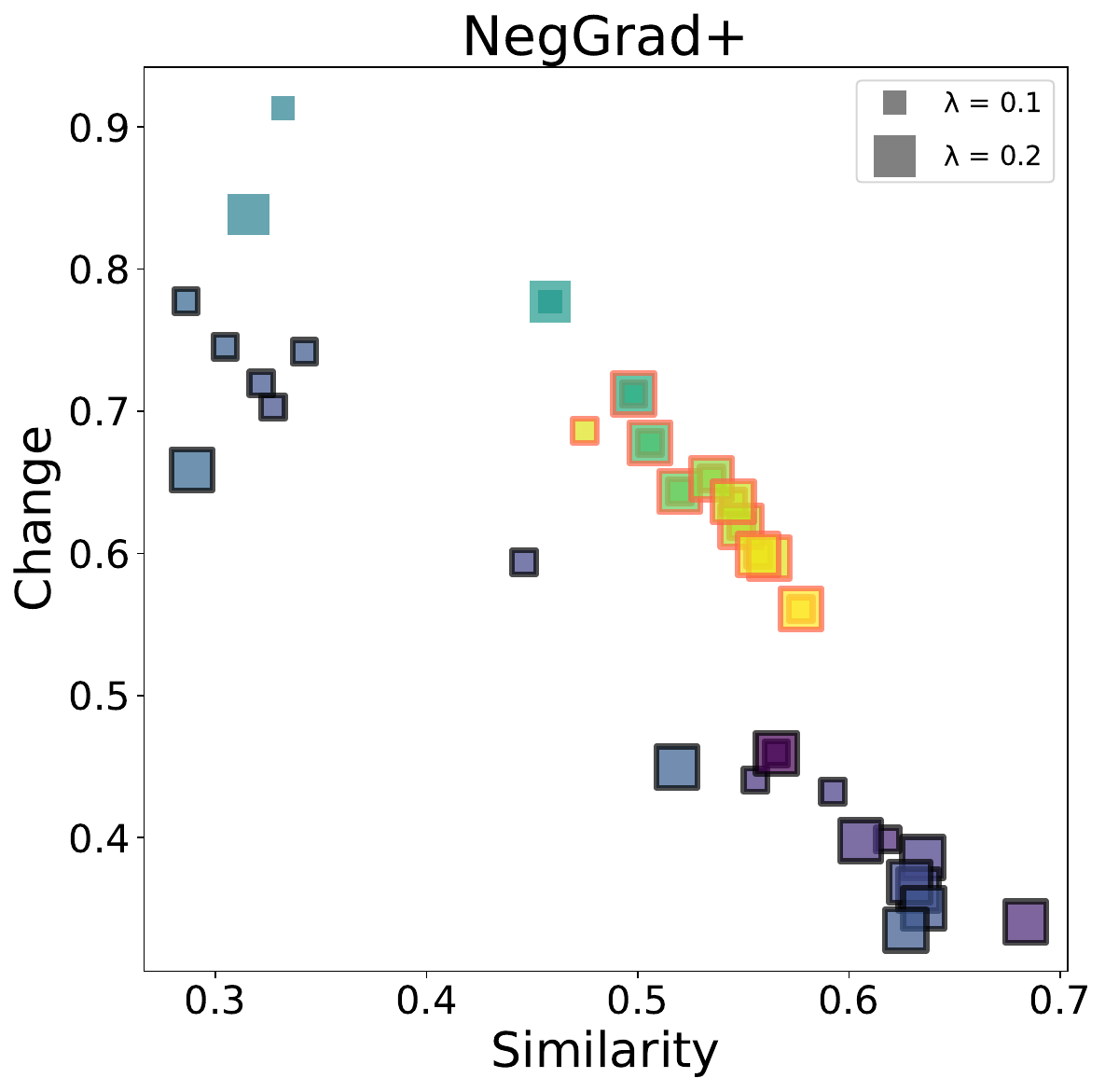}
    \includegraphics[height=0.3\textwidth]{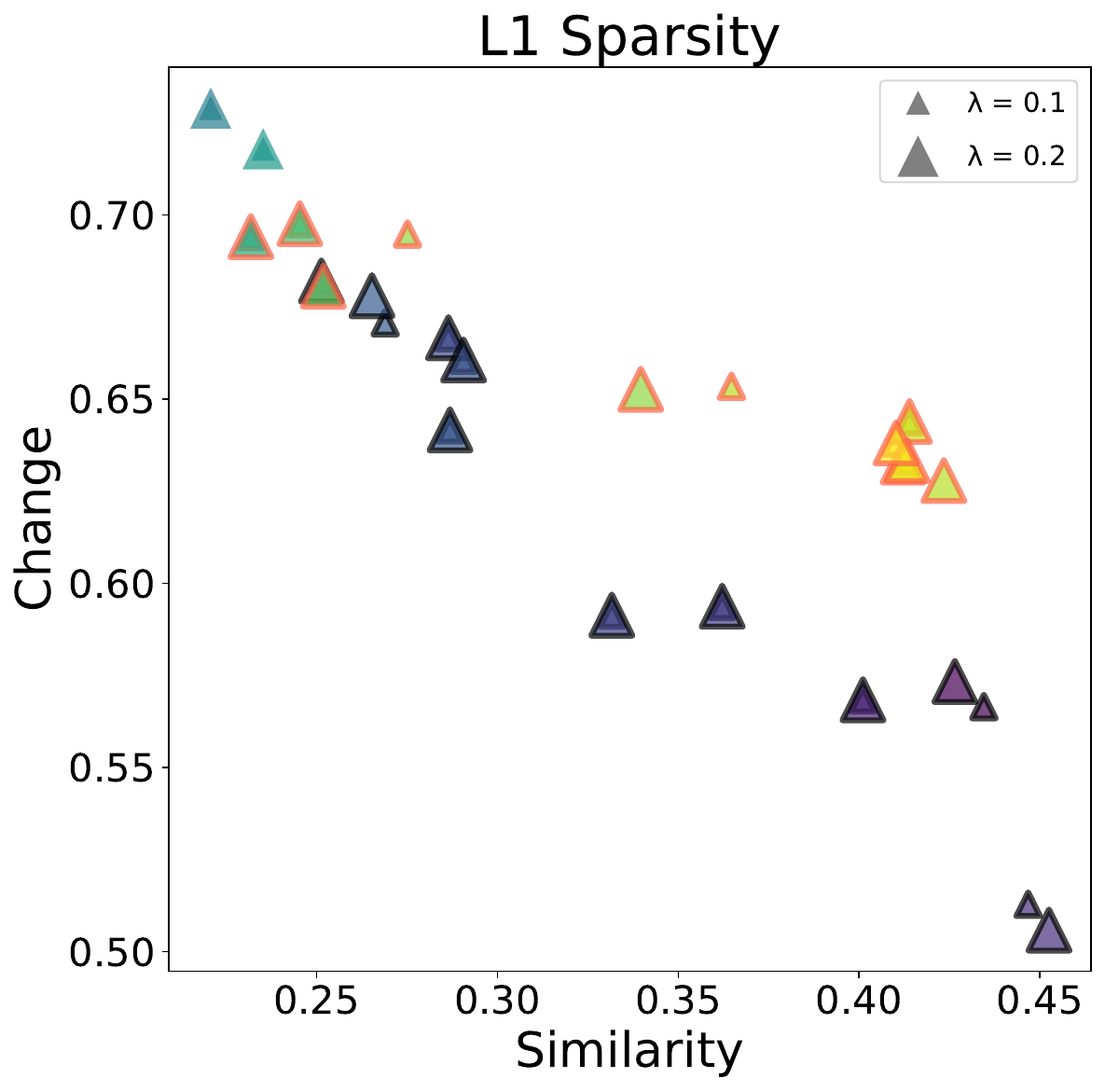}
    \includegraphics[height=0.3\textwidth]{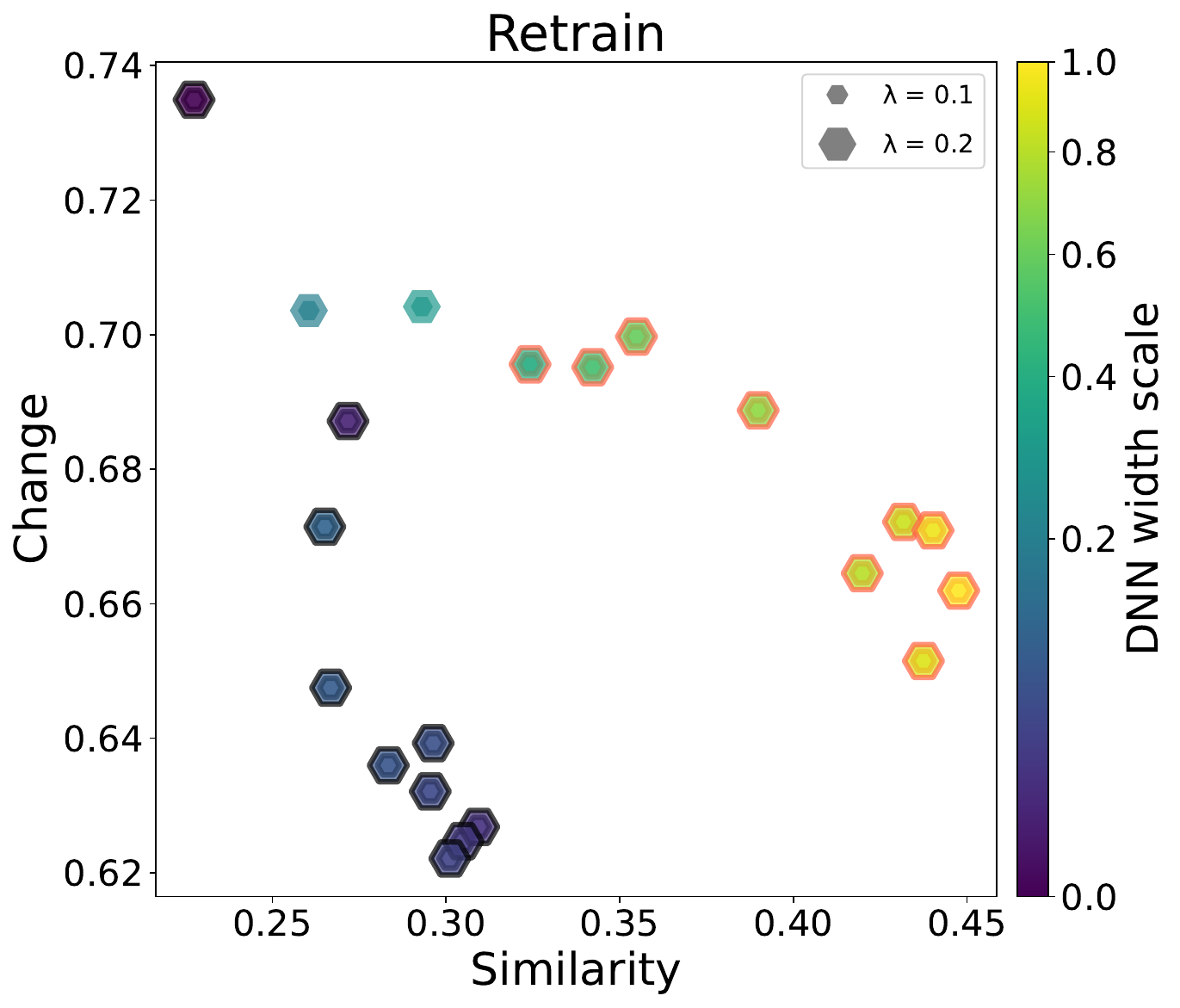}
    \caption{Decision-region similarity and change scores for \textbf{privacy} unlearning. ResNet-34 on Tiny ImageNet (1000 unlearned examples, $\delta=10$). Markers closer to the top-right corner at each diagram denote unlearning with more-local changes of decision regions.}
    \label{fig:similarity_change_privacy_resnet50_tinet_1000unlearned}
\end{figure*}

\begin{figure*}[]
    \centering    
    \includegraphics[height=0.3\textwidth]{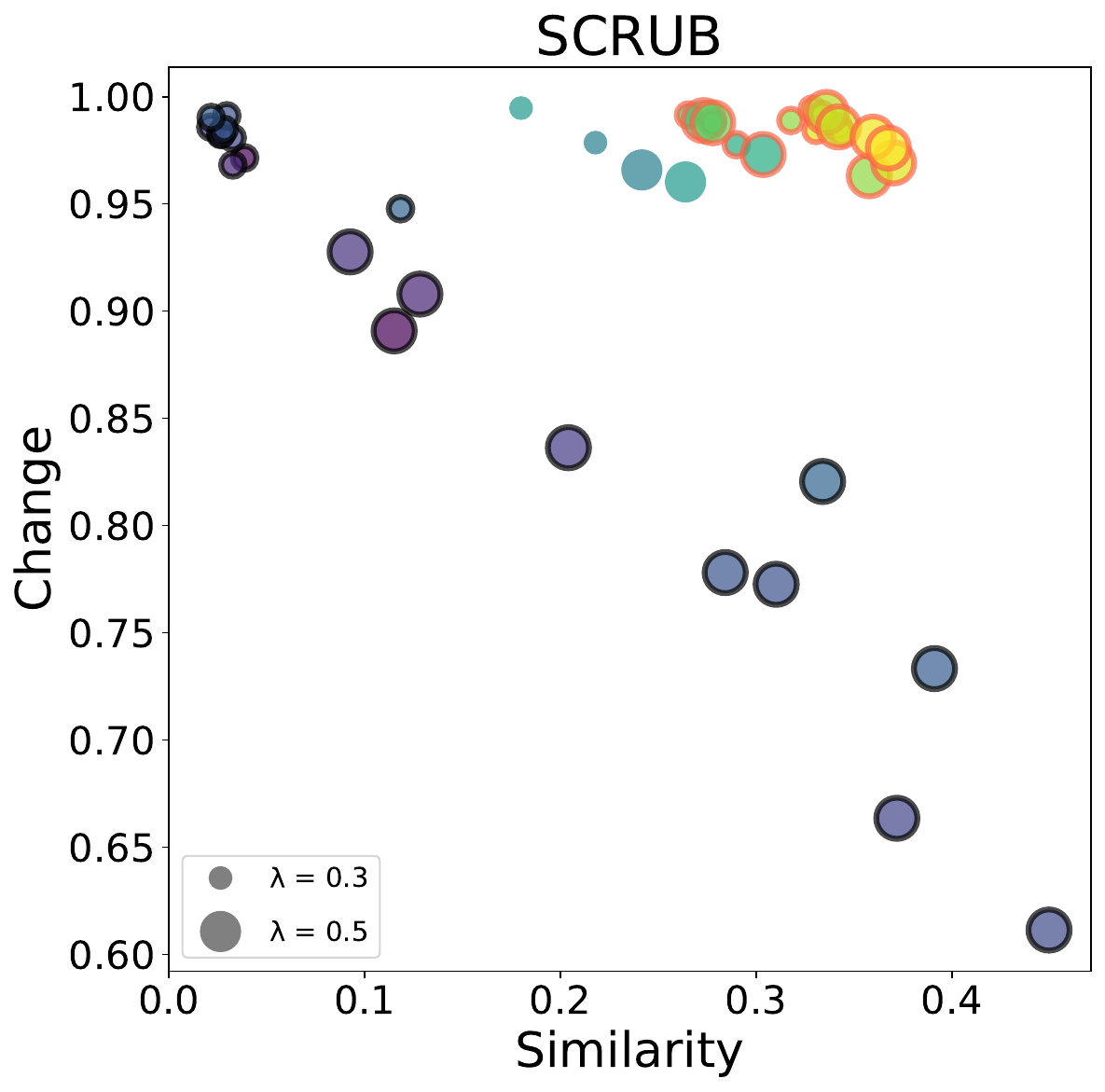}
    \includegraphics[height=0.3\textwidth]{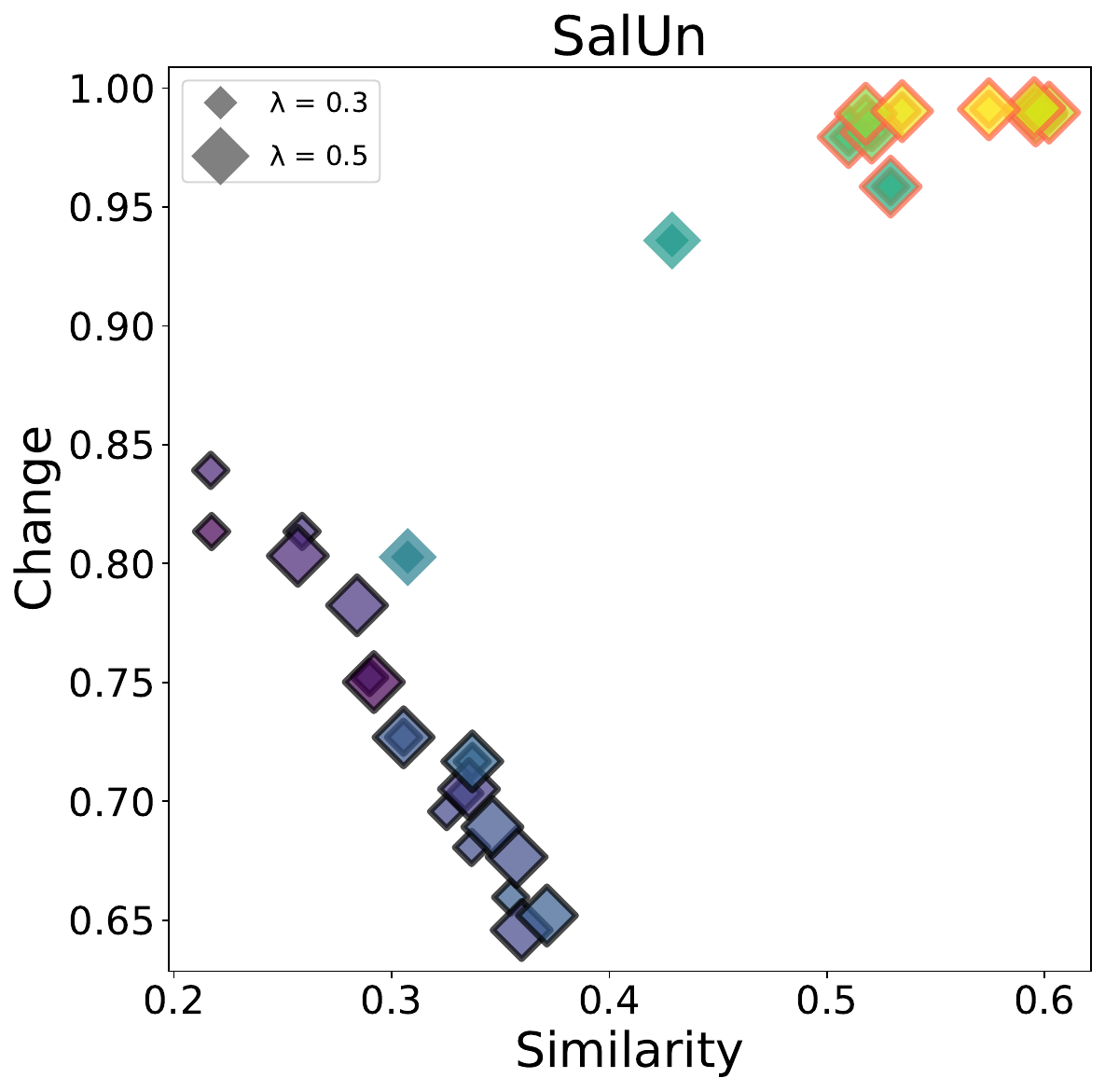}
    \includegraphics[height=0.3\textwidth]{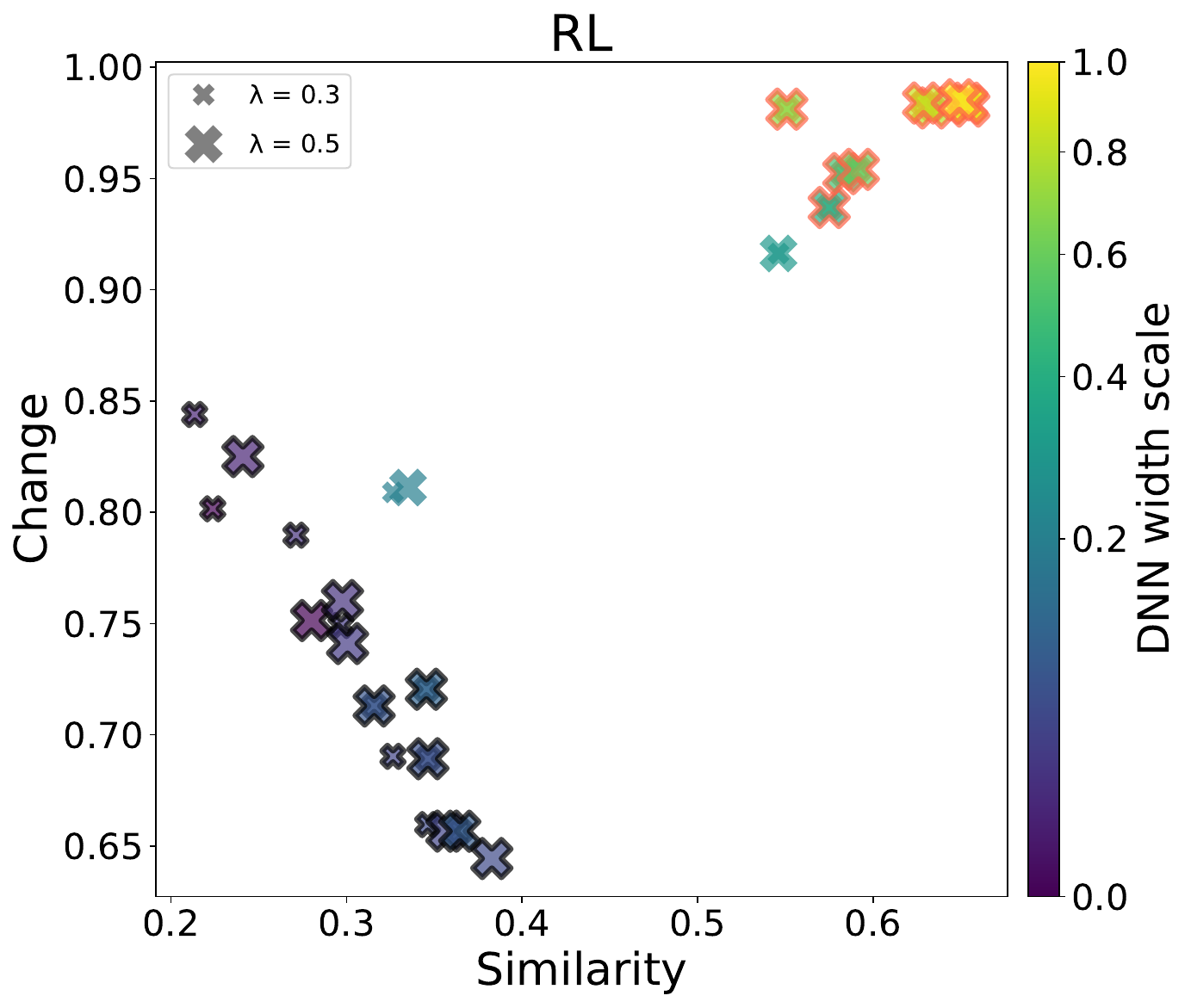}
    \\
    \includegraphics[height=0.3\textwidth]{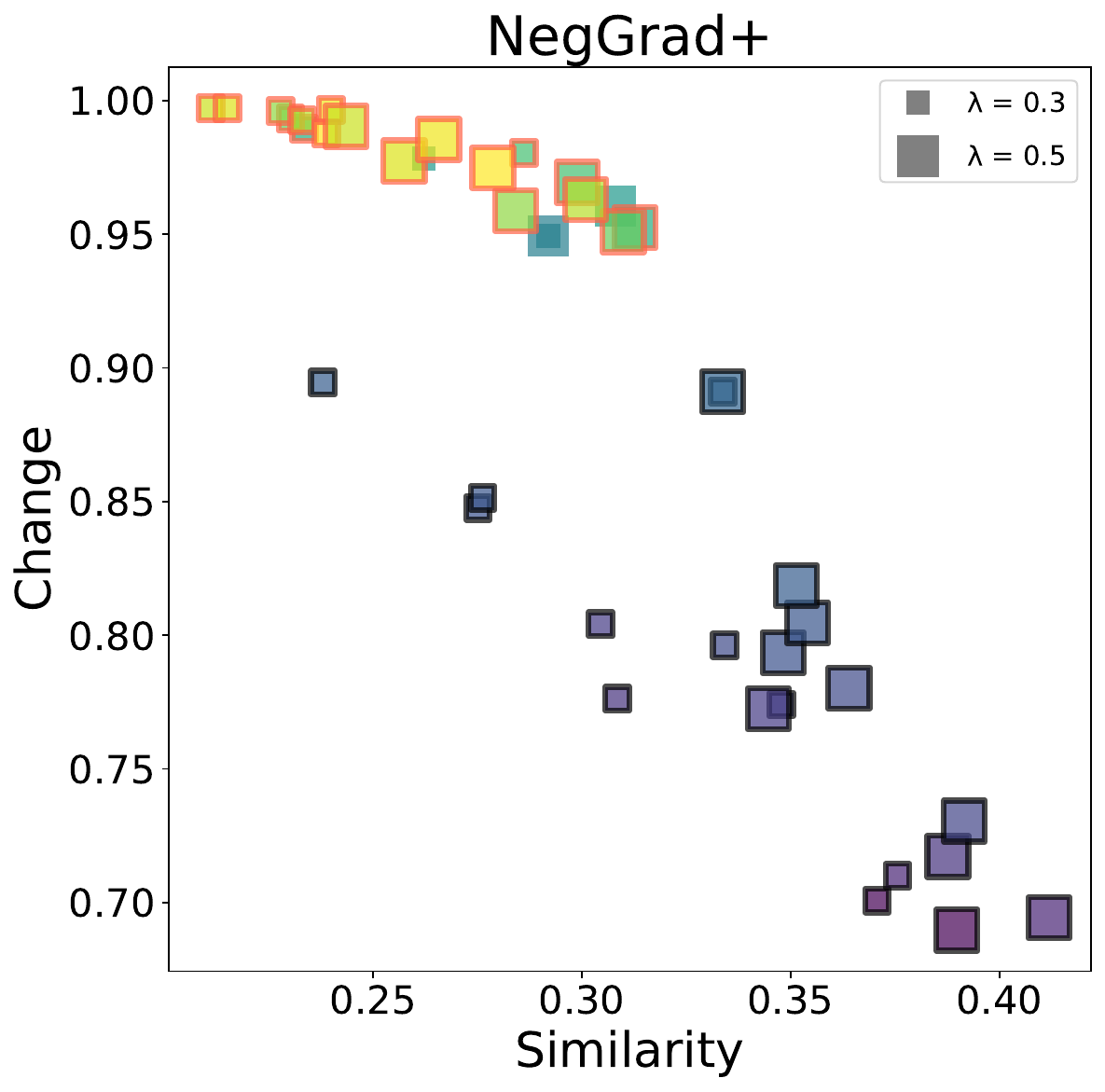}
    \includegraphics[height=0.3\textwidth]{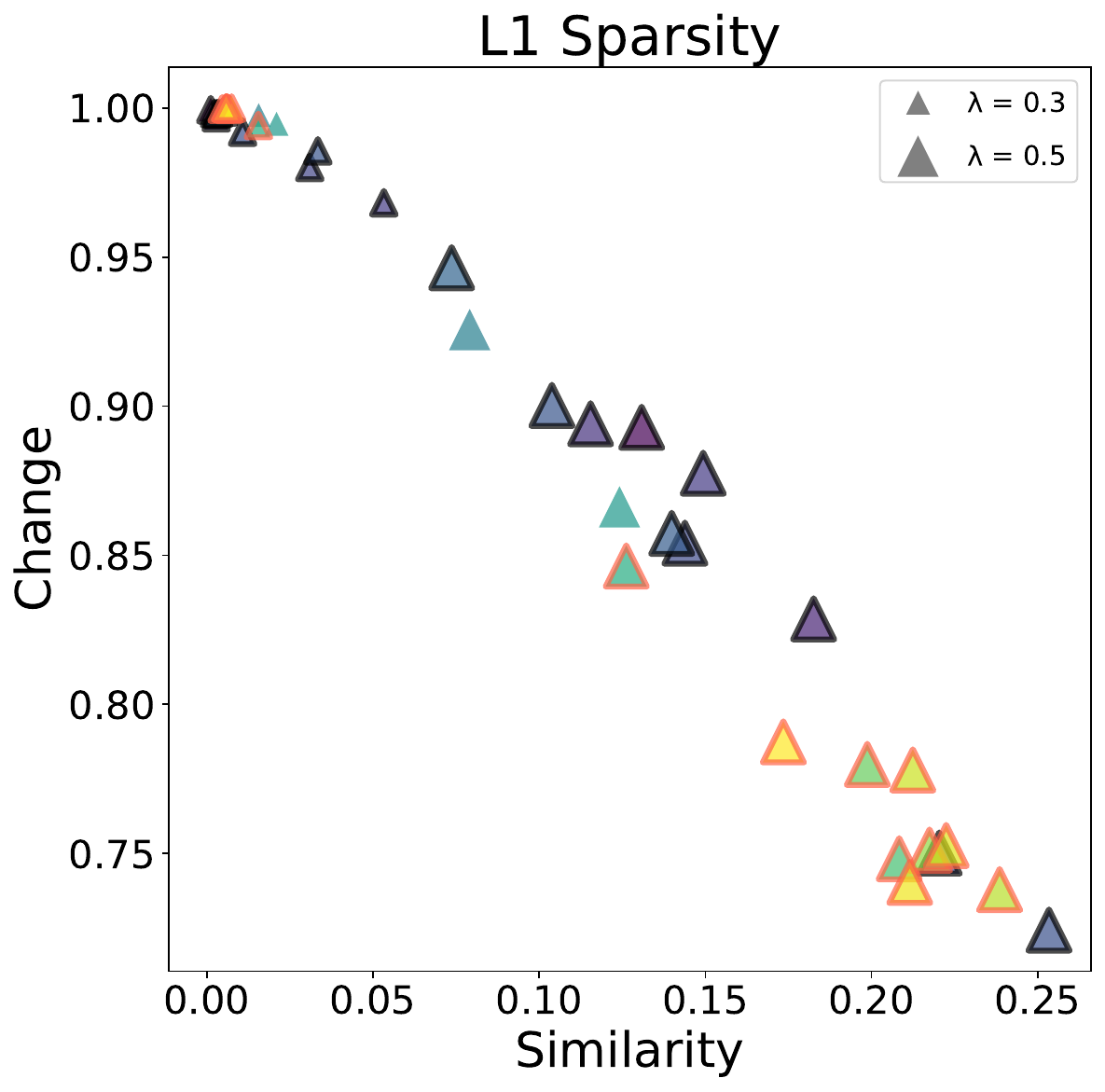}
    \includegraphics[height=0.3\textwidth]{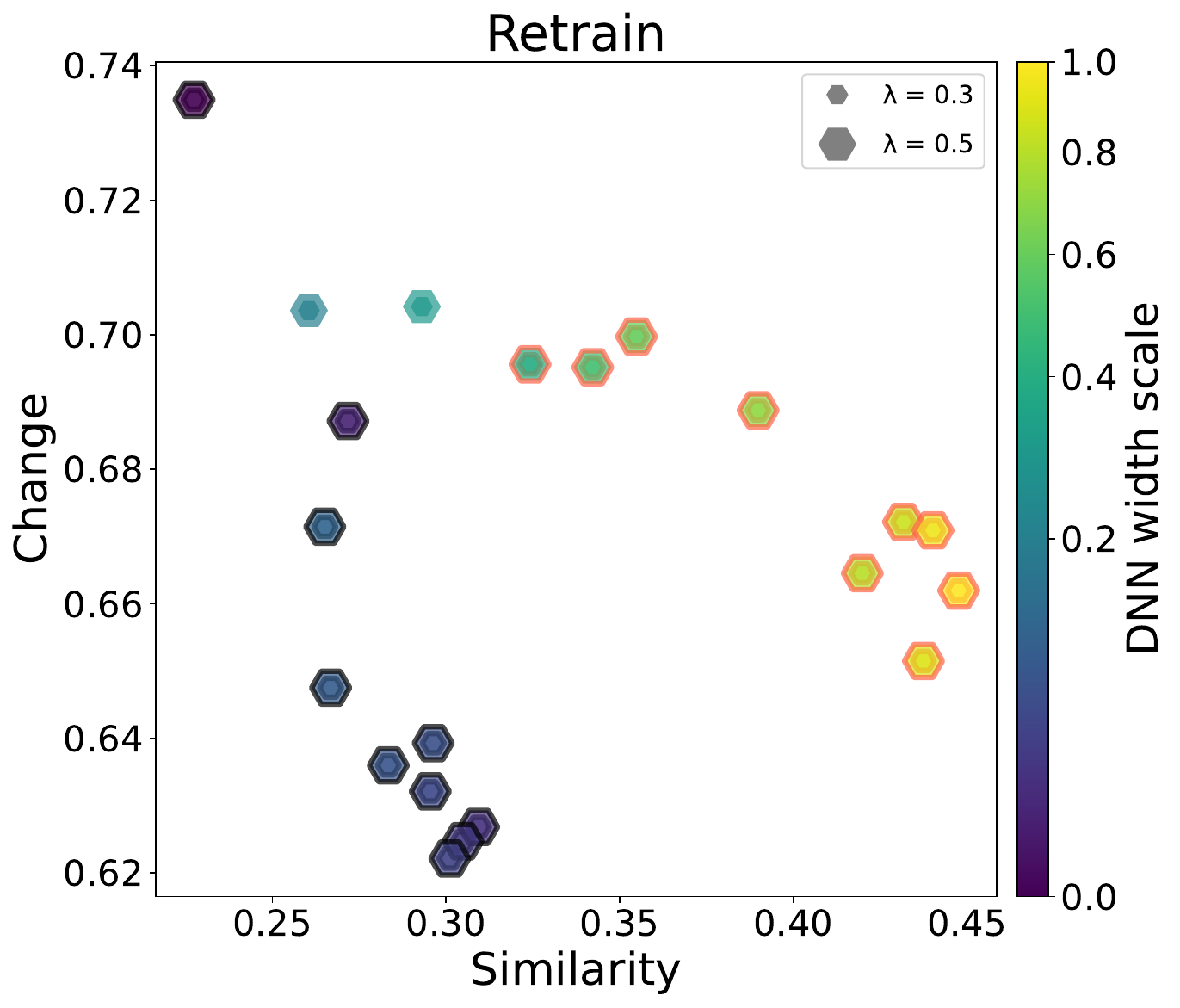}
    \caption{Decision-region similarity and change scores for \textbf{bias removal} unlearning. ResNet-34 on Tiny ImageNet (1000 unlearned examples, $\delta=10$).  Markers closer to the top-right corner at each diagram denote unlearning with more-local changes of decision regions.}
    \label{fig:similarity_change_bias_resnet50_tinet_1000unlearned}
\end{figure*}

\begin{figure*}[]
    \centering    
    \includegraphics[height=0.3\textwidth]{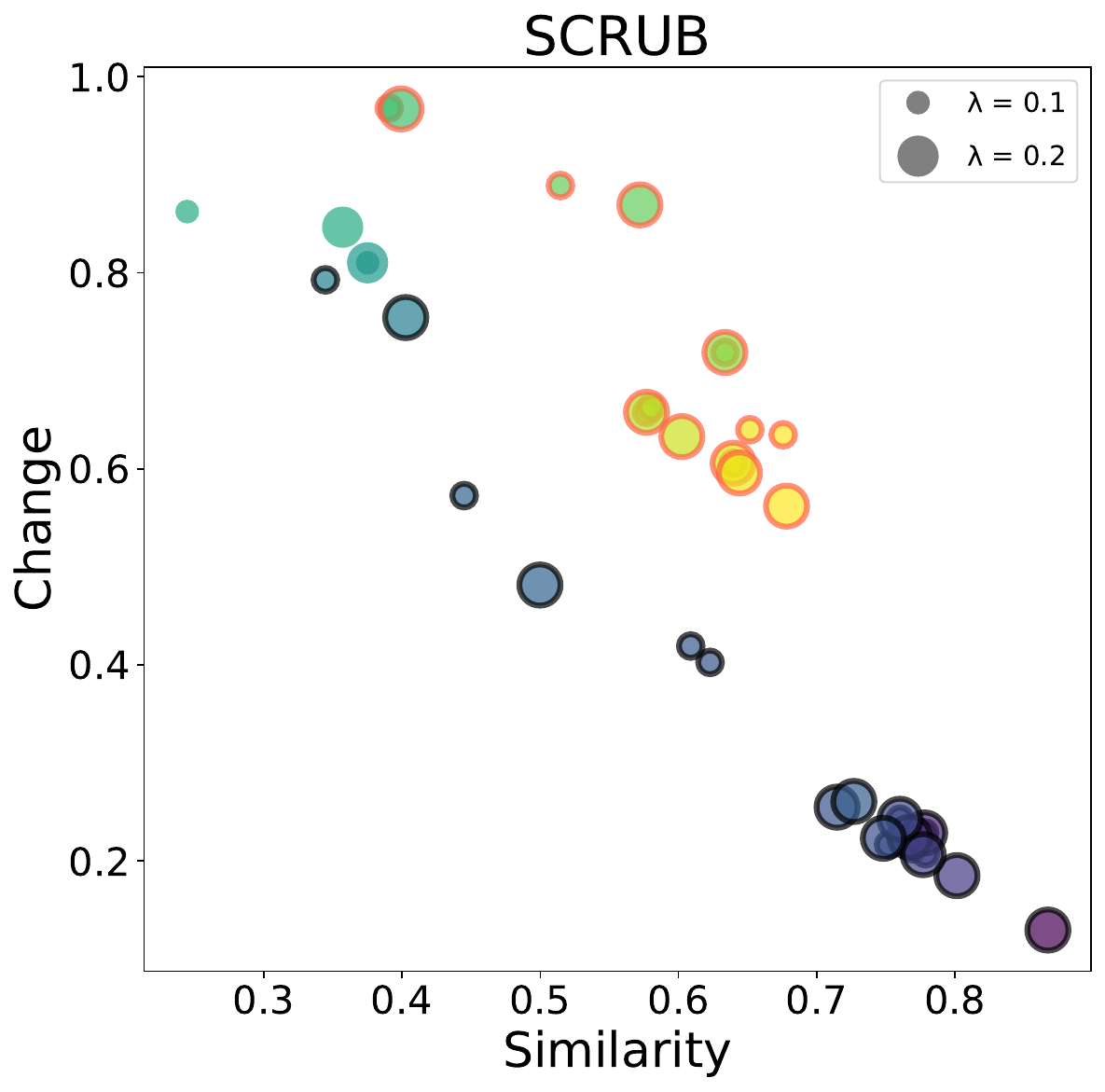}
    \includegraphics[height=0.3\textwidth]{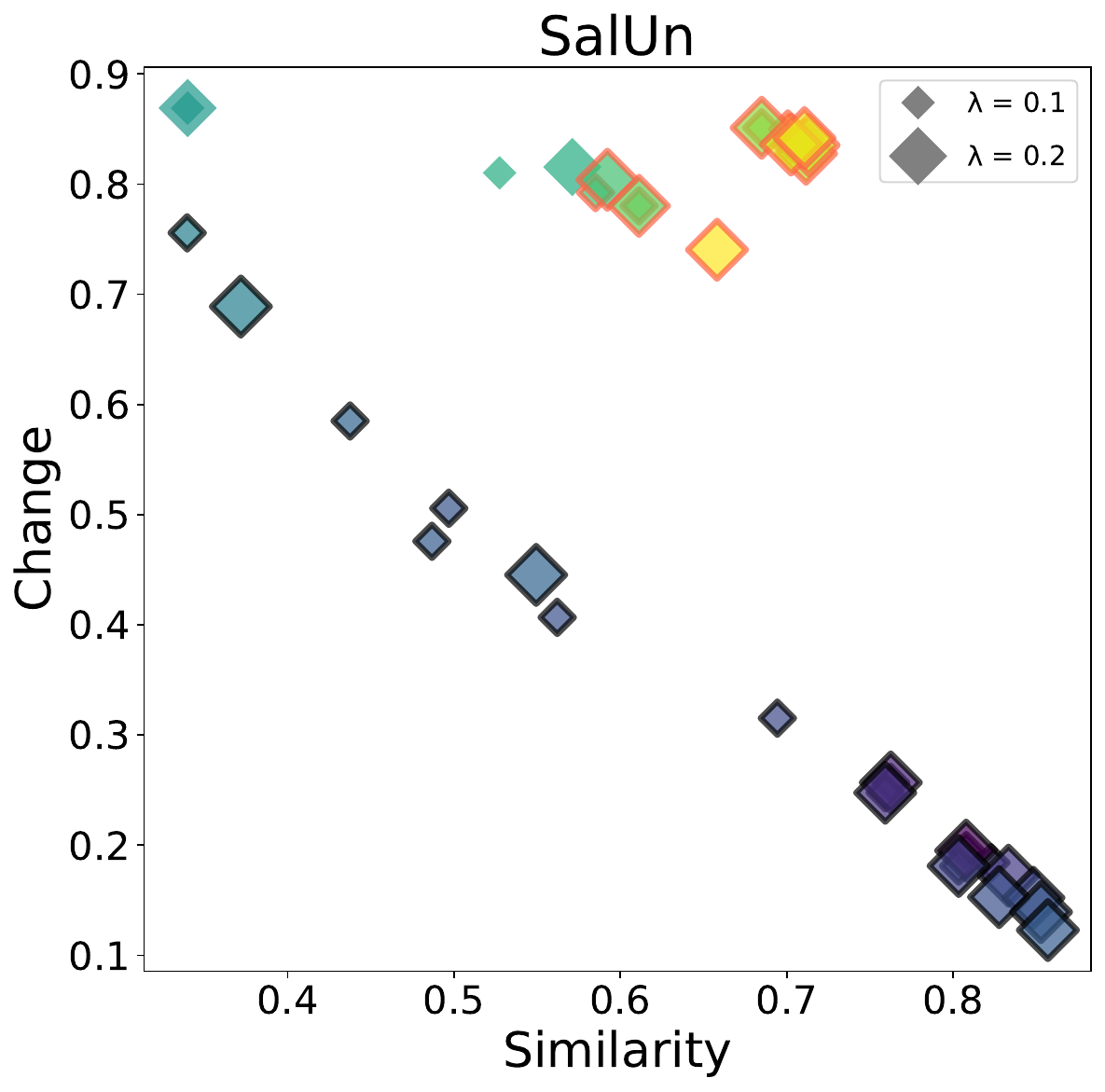}
    \includegraphics[height=0.3\textwidth]{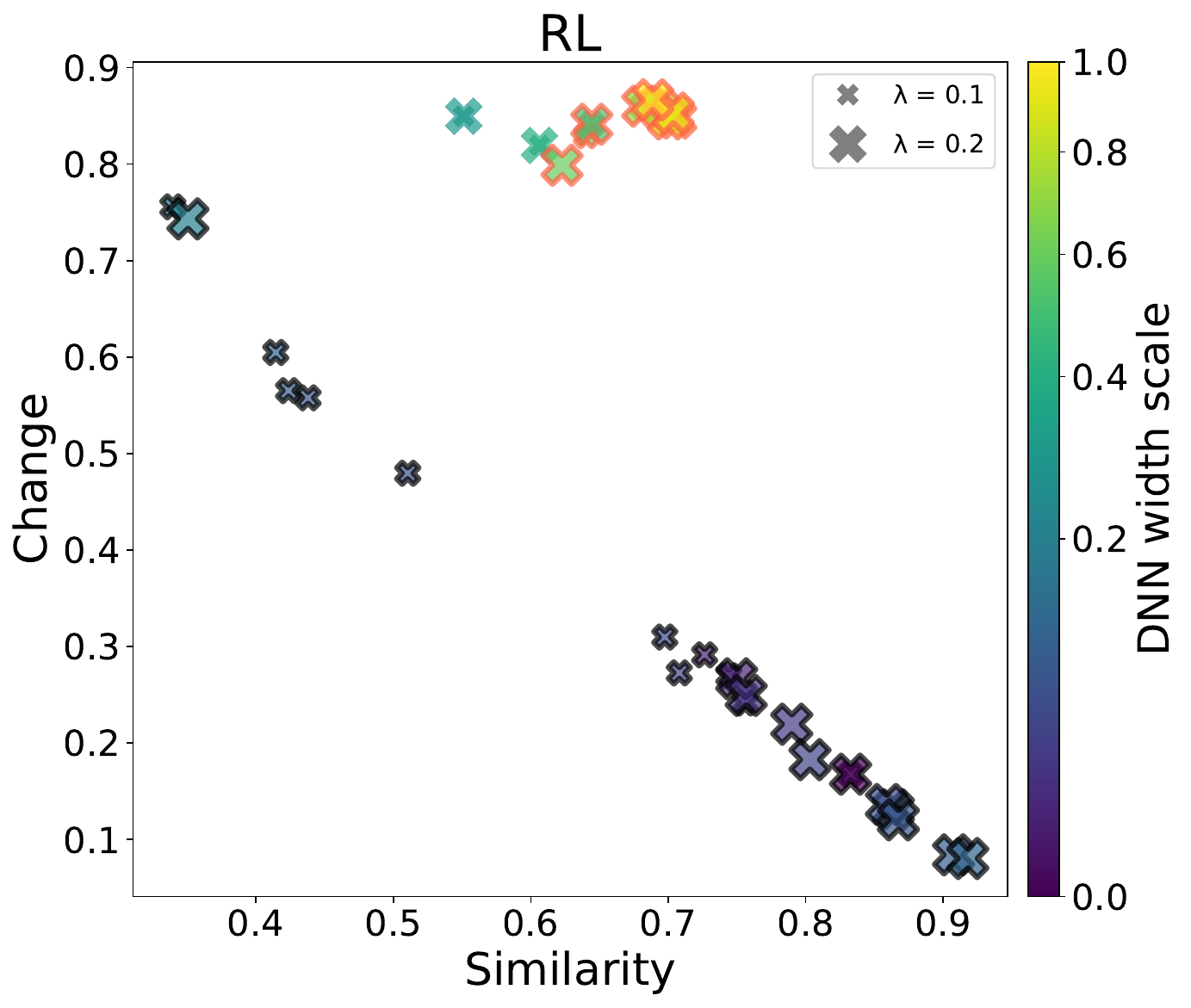}
    \\
    \includegraphics[height=0.3\textwidth]{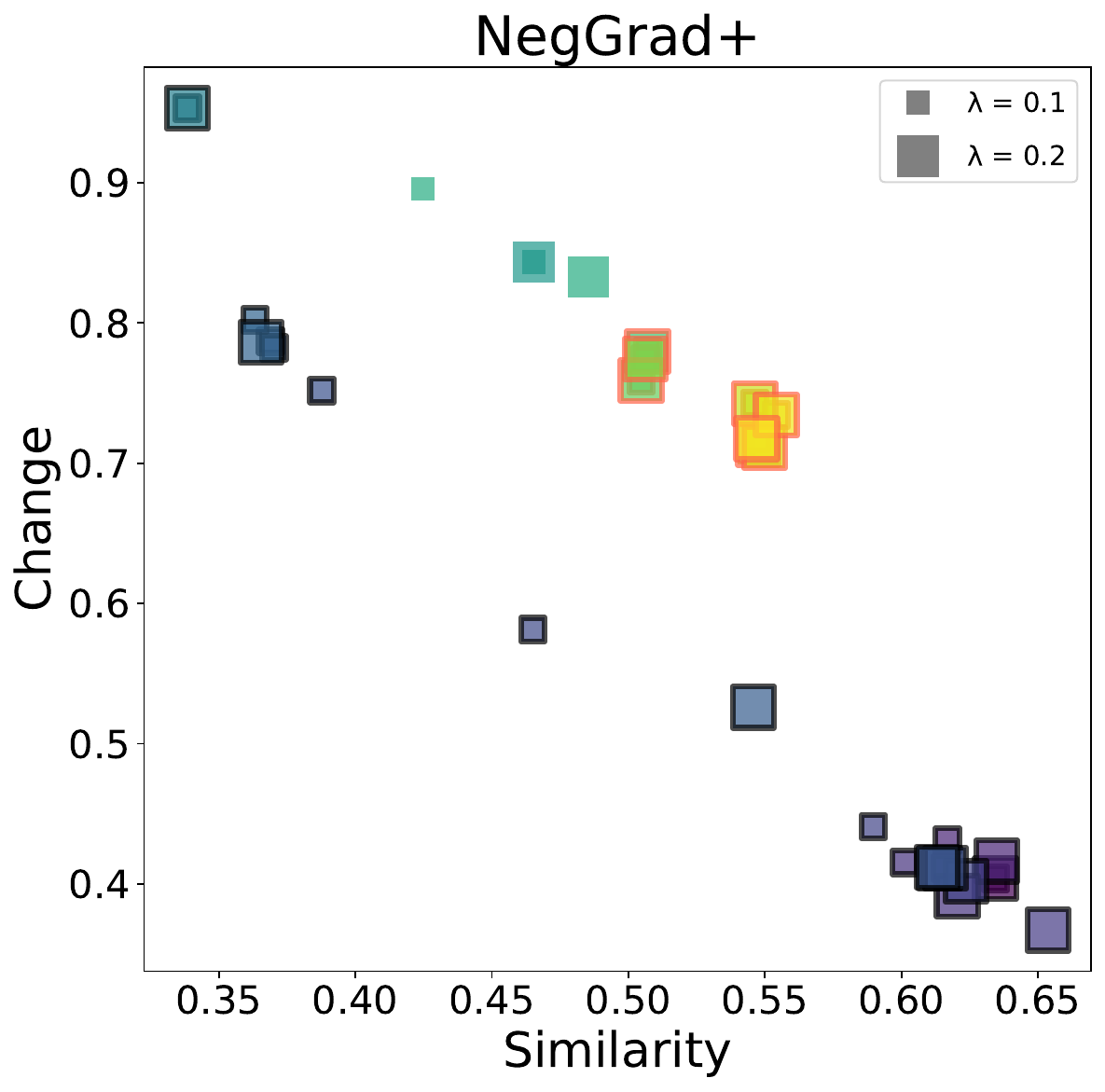}
    \includegraphics[height=0.3\textwidth]{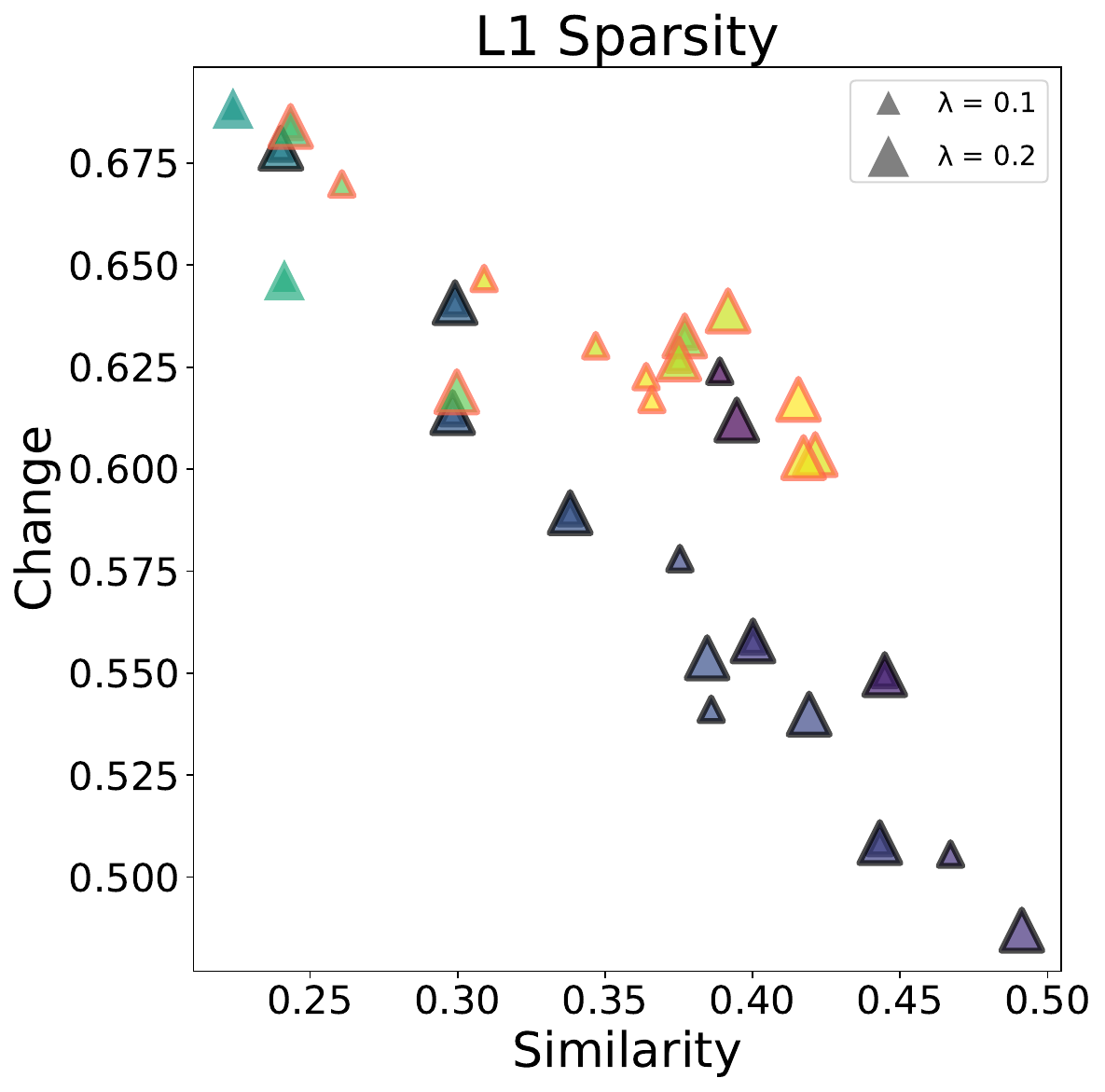}
    \includegraphics[height=0.3\textwidth]{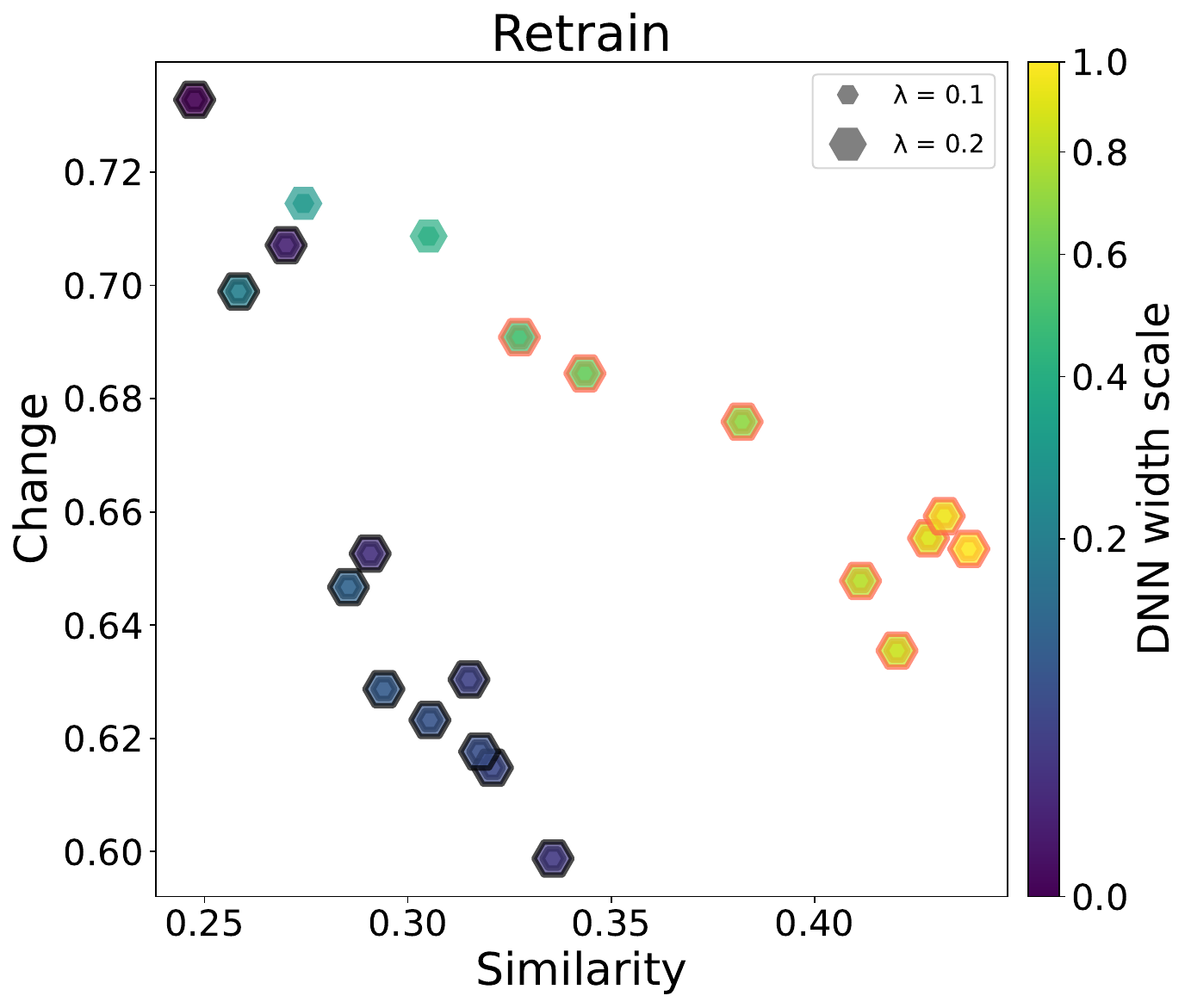}
    \caption{Decision-region similarity and change scores for \textbf{privacy} unlearning. ResNet-18 on Tiny ImageNet (500 unlearned examples, $\delta=10$).  Markers closer to the top-right corner at each diagram denote unlearning with more-local changes of decision regions.}
    \label{fig:similarity_change_privacy_resnet18_tinet_500unlearned}
\end{figure*}

\begin{figure*}[]
    \centering    
    \includegraphics[height=0.3\textwidth]{figures/SC_bias_resnet18_tinet_forget500__scrub_no_colorbar.pdf}
    \includegraphics[height=0.3\textwidth]{figures/SC_bias_resnet18_tinet_forget500__salun_no_colorbar.pdf}
    \includegraphics[height=0.3\textwidth]{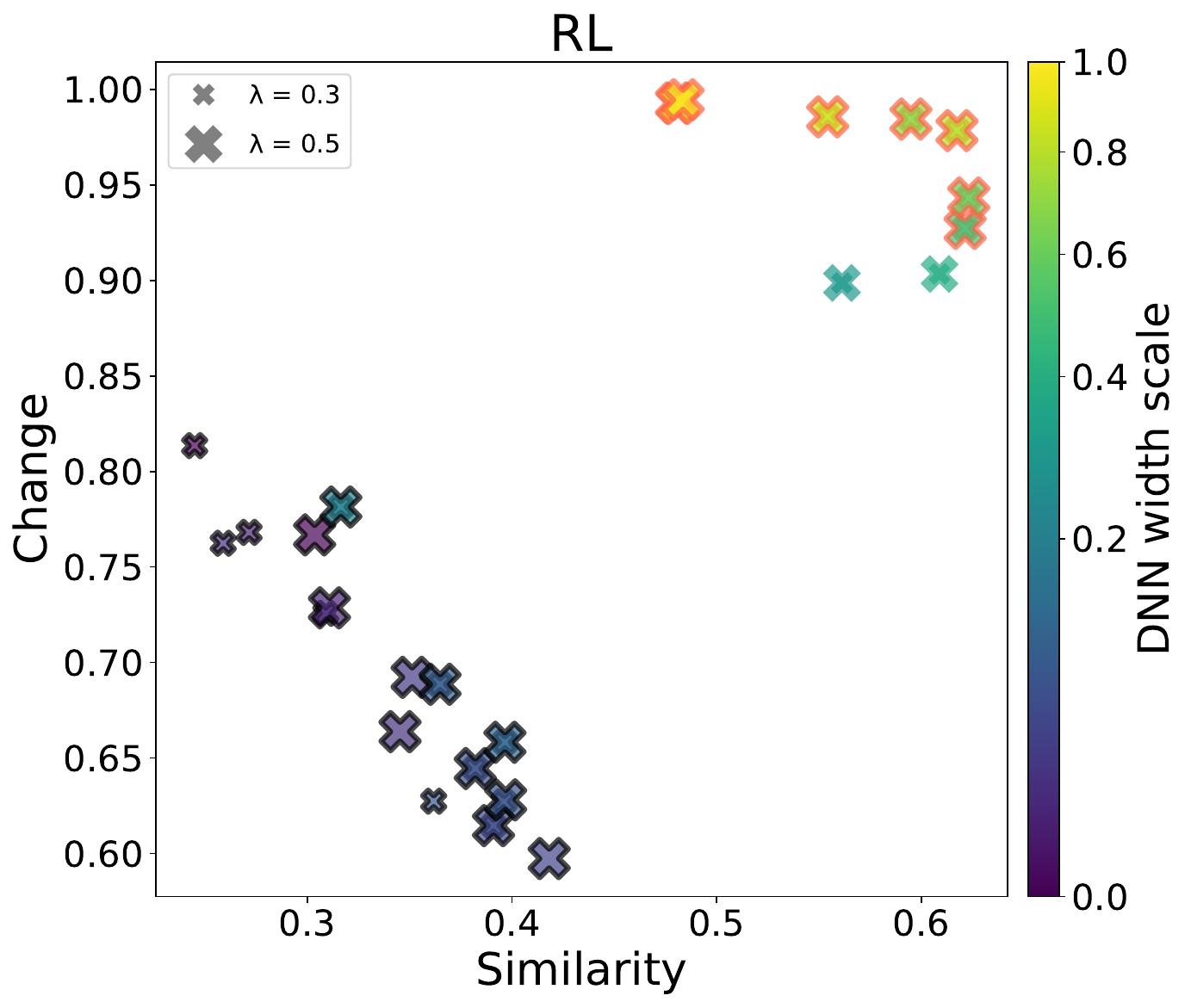}
    \\
    \includegraphics[height=0.3\textwidth]{figures/SC_bias_resnet18_tinet_forget500__negGrad_no_colorbar.pdf}
    \includegraphics[height=0.3\textwidth]{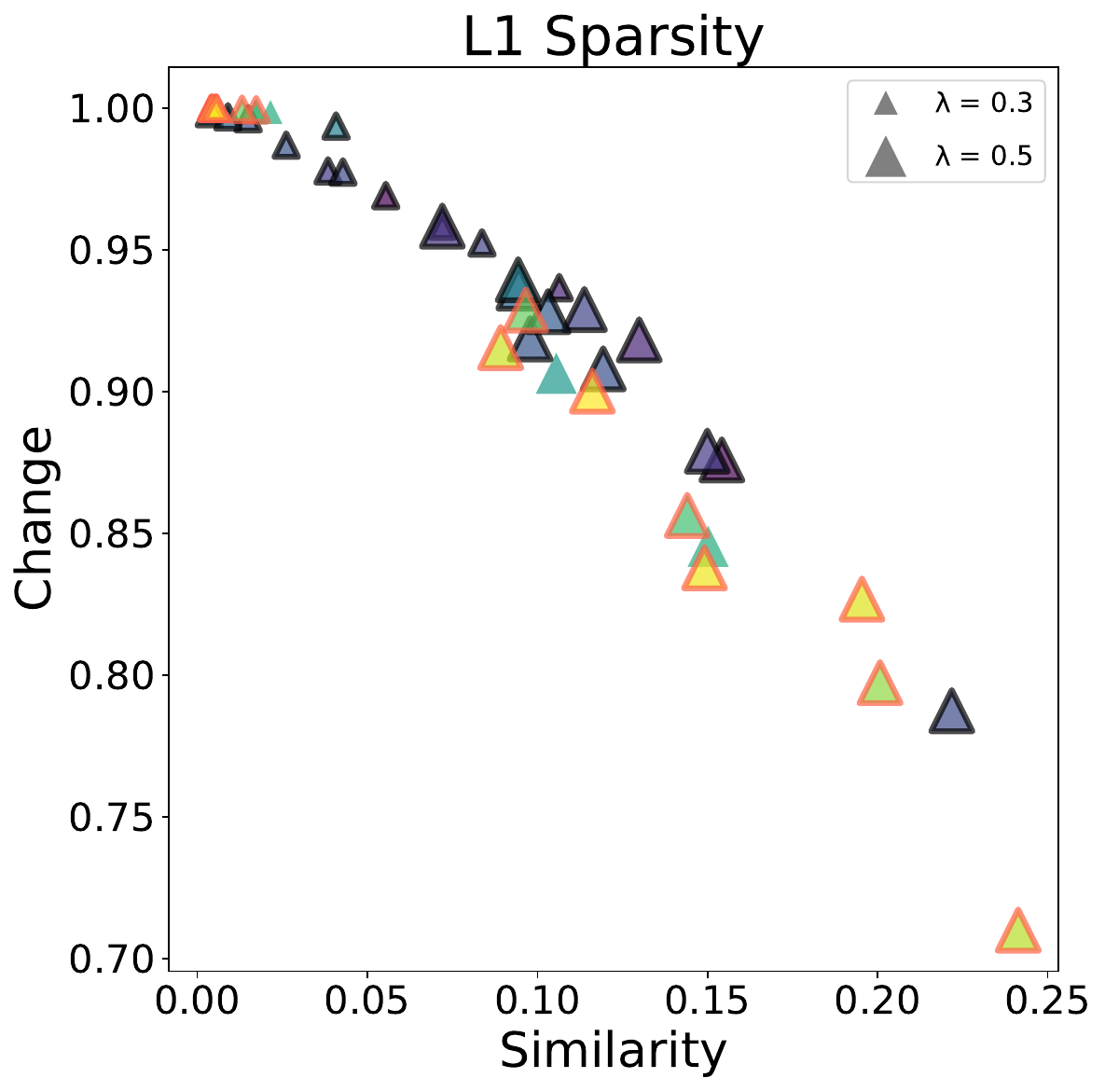}
    \includegraphics[height=0.3\textwidth]{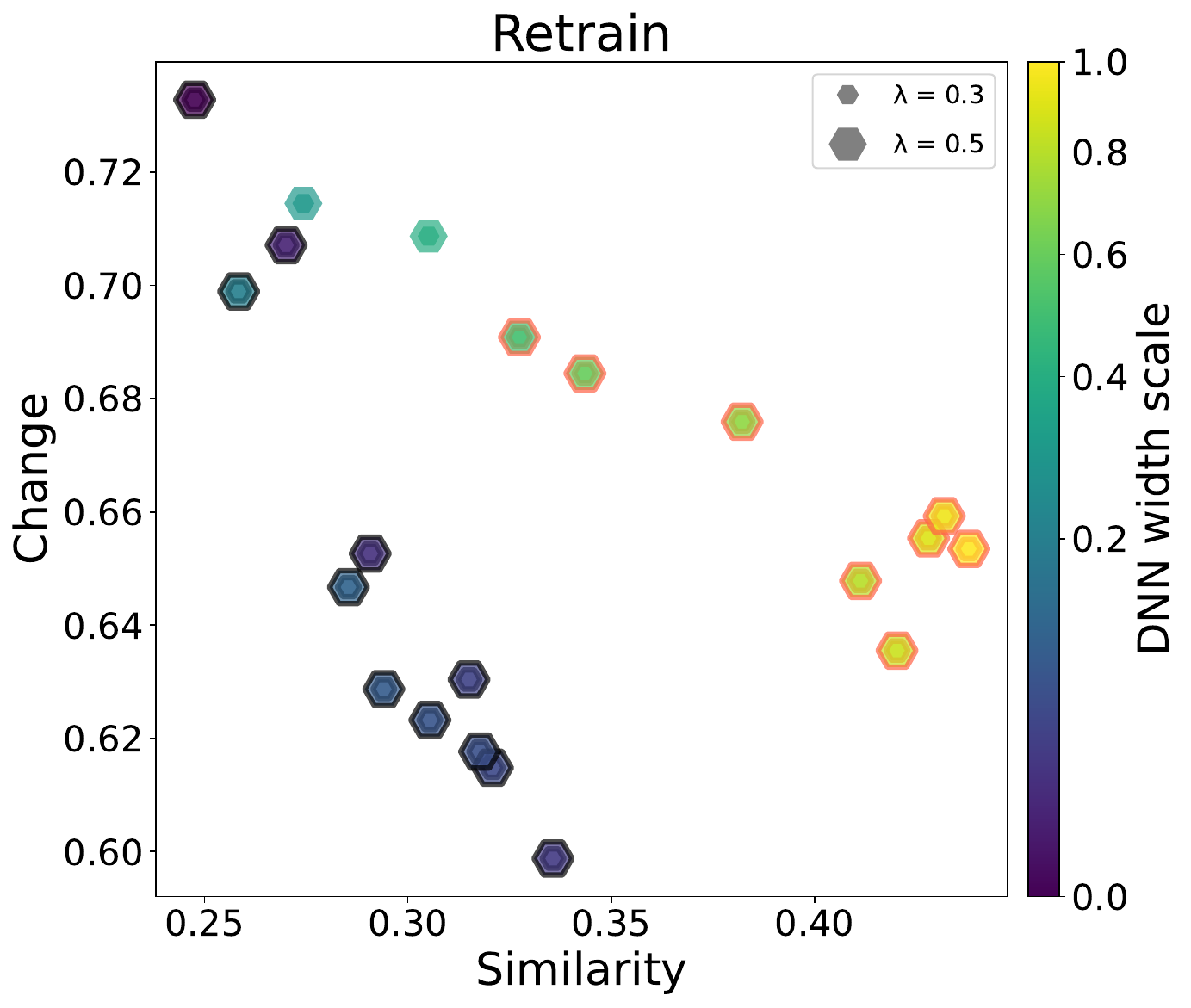}
    \caption{Decision-region similarity and change scores for \textbf{bias removal} unlearning. ResNet-18 on Tiny ImageNet (500 unlearned examples, $\delta=10$).  Markers closer to the top-right corner at each diagram denote unlearning with more-local changes of decision regions.}
    \label{fig:similarity_change_bias_resnet18_tinet_500unlearned}
\end{figure*}

\begin{figure*}[]
    \centering    
    \includegraphics[height=0.3\textwidth]{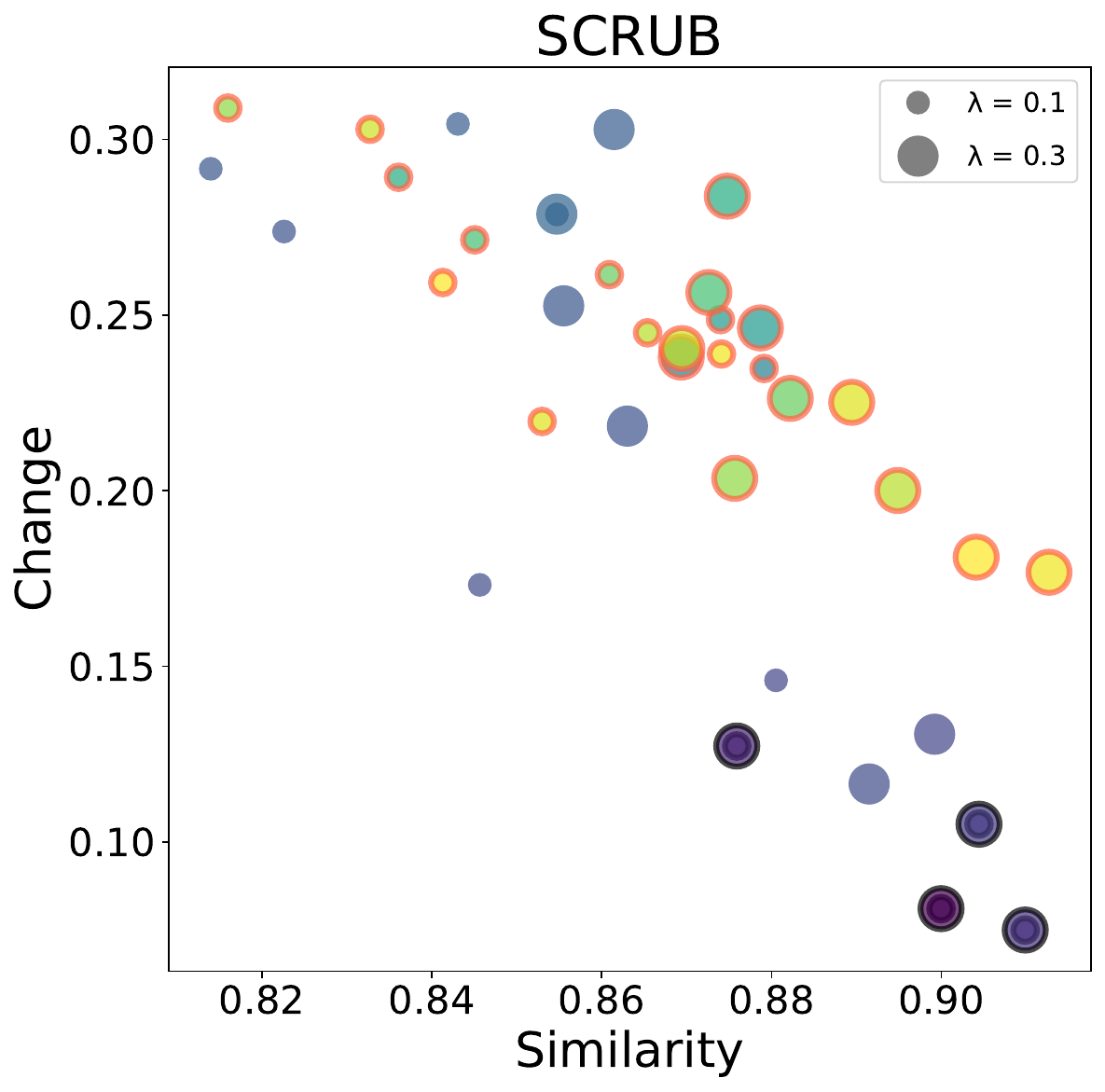}
    \includegraphics[height=0.3\textwidth]{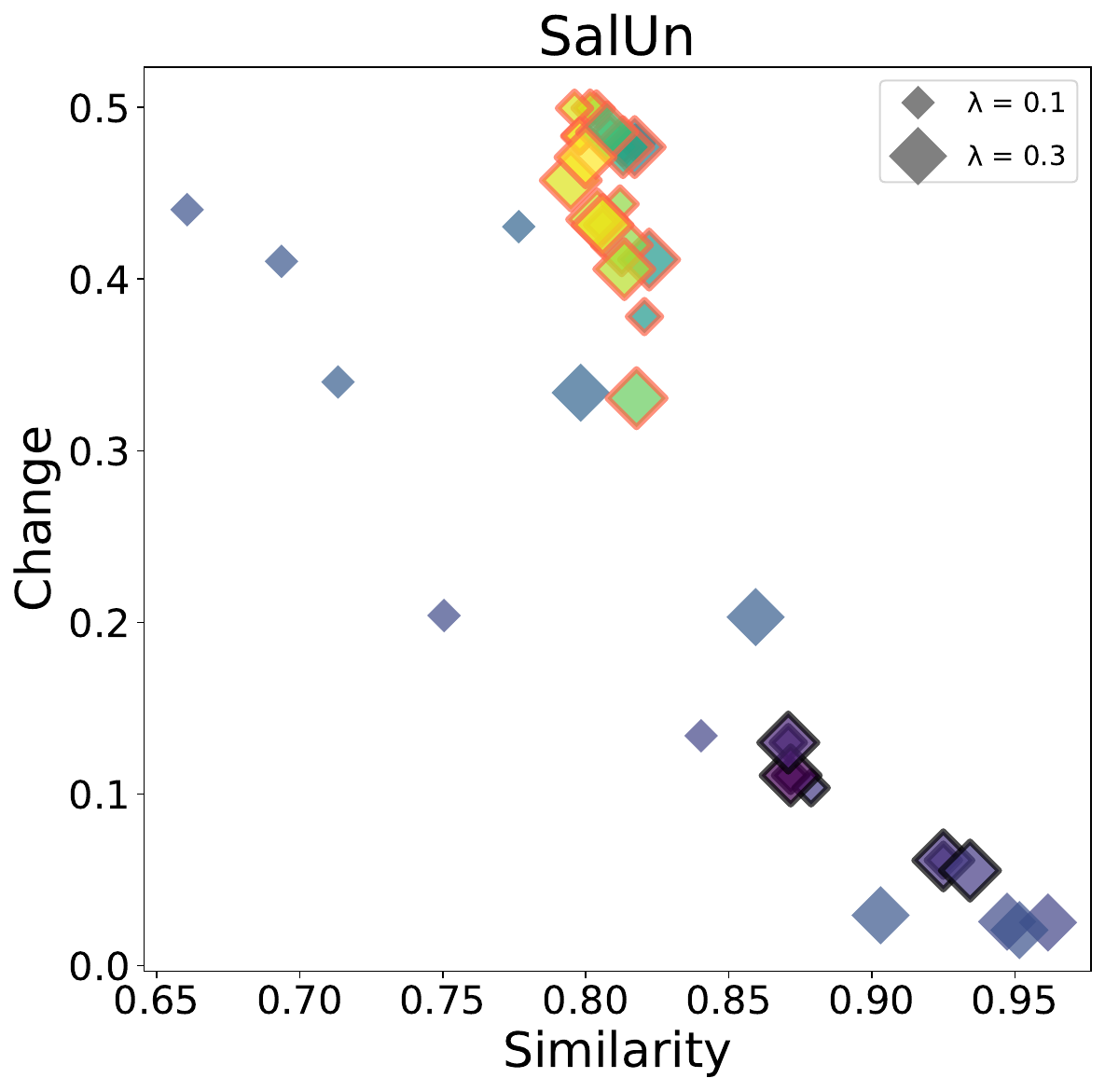}
    \includegraphics[height=0.3\textwidth]{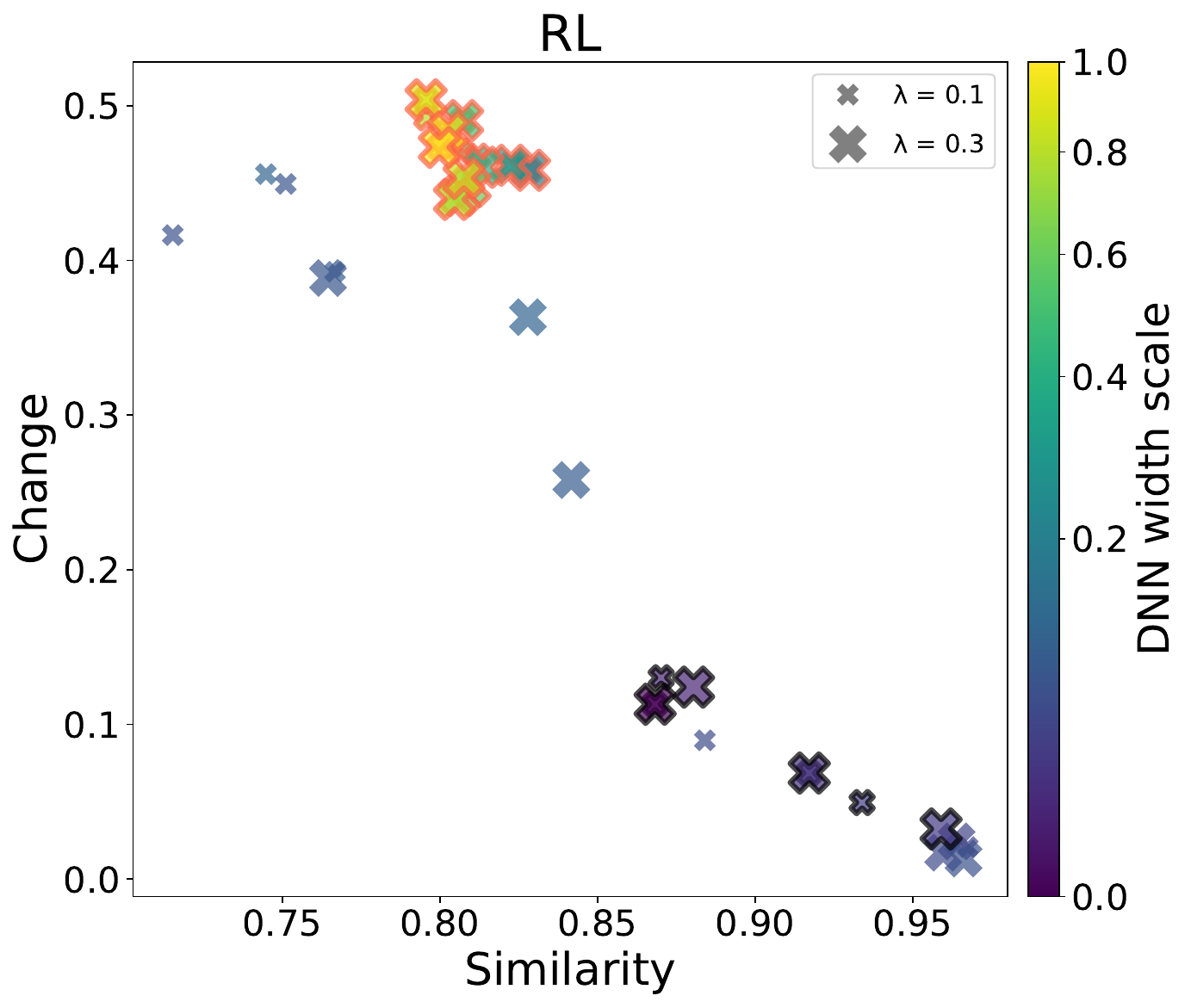}
    \\
    \includegraphics[height=0.3\textwidth]{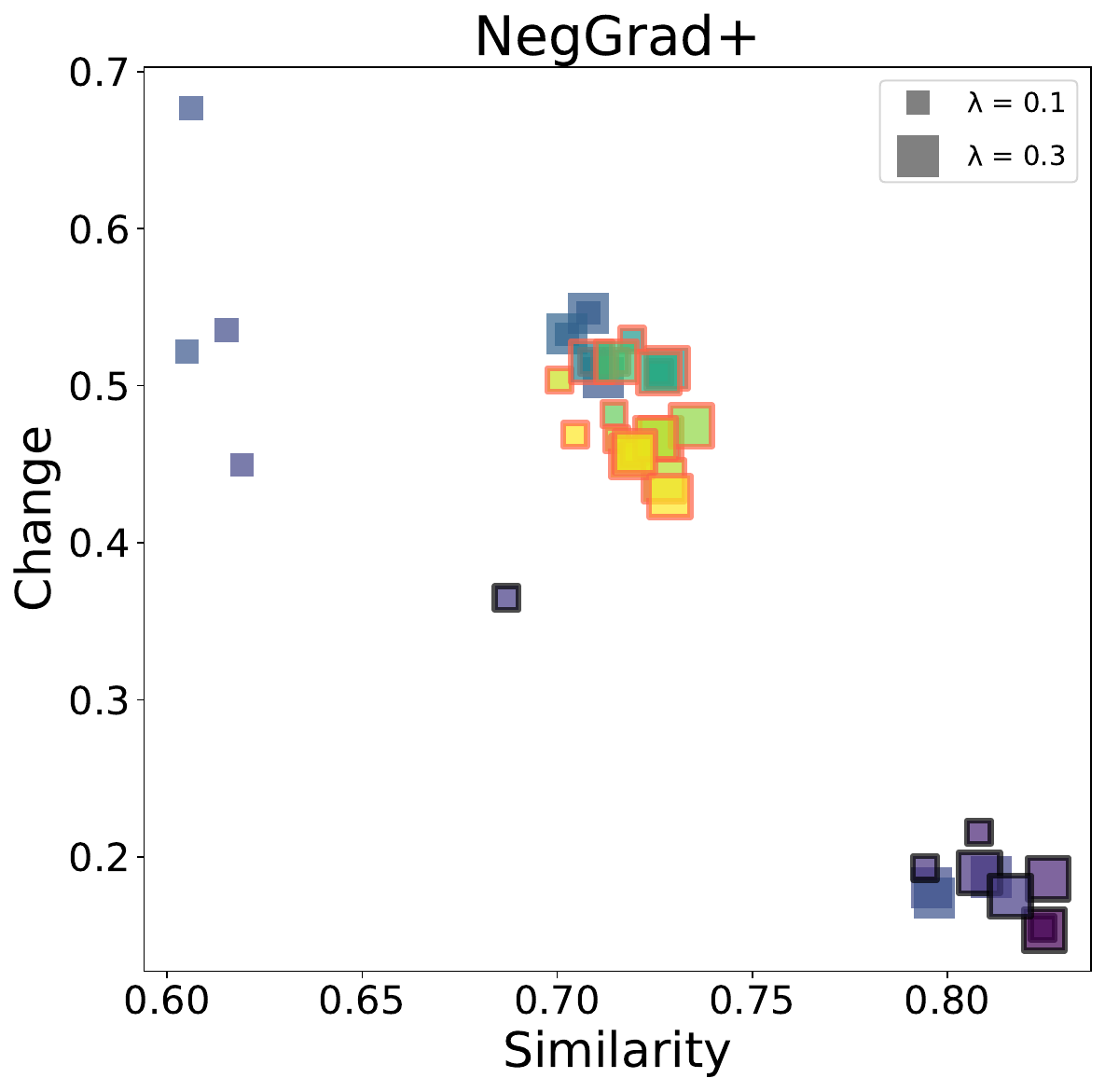}
    \includegraphics[height=0.3\textwidth]{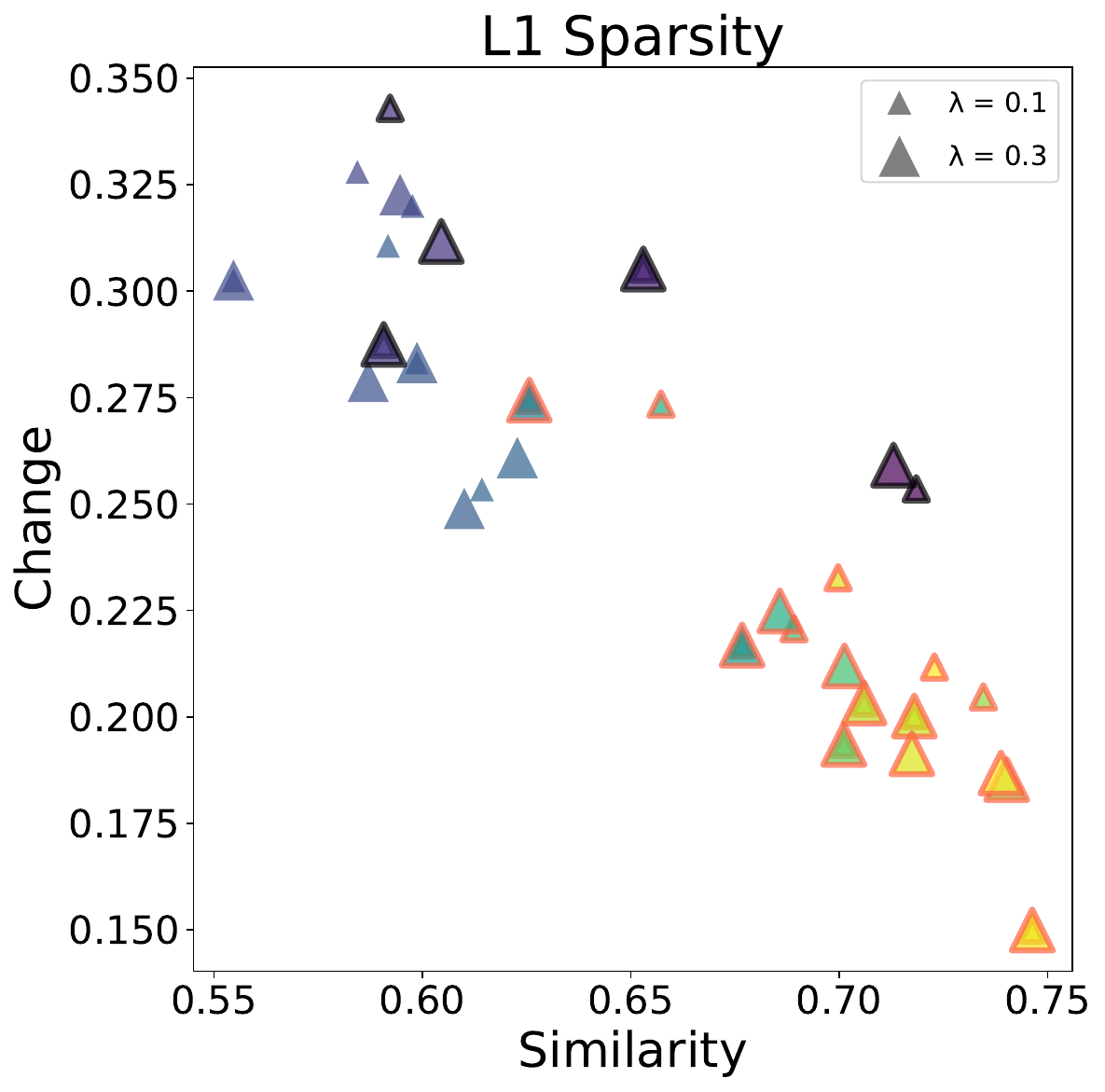}
    \includegraphics[height=0.3\textwidth]{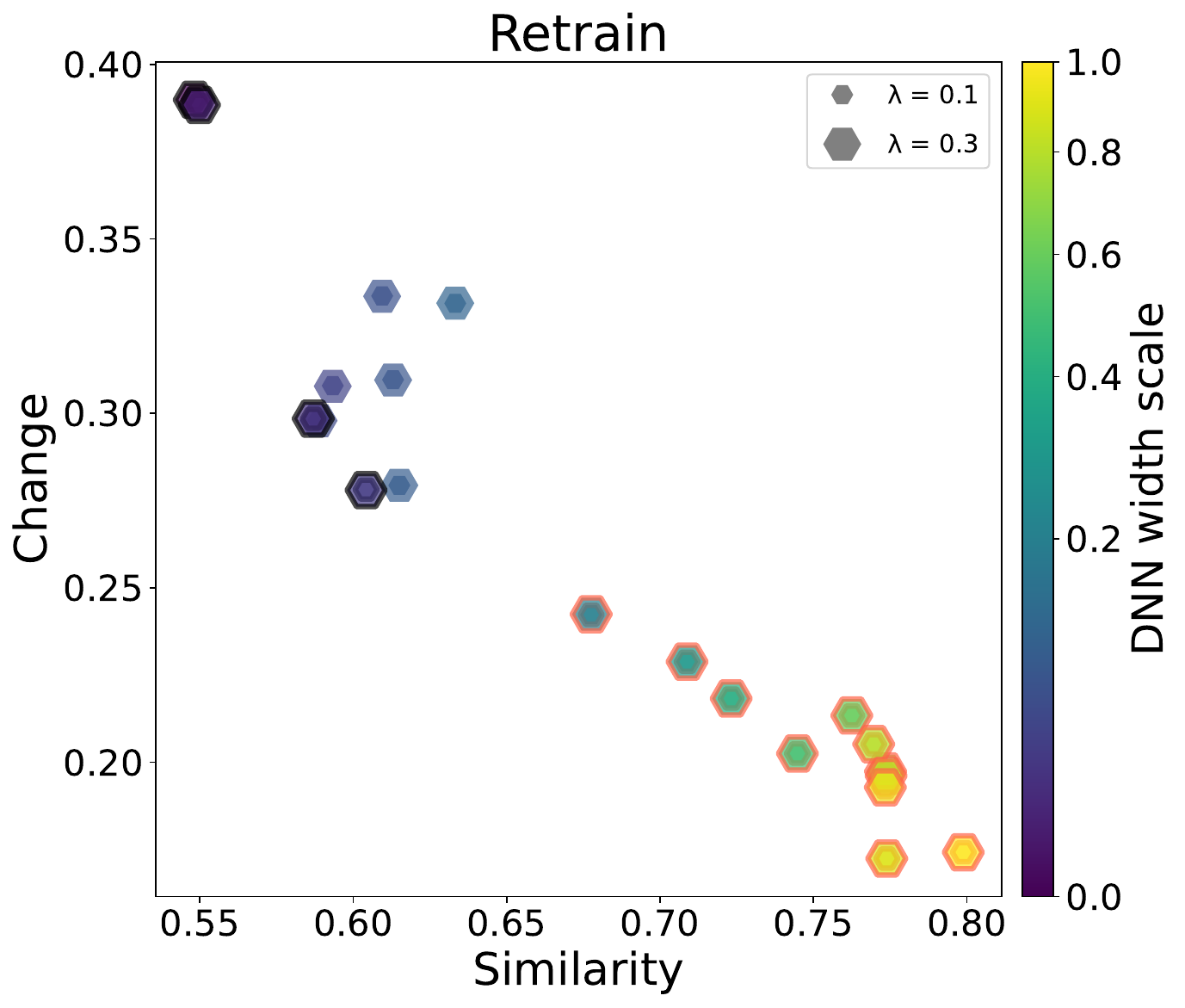}
    \caption{Decision-region similarity and change scores for \textbf{privacy} unlearning. ResNet-18 on CIFAR-10 (200 unlearned examples, $\delta=10$).  Markers closer to the top-right corner at each diagram denote unlearning with more-local changes of decision regions.}
    \label{fig:similarity_change_privacy_resnet_cifar10_200unlearned}
\end{figure*}

\begin{figure*}[]
    \centering    
    \includegraphics[height=0.3\textwidth]{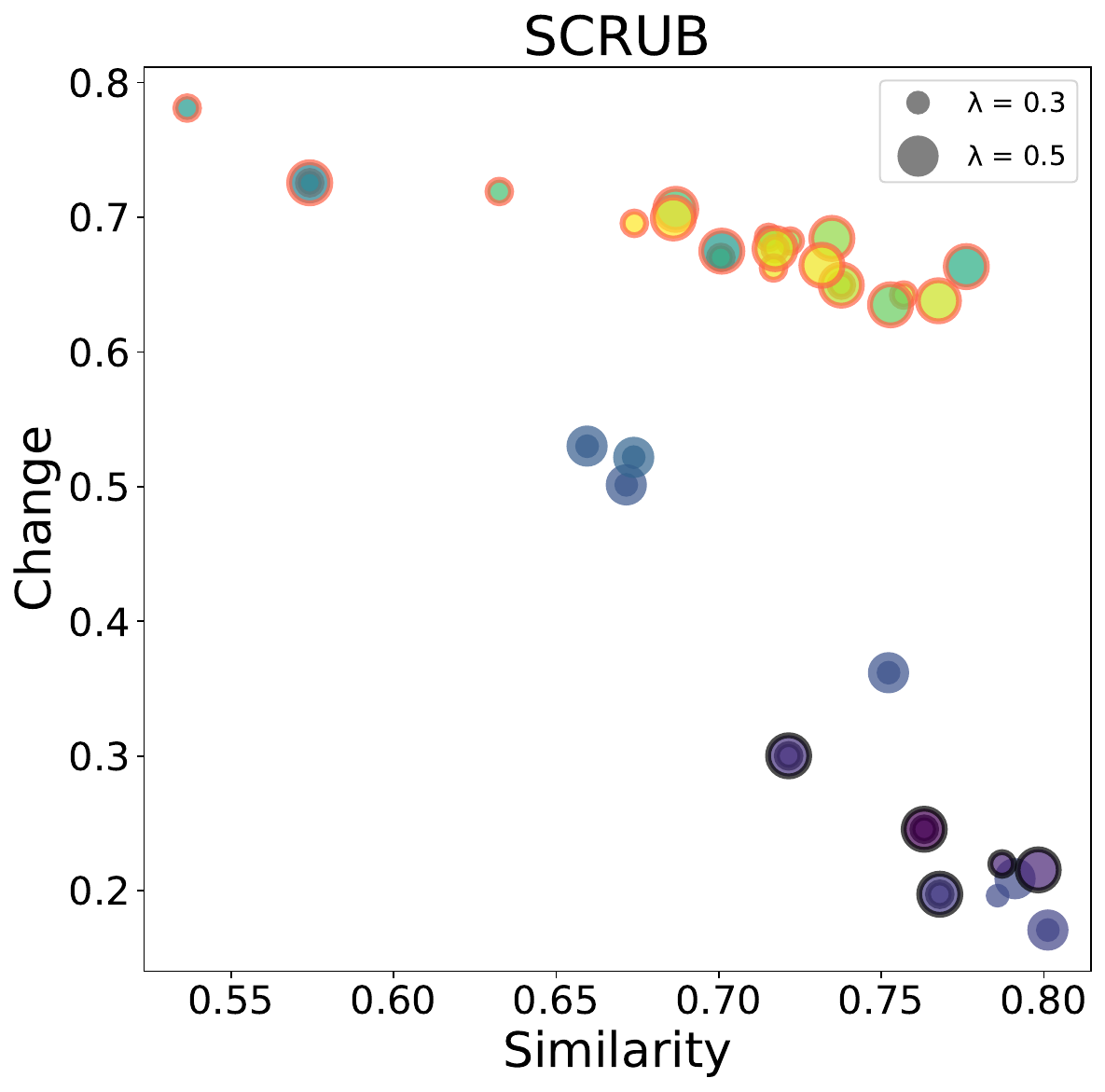}
    \includegraphics[height=0.3\textwidth]{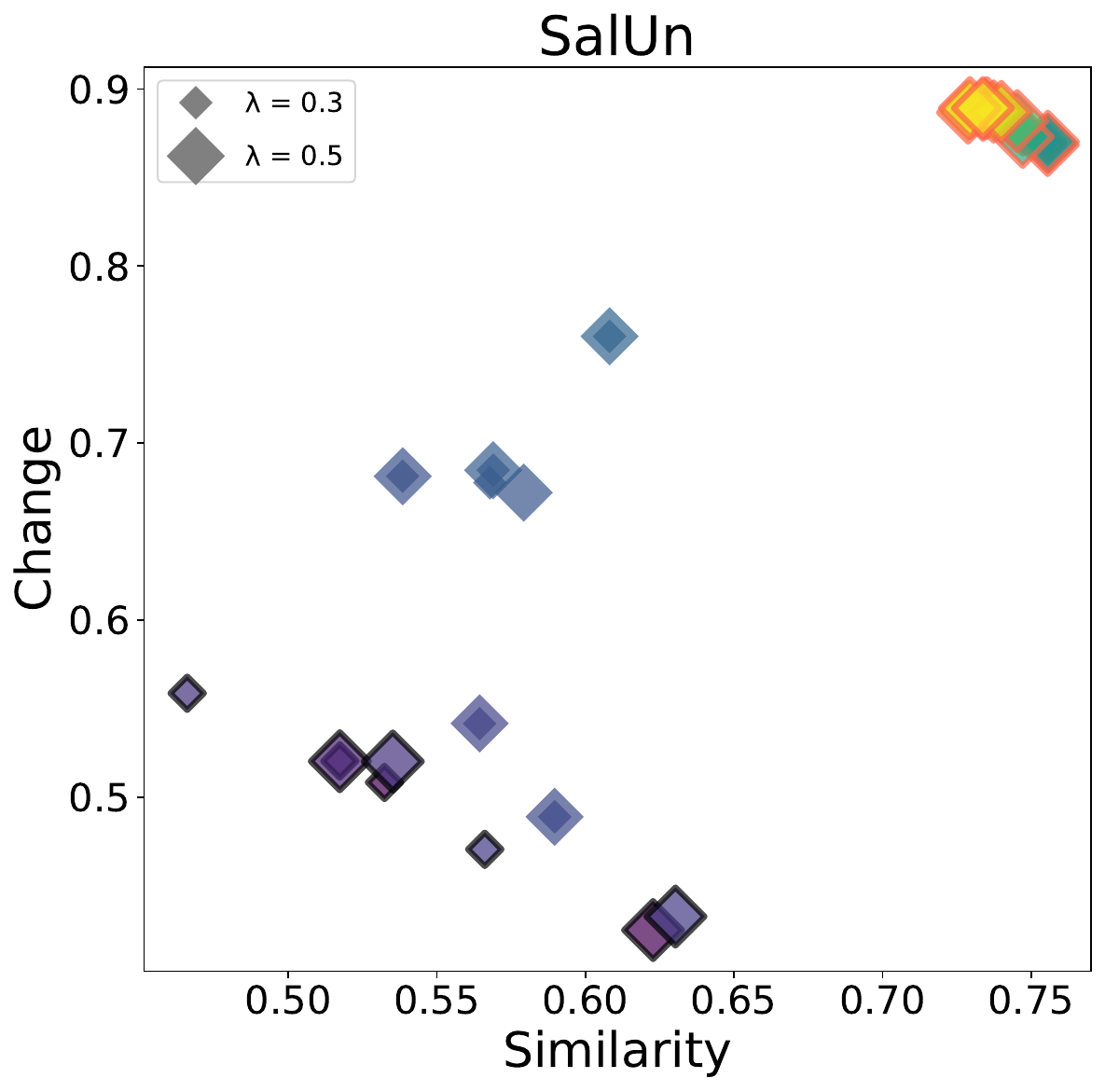}
    \includegraphics[height=0.3\textwidth]{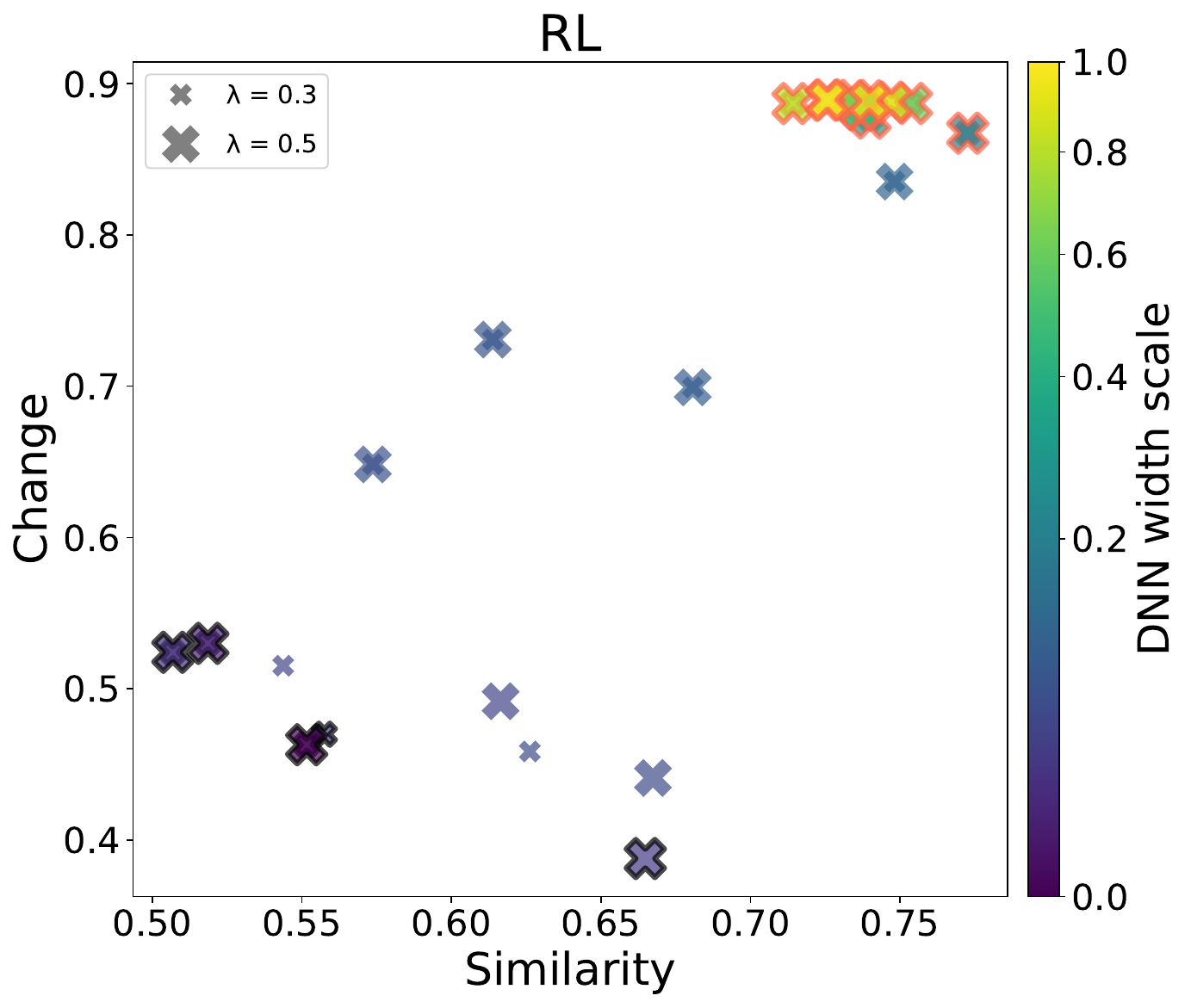}
    \\
    \includegraphics[height=0.3\textwidth]{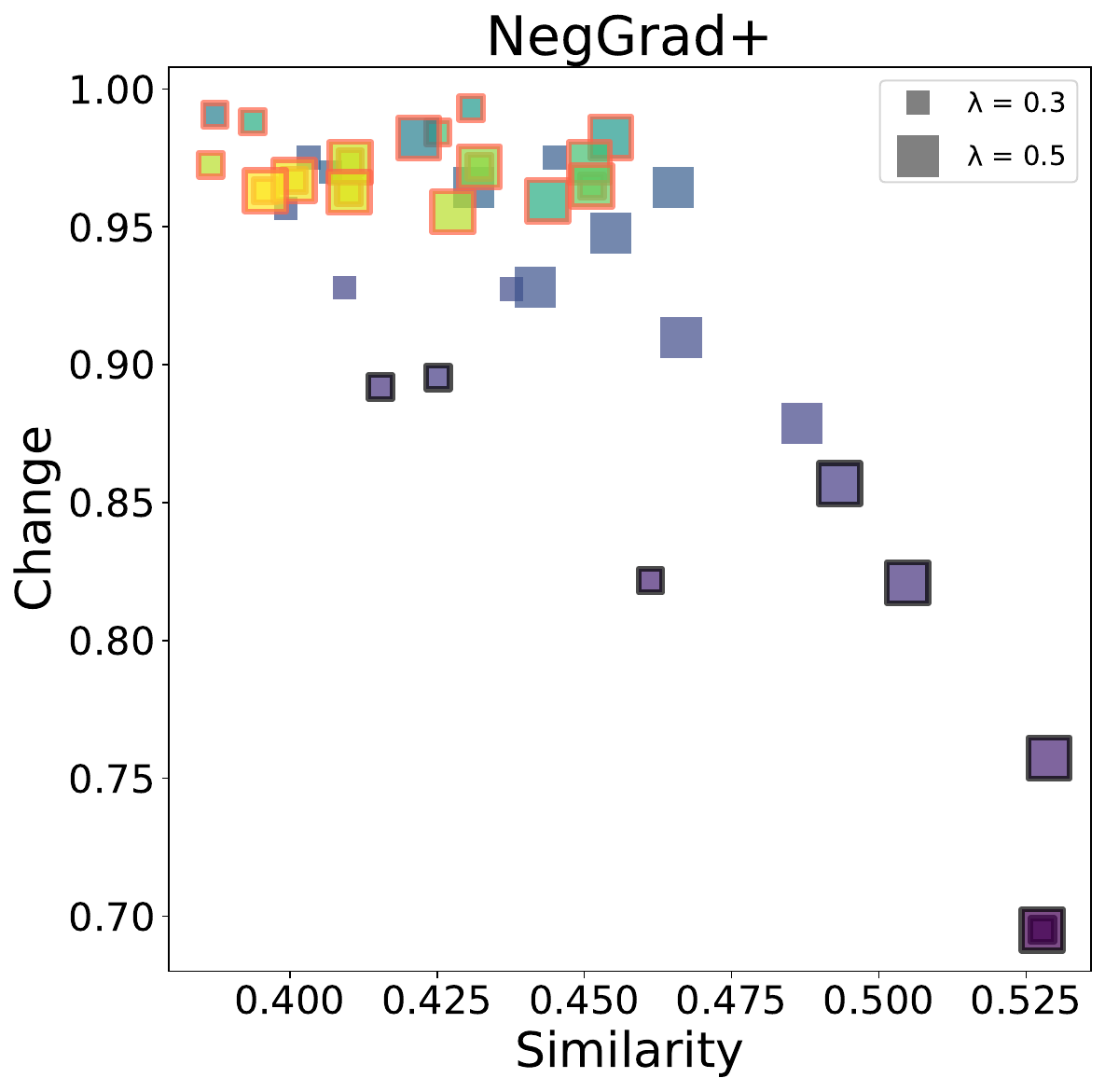}
    \includegraphics[height=0.3\textwidth]{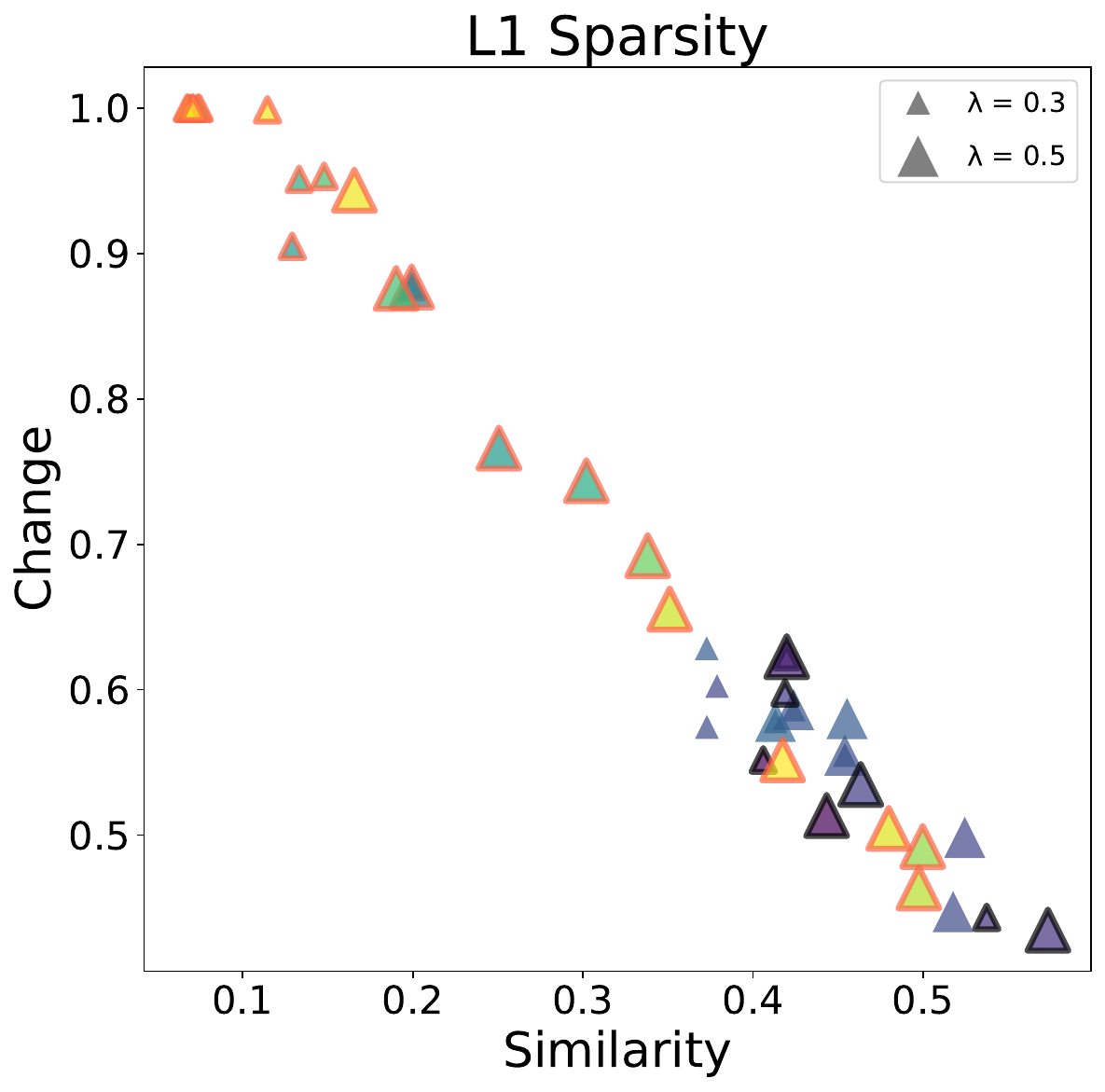}
    \includegraphics[height=0.3\textwidth]{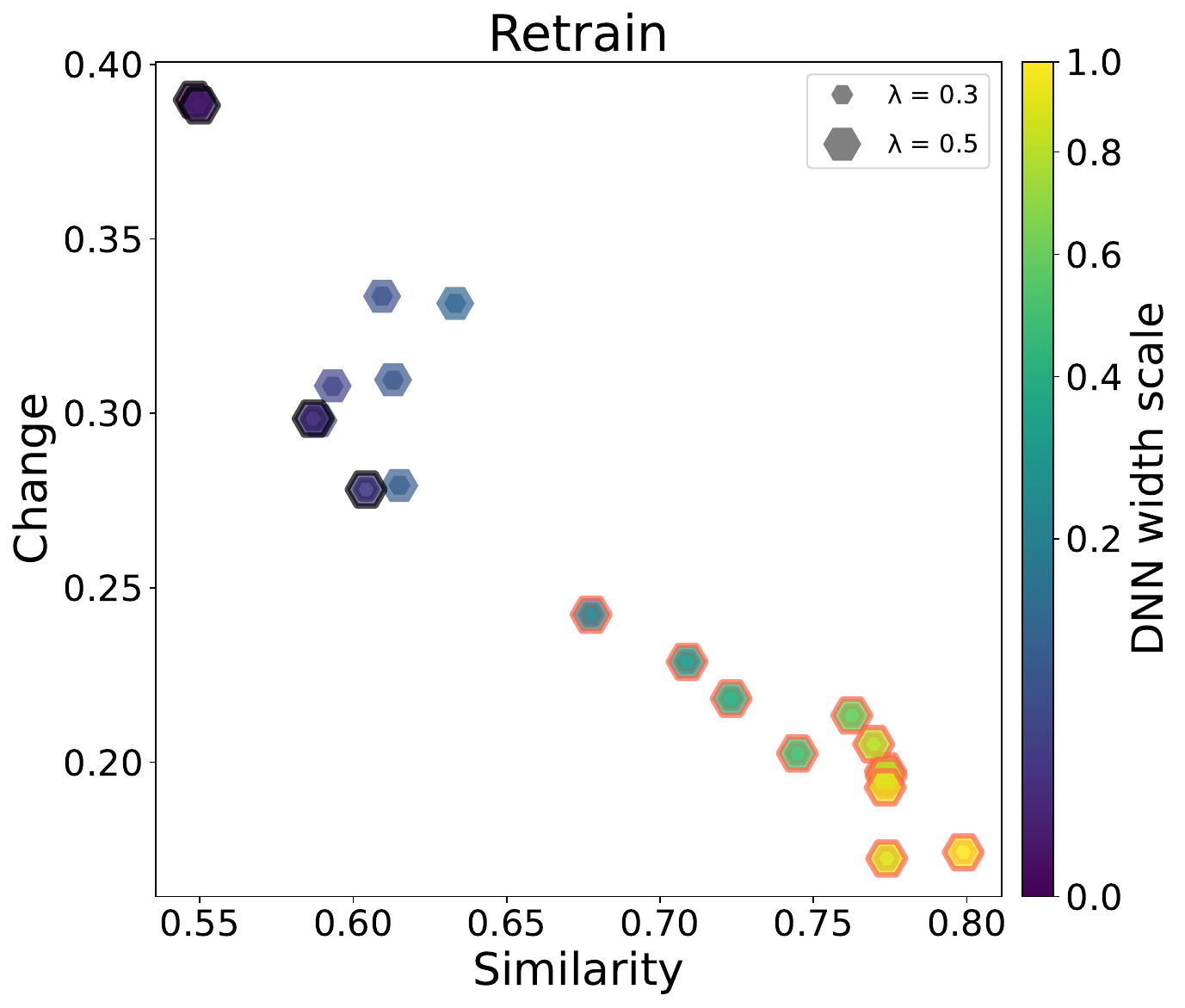}
    \caption{Decision-region similarity and change scores for \textbf{bias removal} unlearning. ResNet-18 on CIFAR-10 (200 unlearned examples, $\delta=10$).  Markers closer to the top-right corner at each diagram denote unlearning with more-local changes of decision regions.}
    \label{fig:similarity_change_bias_resnet_cifar10_200unlearned}
\end{figure*}

\begin{figure*}[]
    \centering    
    \includegraphics[height=0.3\textwidth]{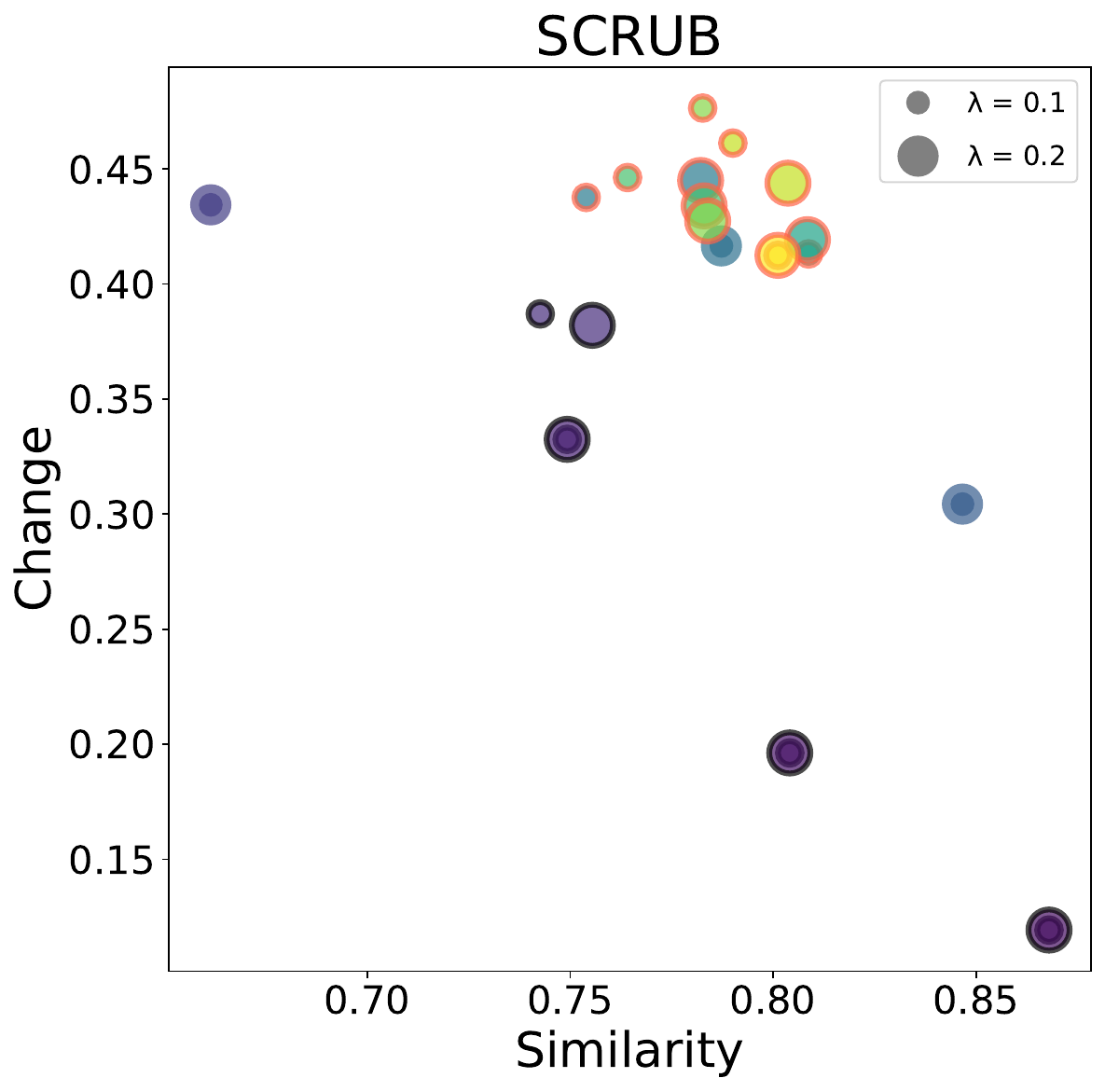}
    \includegraphics[height=0.3\textwidth]{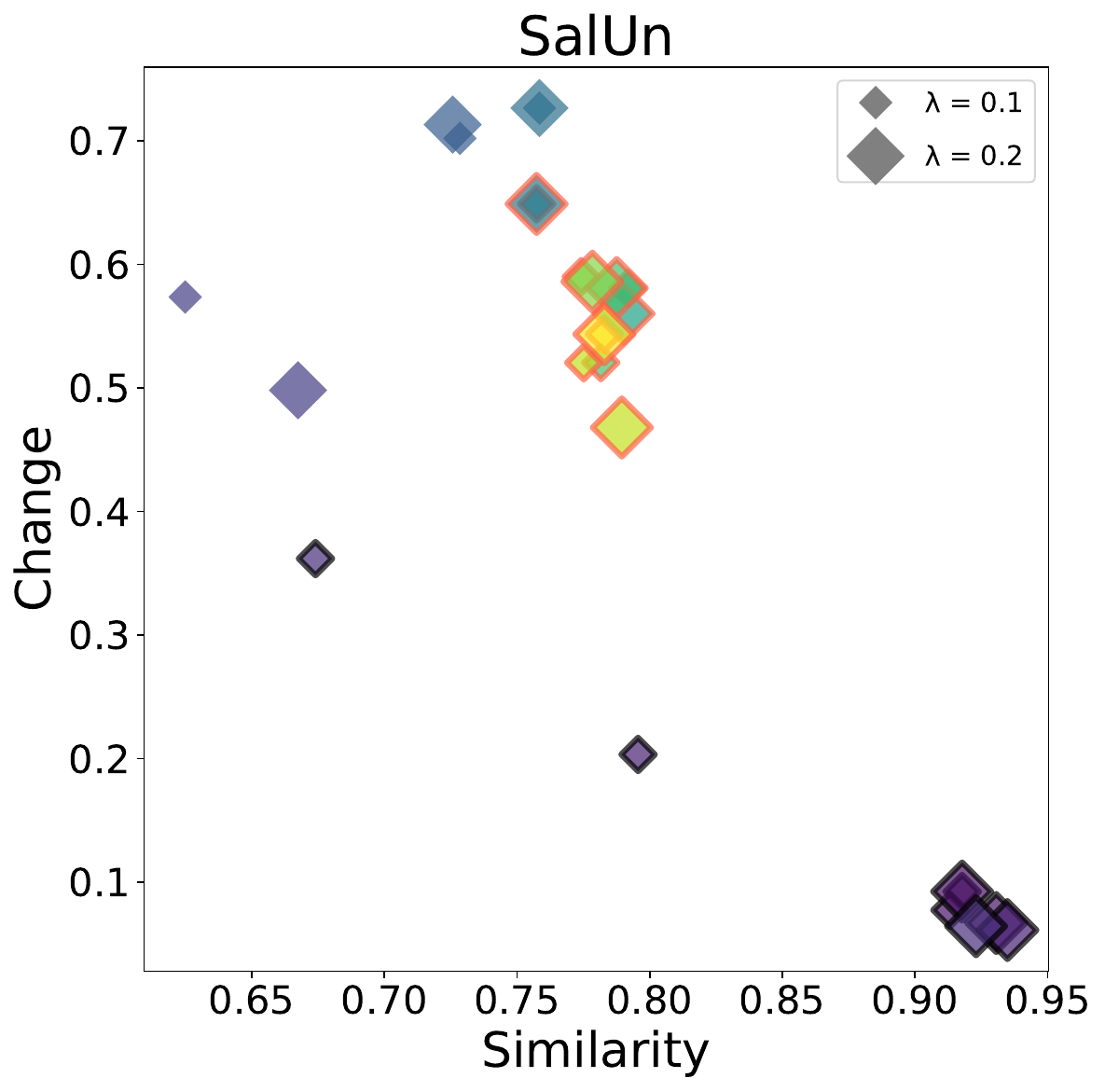}
    \includegraphics[height=0.3\textwidth]{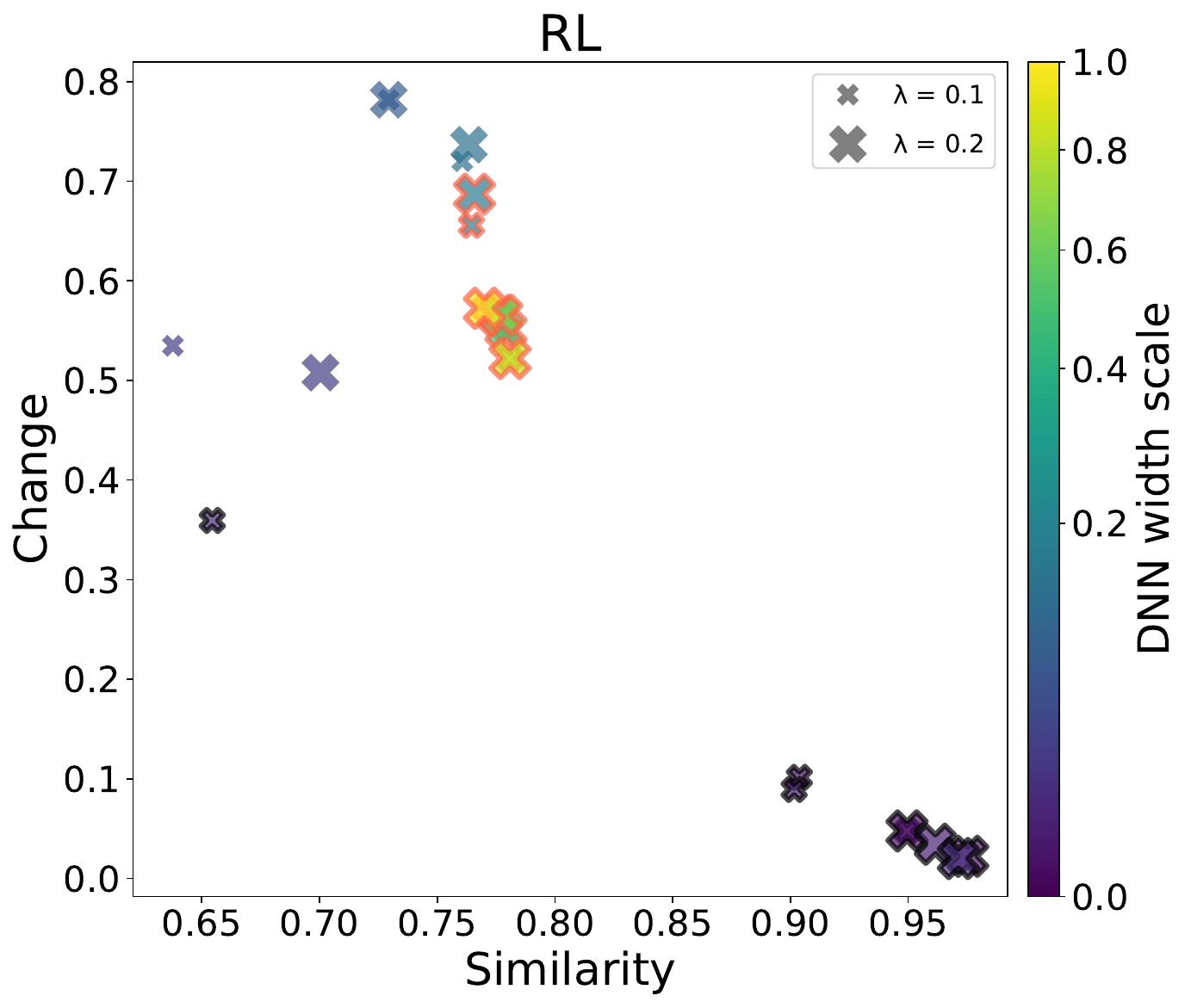}
    \\
    \includegraphics[height=0.3\textwidth]{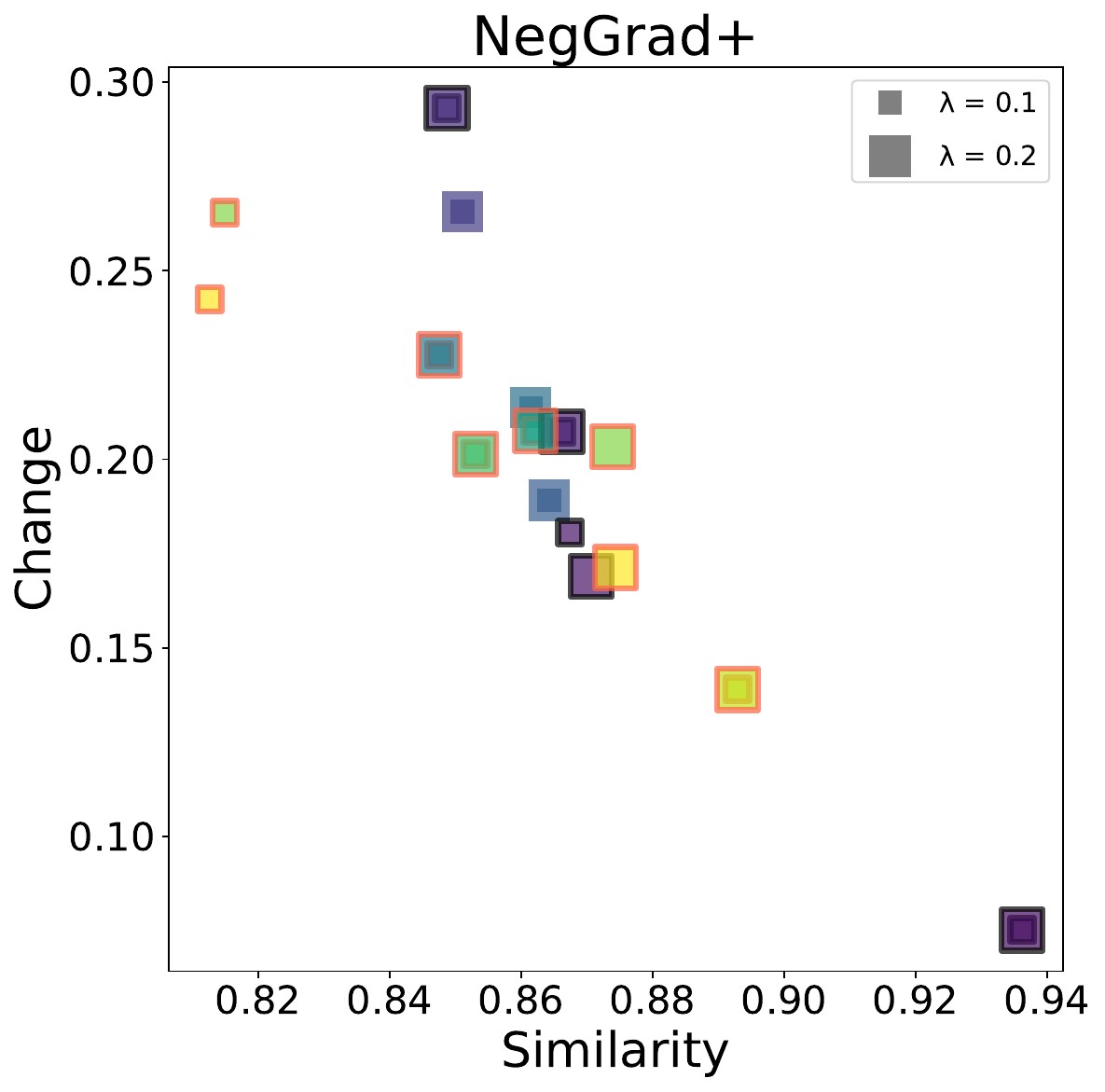}
    \includegraphics[height=0.3\textwidth]{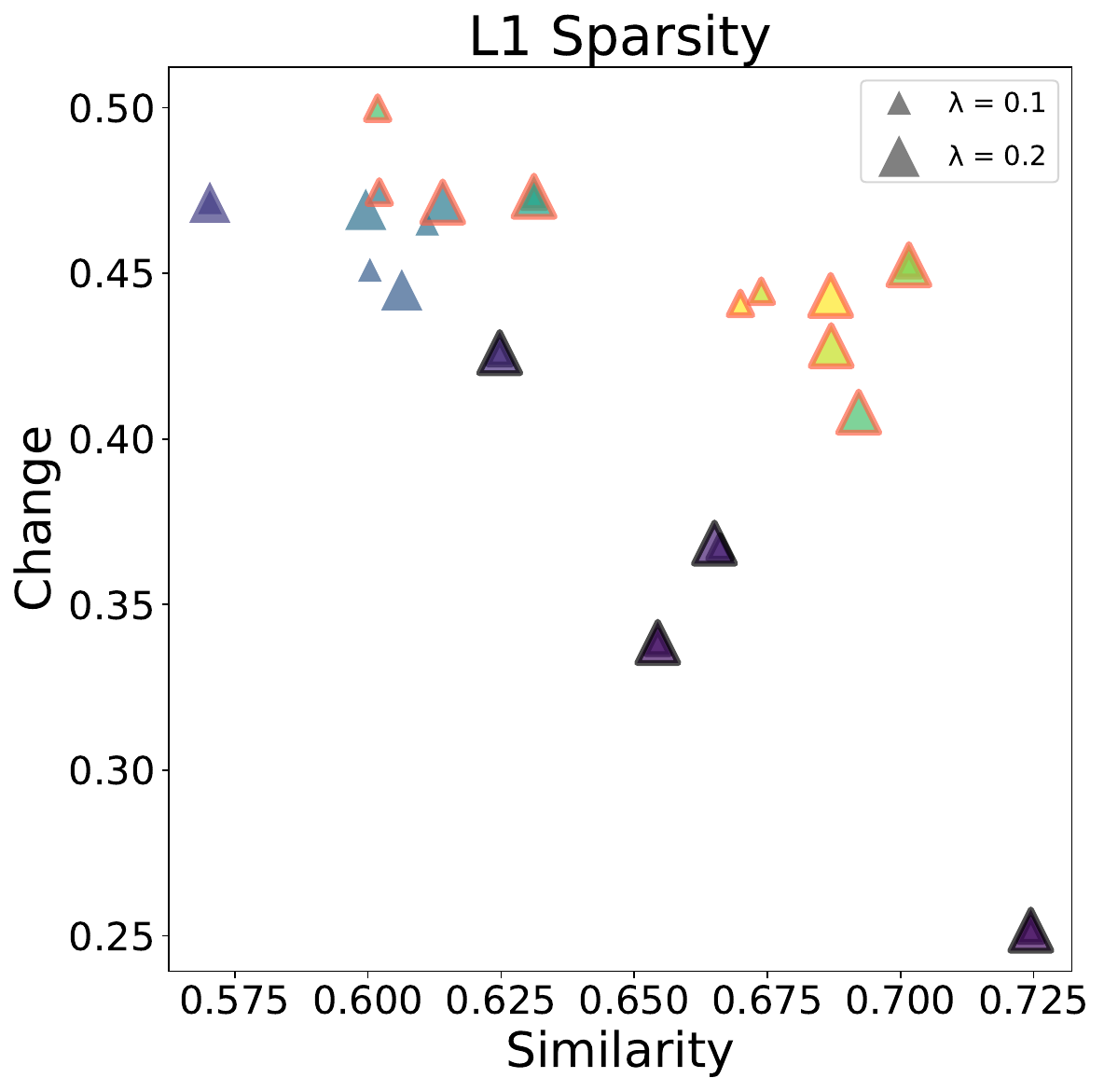}
    \includegraphics[height=0.3\textwidth]{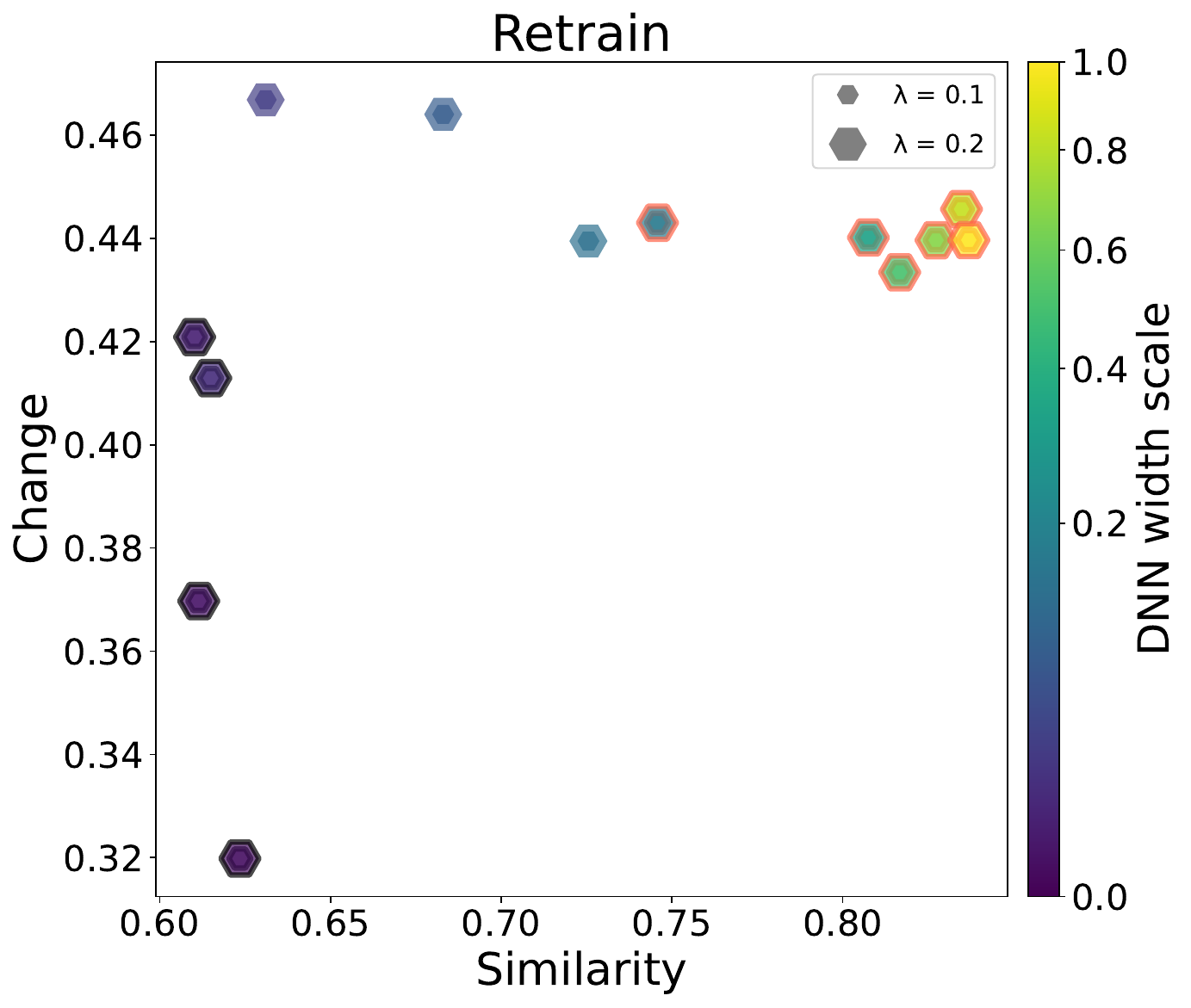}
    \caption{Decision-region similarity and change scores for \textbf{privacy} unlearning. 3-layer FC network on CIFAR-10 (200 unlearned examples, $\delta=10$).  Markers closer to the top-right corner at each diagram denote unlearning with more-local changes of decision regions.}
    \label{fig:similarity_change_privacy_fcnet3layer_cifar10_200unlearned}
\end{figure*}

\begin{figure*}[]
    \centering    
    \includegraphics[height=0.3\textwidth]{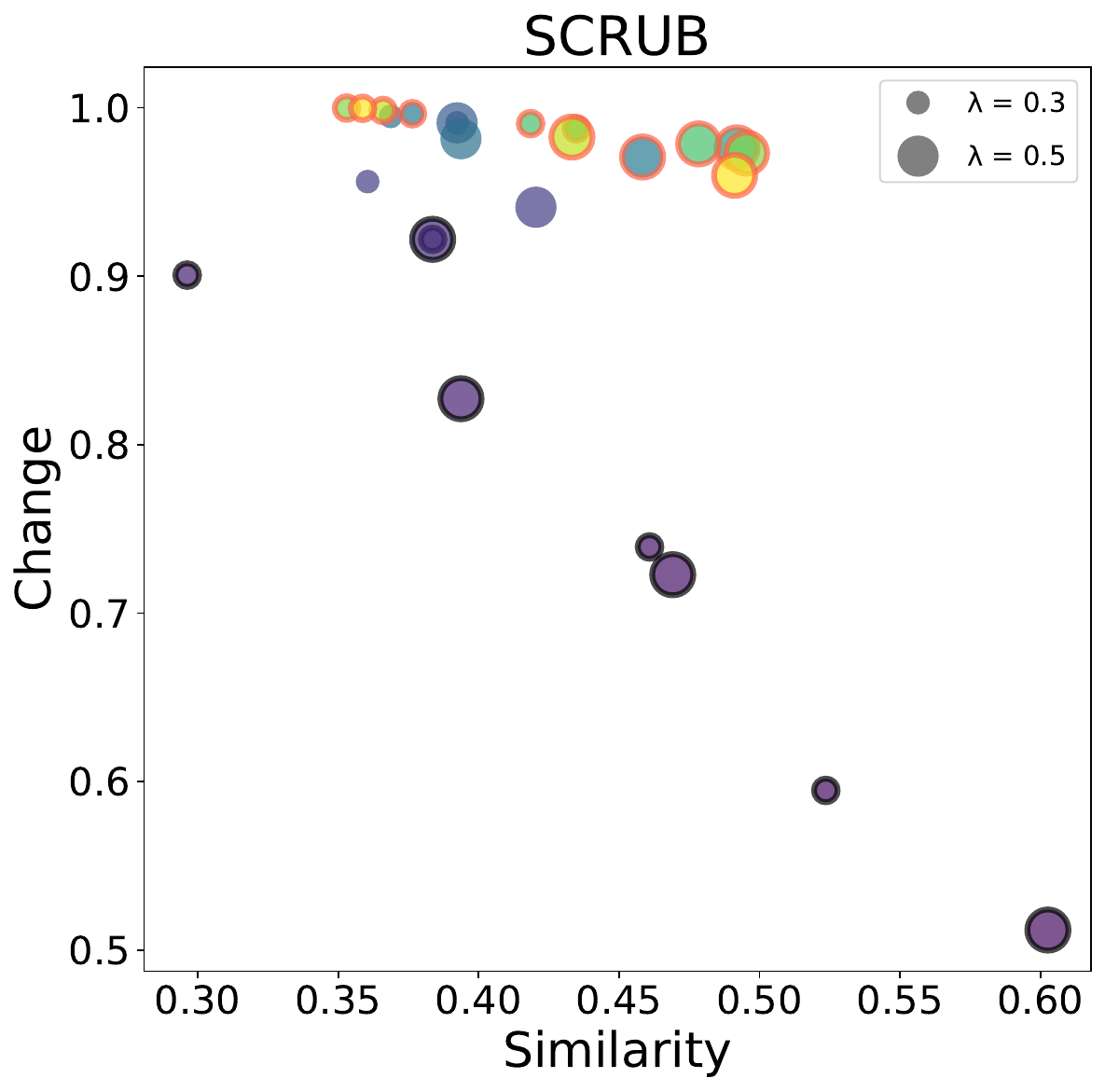}
    \includegraphics[height=0.3\textwidth]{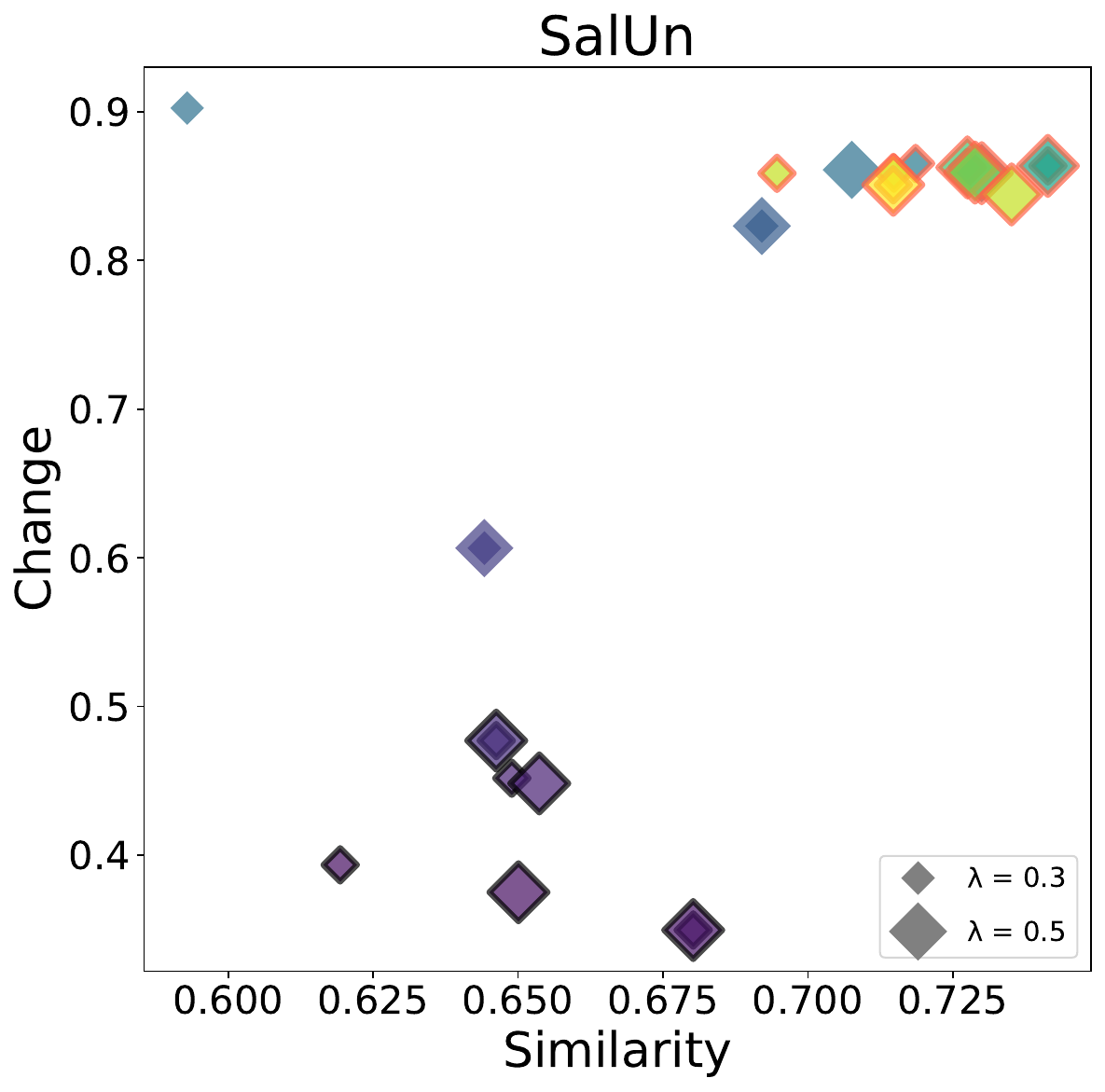}
    \includegraphics[height=0.3\textwidth]{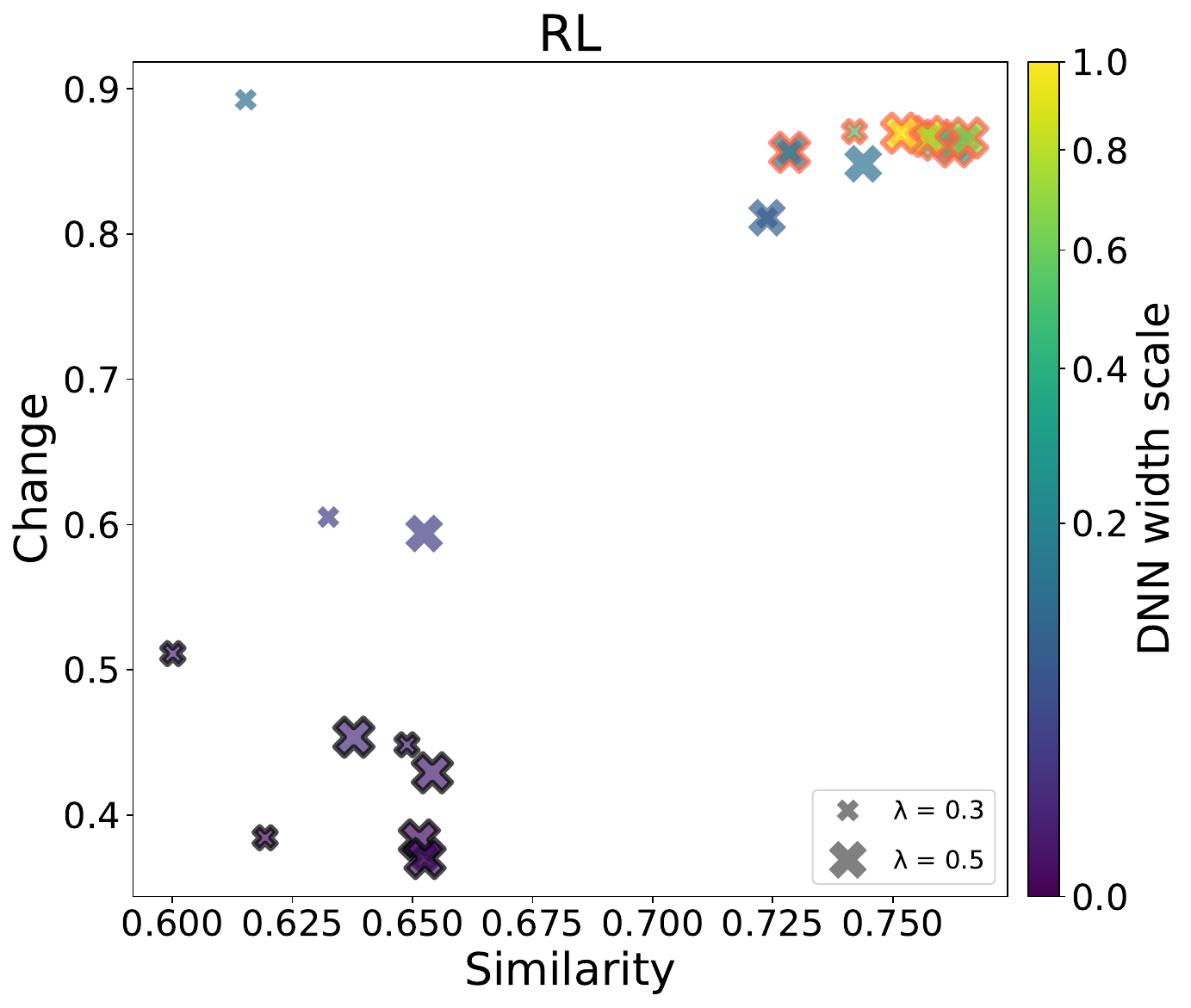}
    \\
    \includegraphics[height=0.3\textwidth]{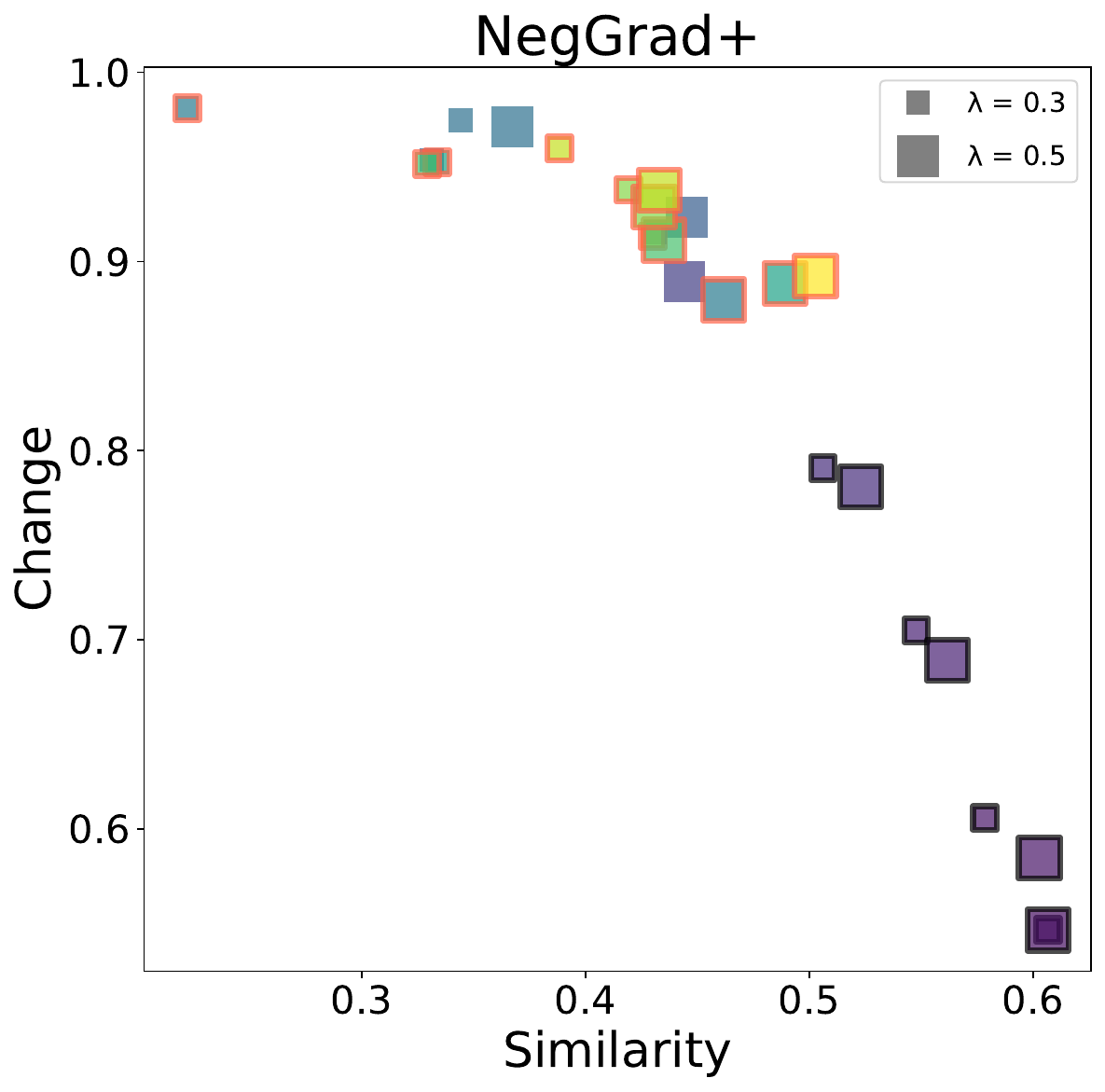}
    \includegraphics[height=0.3\textwidth]{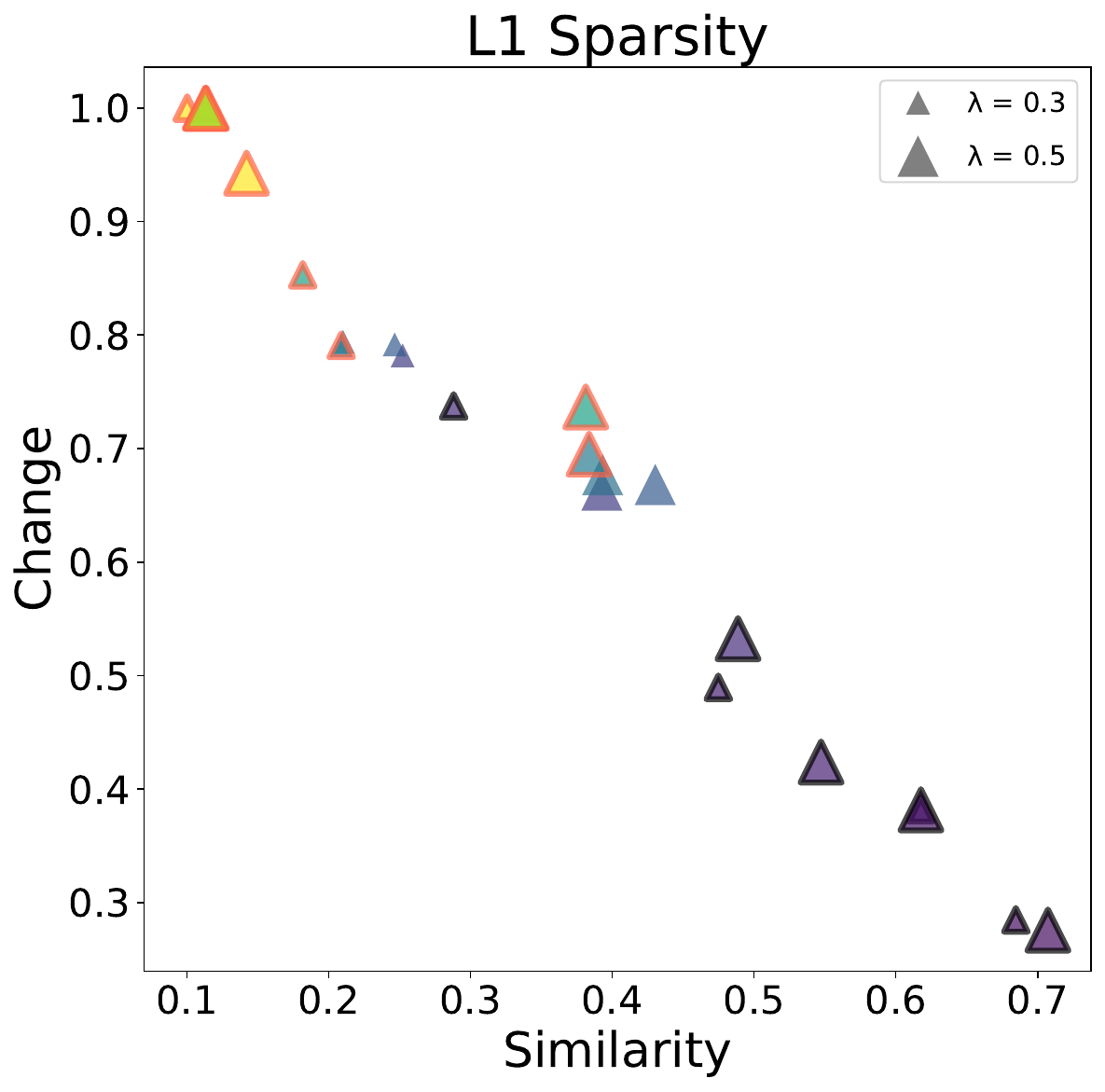}
    \includegraphics[height=0.3\textwidth]{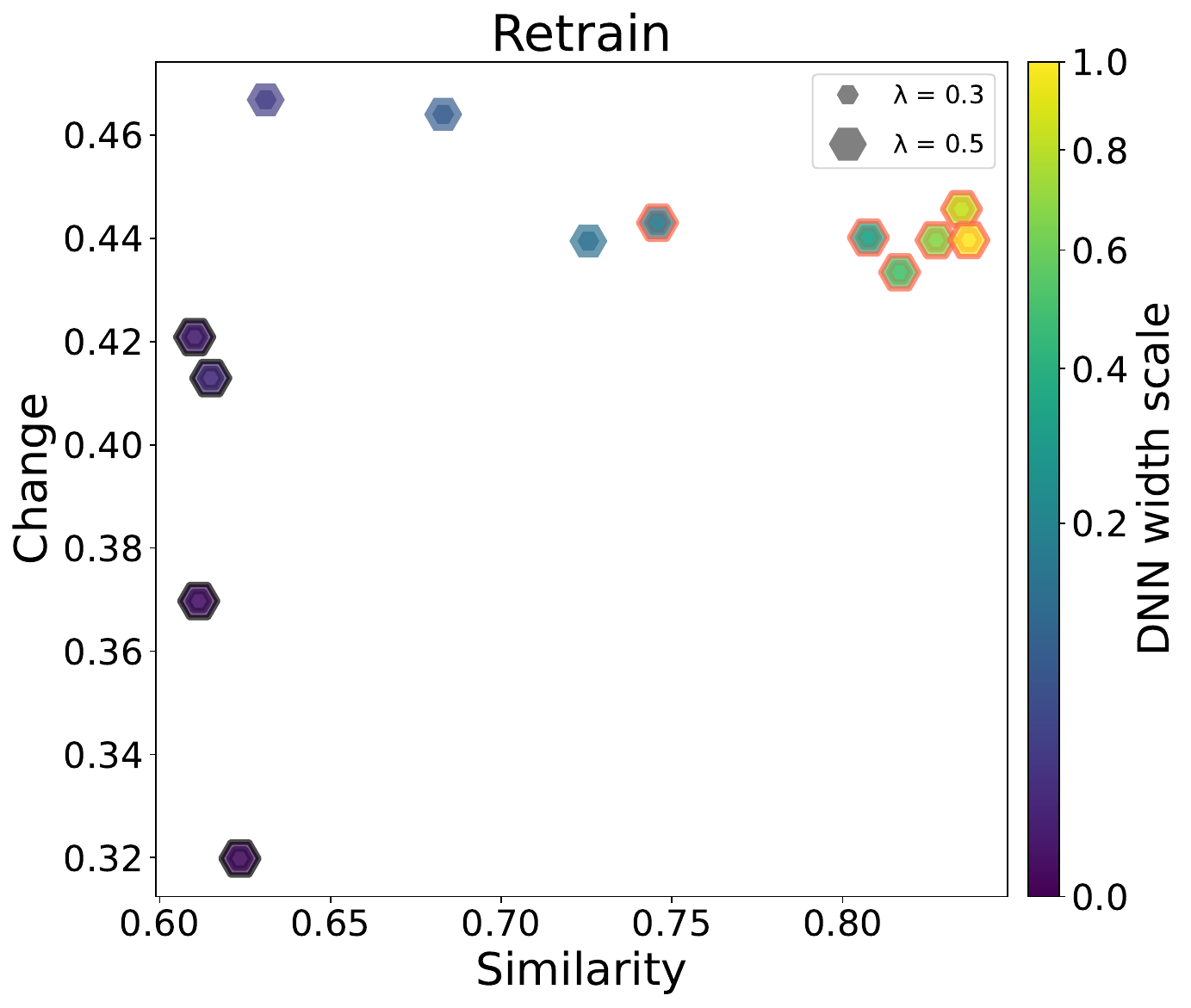}
    \caption{Decision-region similarity and change scores for \textbf{bias removal} unlearning. 3-layer FC network on CIFAR-10 (200 unlearned examples, $\delta=10$).  Markers closer to the top-right corner at each diagram denote unlearning with more-local changes of decision regions.}
    \label{fig:similarity_change_bias_fcnet3layer_cifar10_200unlearned}
\end{figure*}

In this appendix section, we provide additional results that examine the decision regions before and after unlearning. These results include measurements of the similarity and change scores for models, highlighting the changes in decision boundaries (see Figs.~\ref{fig:similarity_change_privacy_resnet50_tinet_1000unlearned}-\ref{fig:similarity_change_bias_fcnet3layer_cifar10_200unlearned}).

\begin{figure*}[]
    \centering    
    \includegraphics[height=0.3\textwidth]{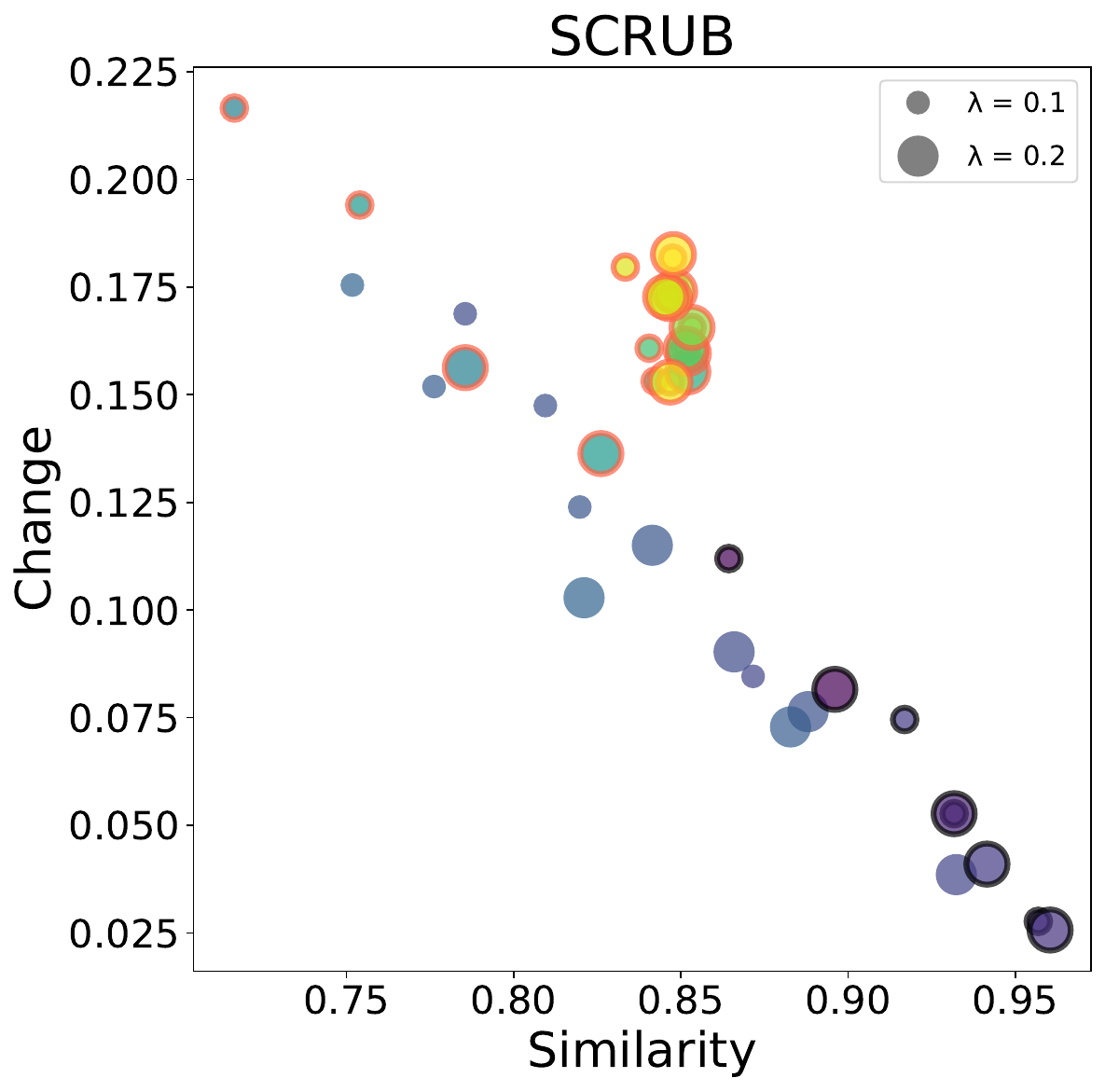}
    \includegraphics[height=0.3\textwidth]{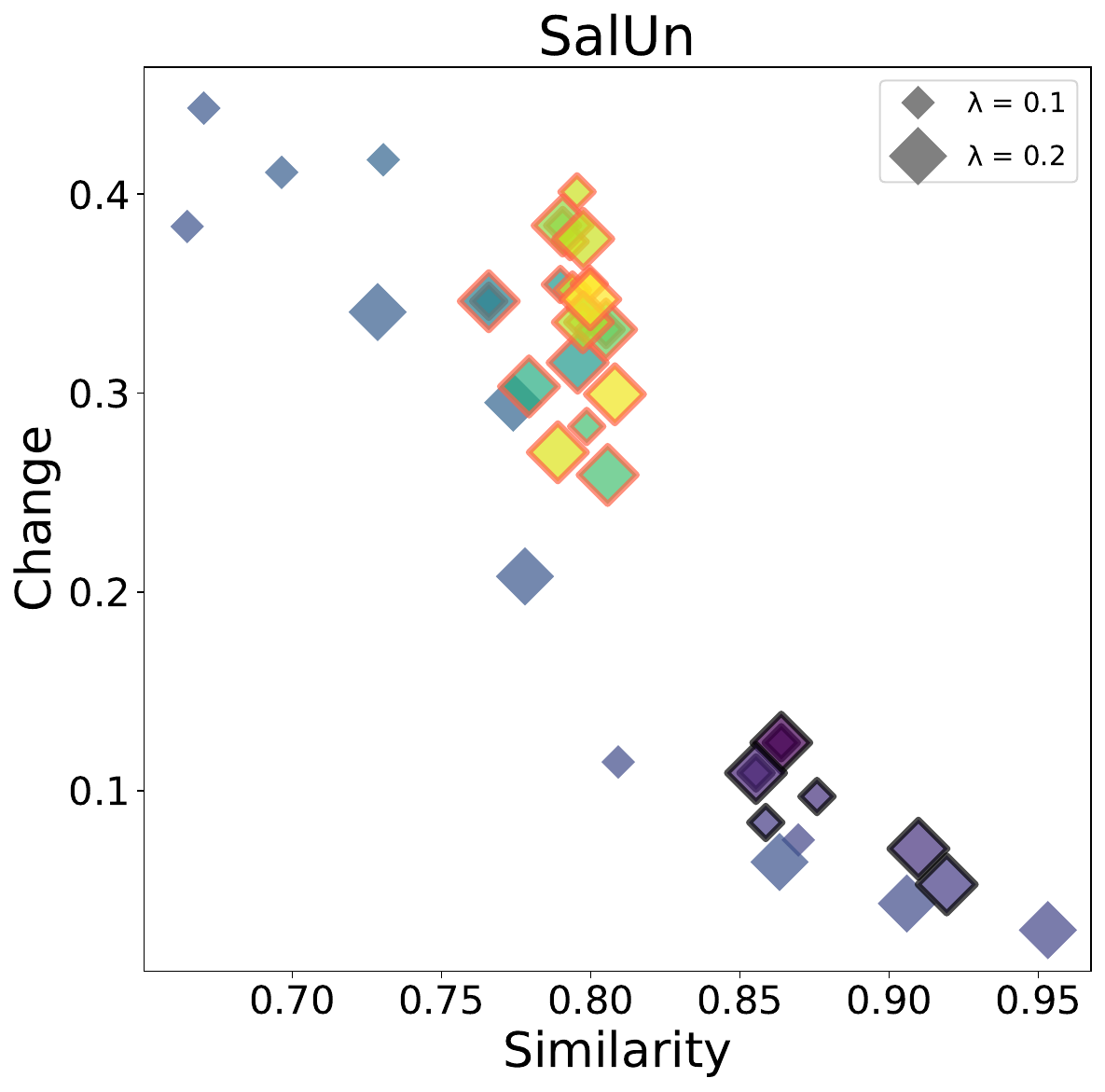}
    \includegraphics[height=0.3\textwidth]{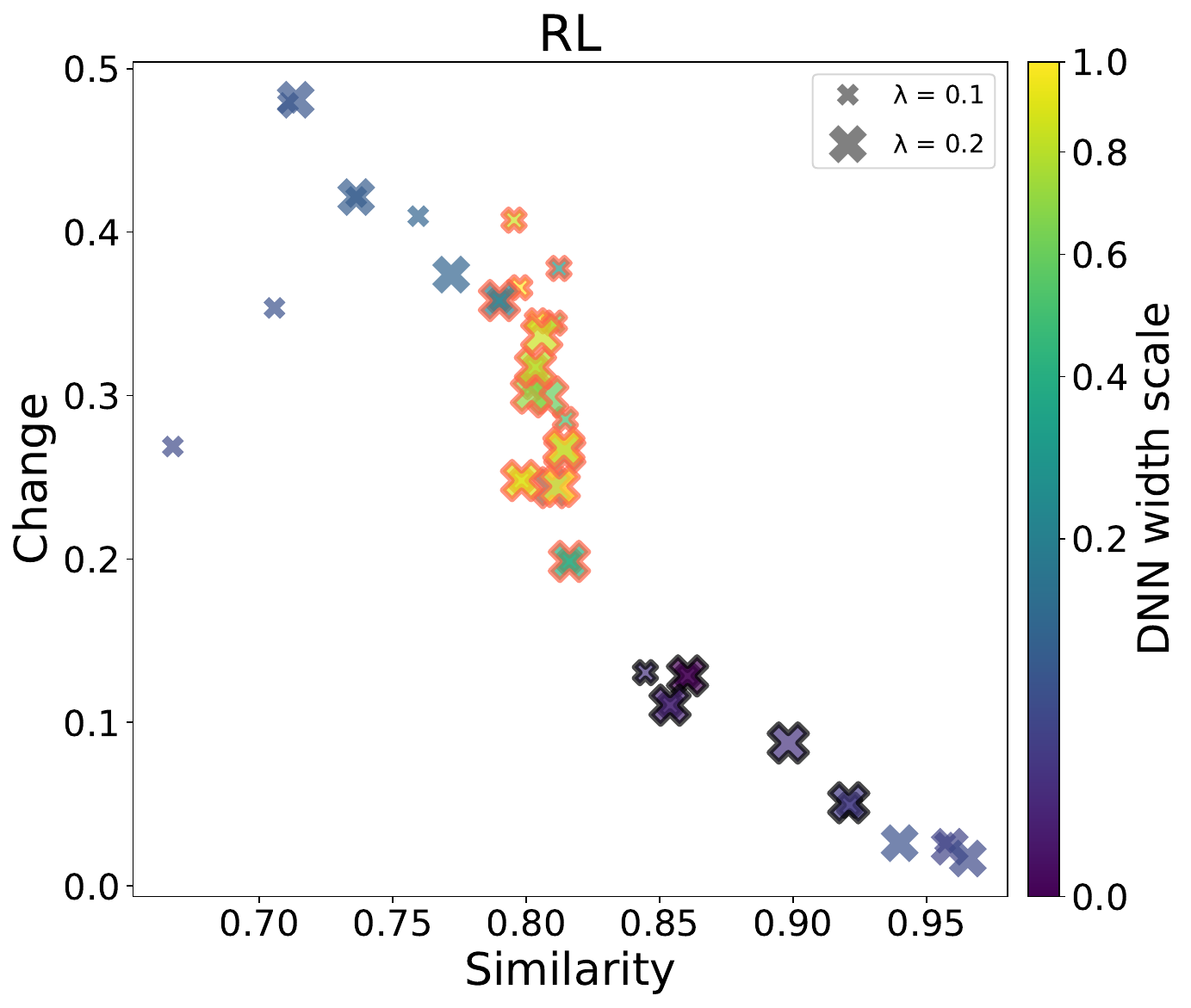}
    \\
    \includegraphics[height=0.3\textwidth]{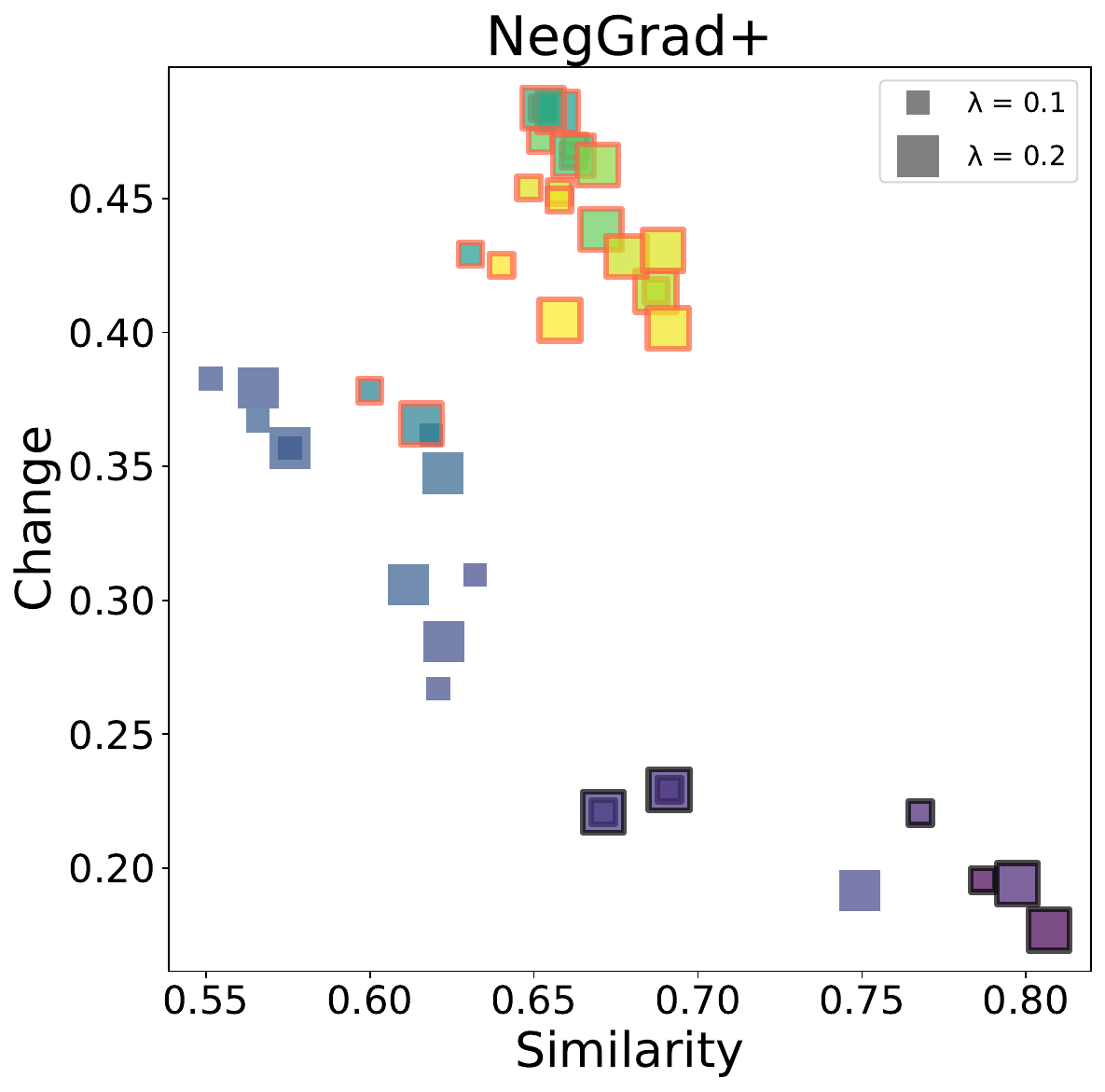}
    \includegraphics[height=0.3\textwidth]{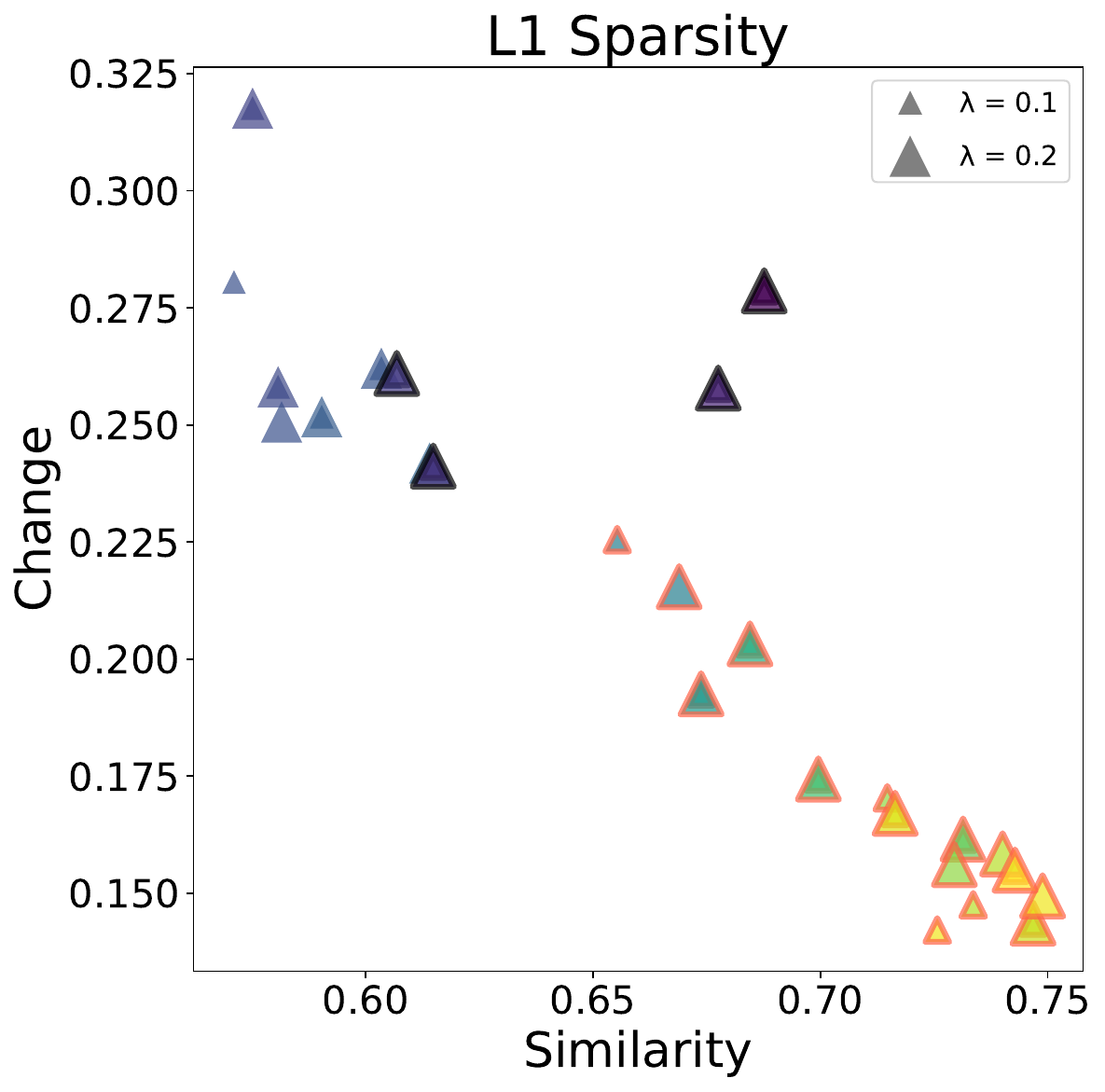}
    \includegraphics[height=0.3\textwidth]{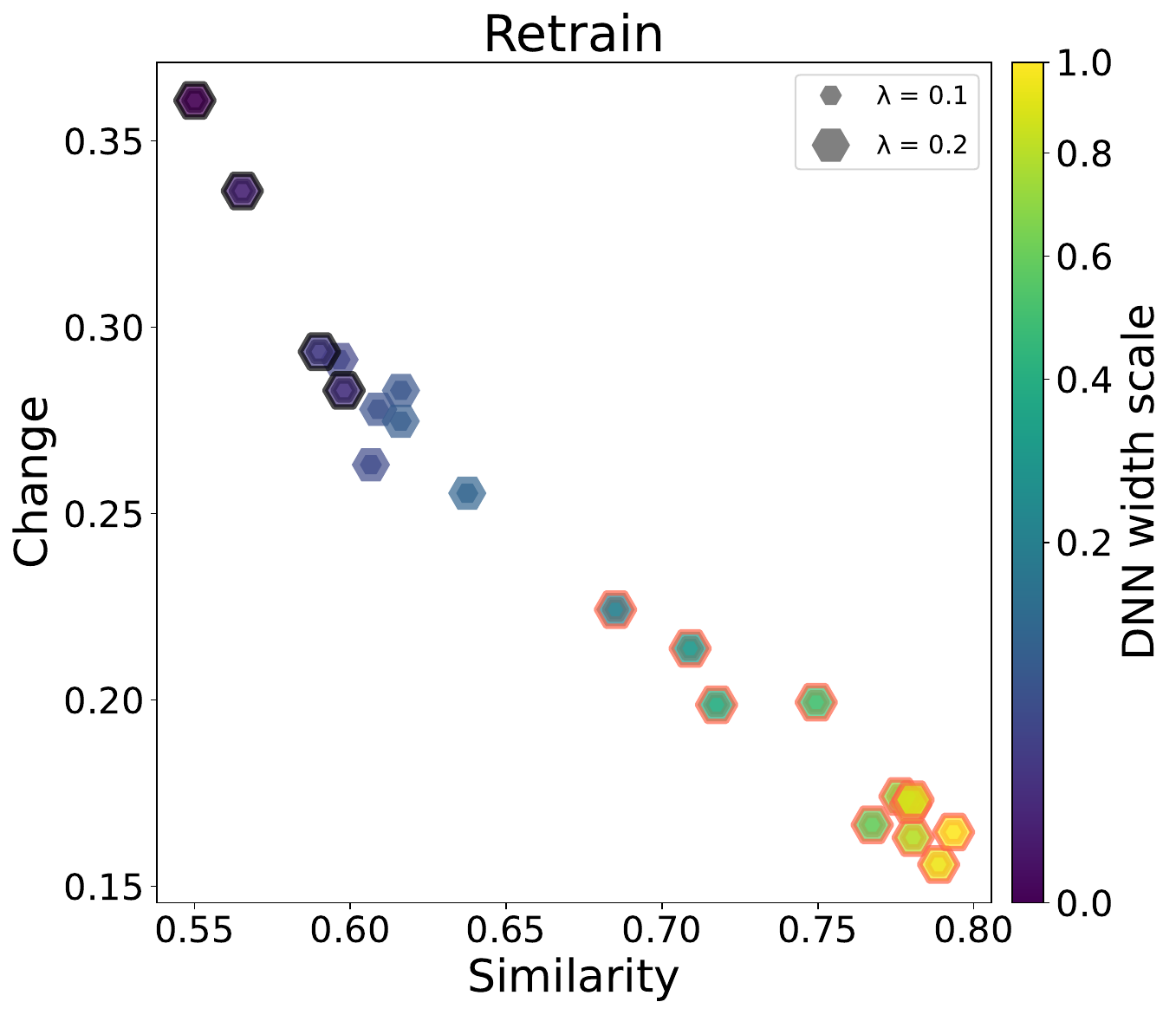}
    \caption{Decision-region similarity and change scores for \textbf{privacy} unlearning. ResNet-18 on CIFAR-10 (600 unlearned examples from three classes, $\delta=10$).  Markers closer to the top-right corner at each diagram denote unlearning with more-local changes of decision regions.}
    \label{fig:similarity_change_privacy_resnet_cifar10_600unlearned}
\end{figure*}

\begin{figure*}[]
    \centering    
    \includegraphics[height=0.3\textwidth]{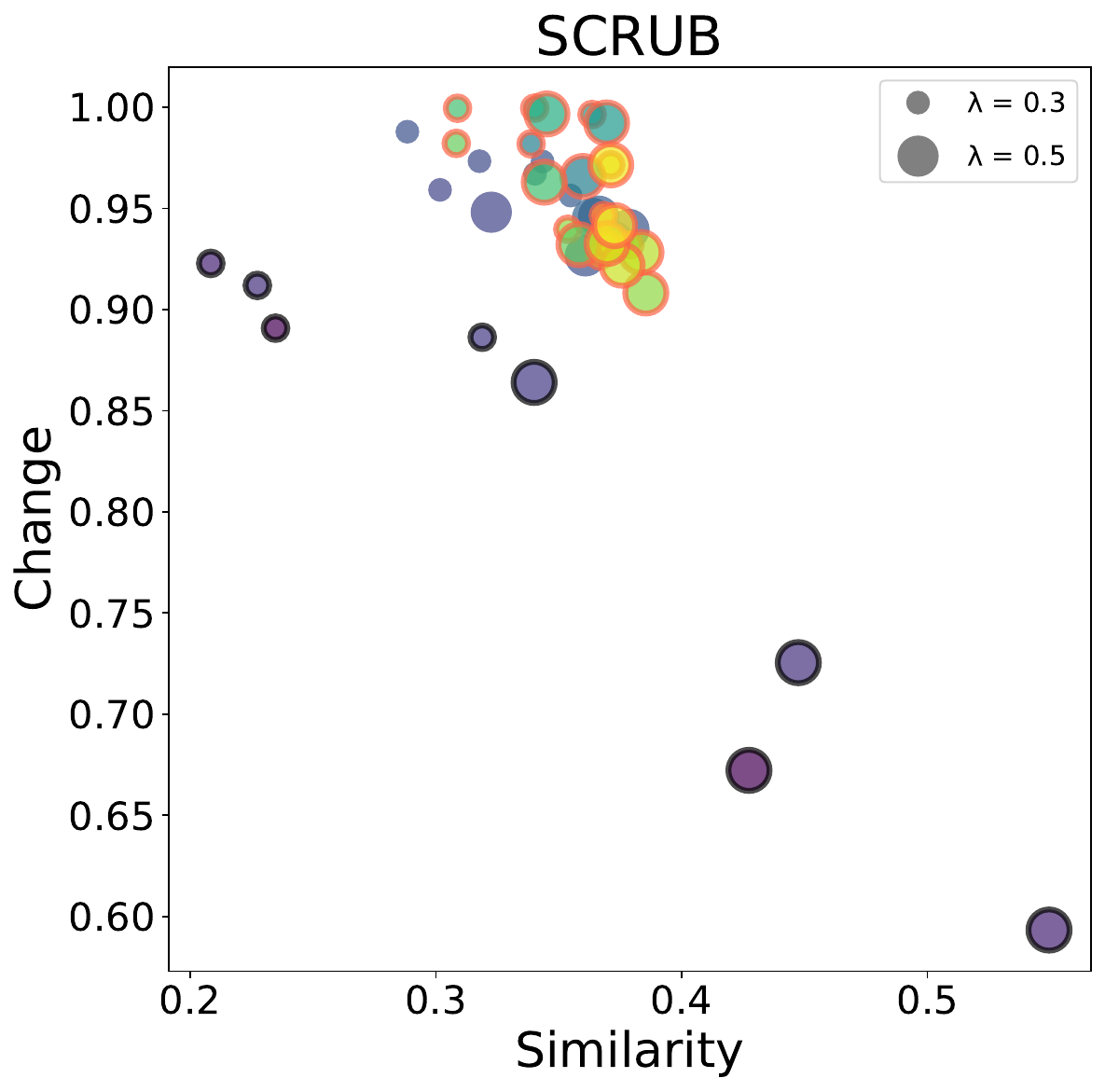}
    \includegraphics[height=0.3\textwidth]{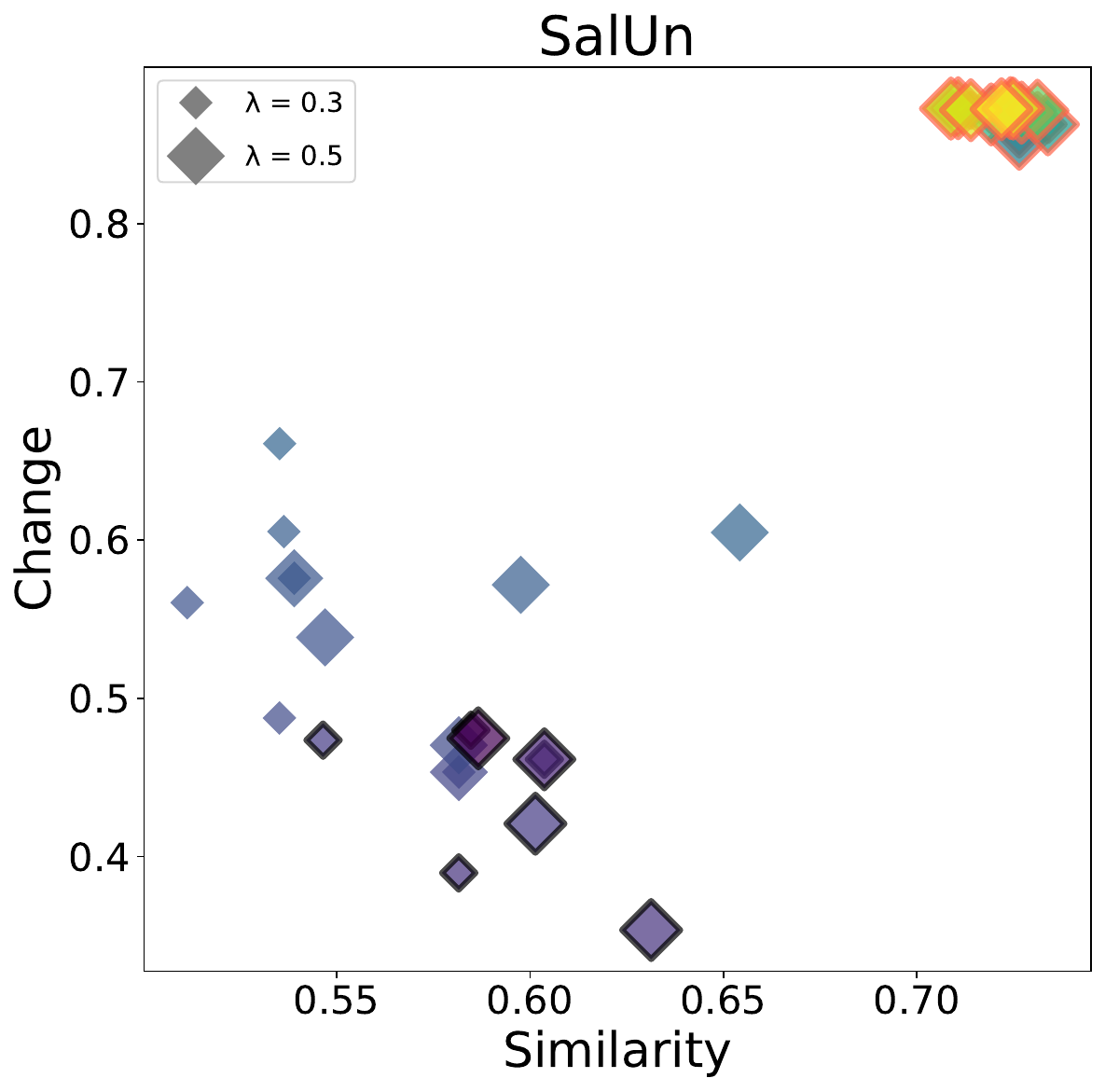}
    \includegraphics[height=0.3\textwidth]{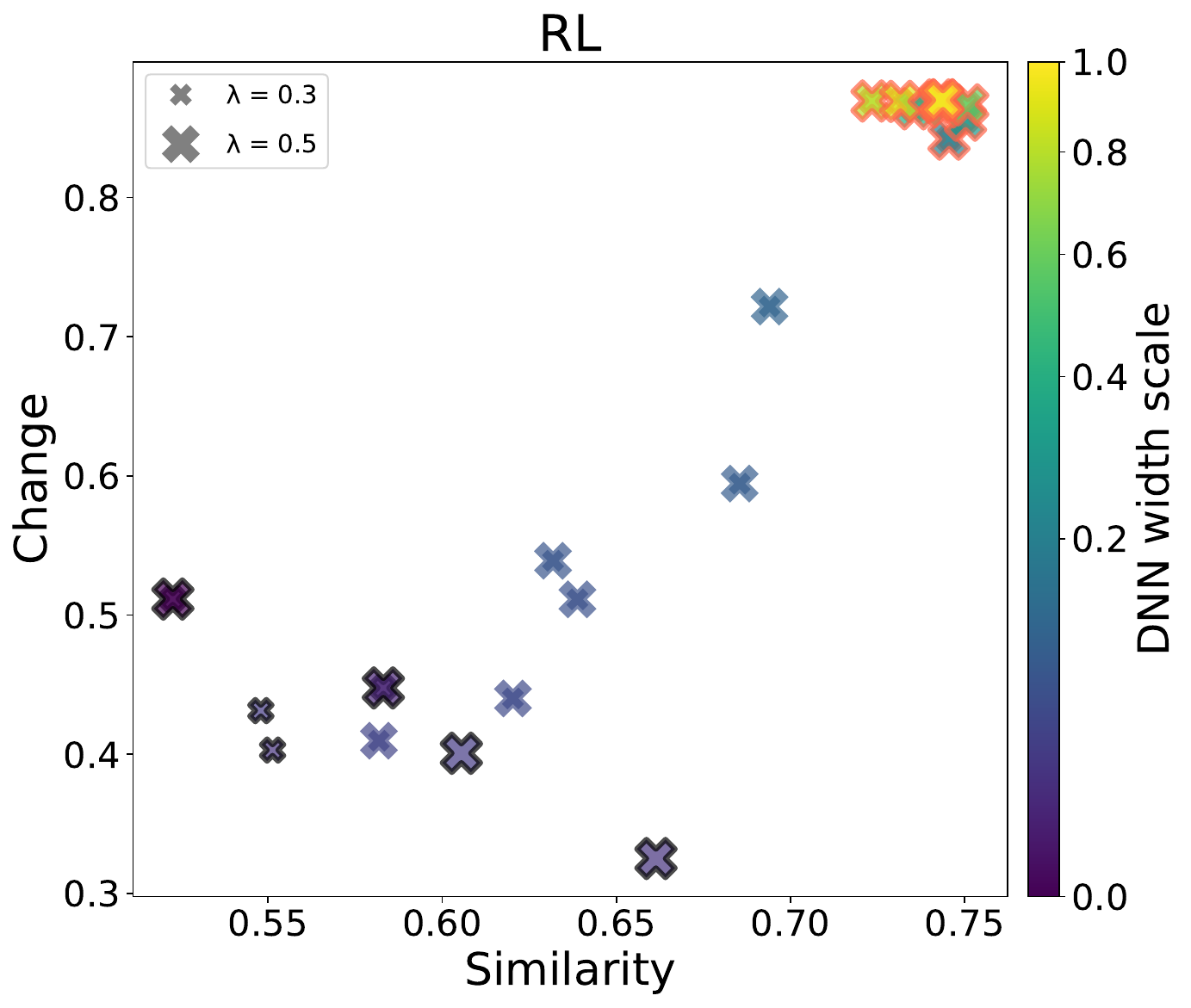}
    \\
    \includegraphics[height=0.3\textwidth]{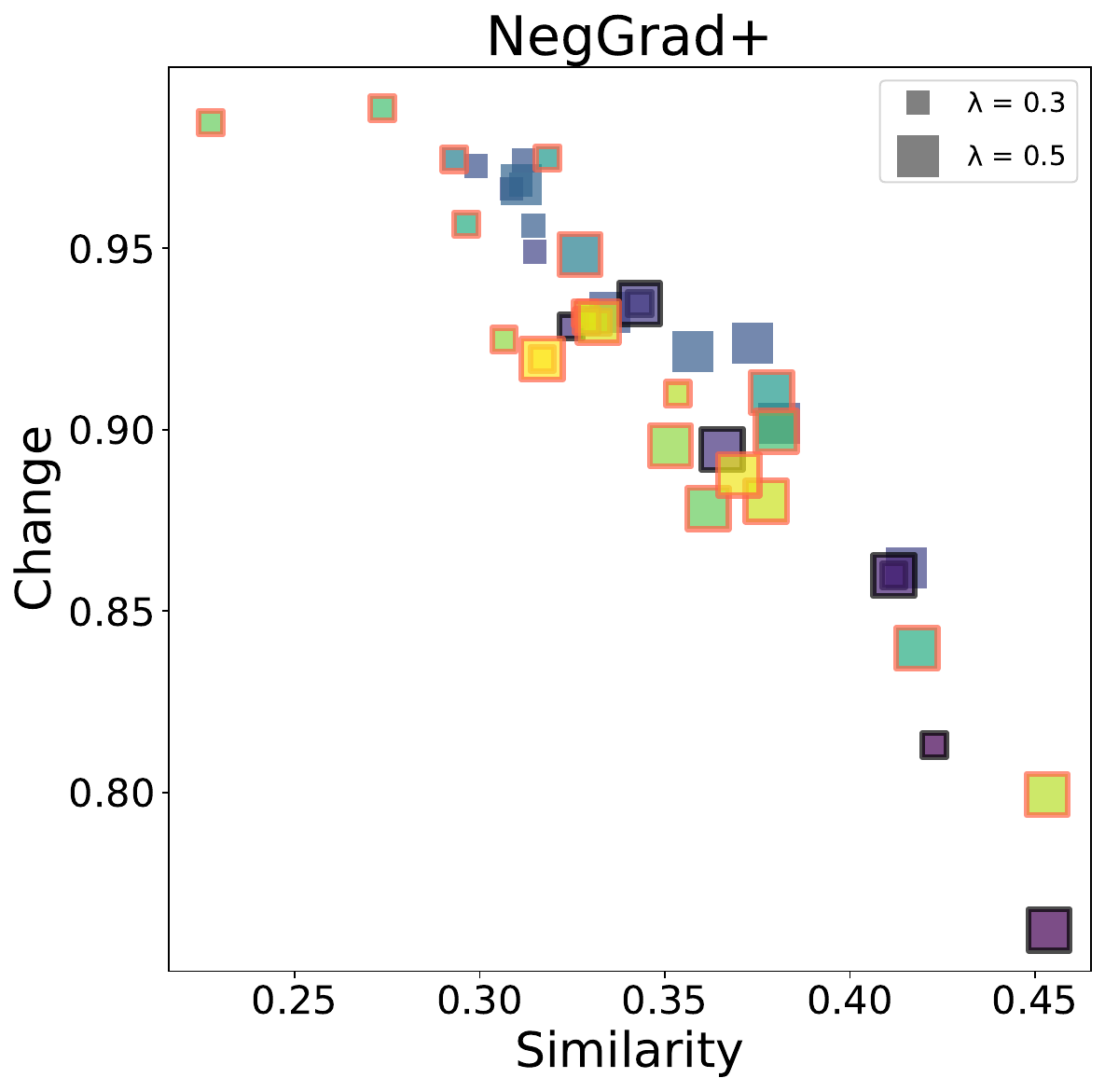}
    \includegraphics[height=0.3\textwidth]{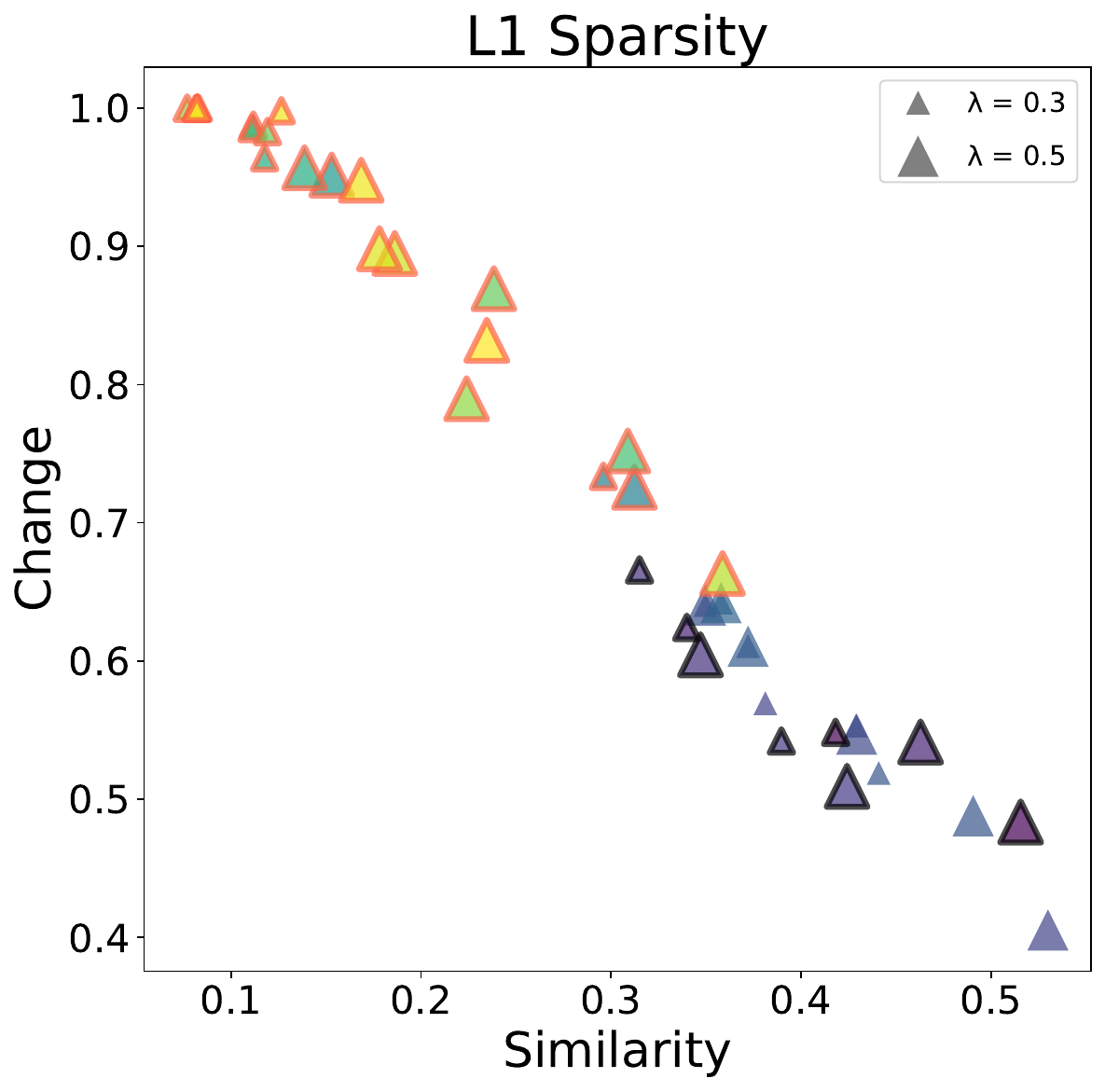}
    \includegraphics[height=0.3\textwidth]{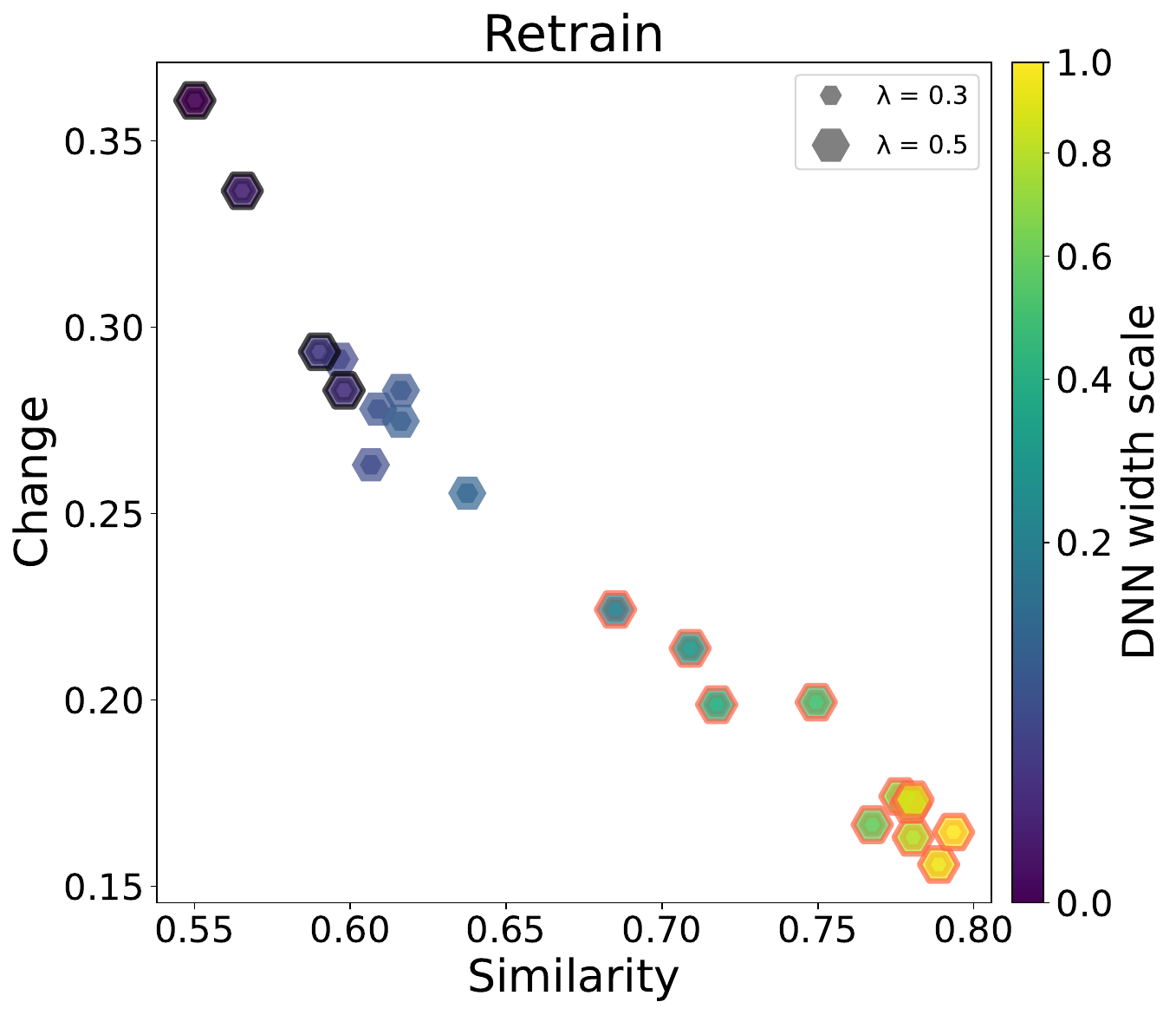}
    \caption{Decision-region similarity and change scores for \textbf{bias removal} unlearning. ResNet-18 on CIFAR-10 (600 unlearned examples, $\delta=10$).  Markers closer to the top-right corner at each diagram denote unlearning with more-local changes of decision regions.}
    \label{fig:similarity_change_privacy_resnet_cifar10_600unlearned}
\end{figure*}

\end{document}